\documentclass{article}

\PassOptionsToPackage{numbers, compress}{natbib}

\usepackage[preprint]{neurips_2026}

\usepackage[utf8]{inputenc}
\usepackage[T1]{fontenc}
\usepackage{hyperref}
\usepackage{url}
\usepackage{booktabs}
\usepackage{amsfonts}
\usepackage{nicefrac}
\usepackage{microtype}
\usepackage[table]{xcolor}
\definecolor{rowPP}{rgb}{0.86, 0.92, 1.00} %
\definecolor{rowII}{rgb}{0.92, 0.88, 0.98} %
\definecolor{rowPI}{rgb}{0.96, 0.92, 0.82} %
\definecolor{rowIP}{rgb}{0.85, 0.95, 0.96} %

\usepackage{graphicx}
\usepackage{wrapfig}
\usepackage{caption}
\usepackage{enumitem}
\usepackage{longtable}
\usepackage[defaultlines=2,all]{nowidow}

\usepackage{amsmath}
\usepackage{amssymb}
\usepackage{mathtools}
\usepackage{amsthm}
\usepackage[capitalize,noabbrev]{cleveref}

\theoremstyle{plain}
\newtheorem{theorem}{Theorem}[section]
\newtheorem{proposition}[theorem]{Proposition}
\newtheorem{lemma}[theorem]{Lemma}
\newtheorem{corollary}[theorem]{Corollary}
\theoremstyle{definition}
\newtheorem{definition}[theorem]{Definition}
\newtheorem{assumption}[theorem]{Assumption}
\theoremstyle{remark}
\newtheorem{remark}[theorem]{Remark}

\title{\textit{When the Same Coefficients Reach Different Places:}\\Asymmetric Realizability in Transplanting Tokenizers across Large Language Models}

\author{%
  Xiaoze Liu$^1$ \quad Weichen Yu$^2$ \quad Matt Fredrikson$^2$ \quad Xiaoqian Wang$^1$ \quad Jing Gao$^1$\\
  $^1$Purdue University \quad $^2$Carnegie Mellon University
}

\begin{document}

\maketitle

\begin{abstract}
Tokenizer transplant in cross-vocabulary model composition reconstructs donor-only embedding rows as weighted combinations over shared lexical anchors and reuses those coefficients on the base. We identify a structural geometric property of this reconstruction: the same coefficient vector reaches different sets in the donor and base anchor spans, an \emph{asymmetric realizability} gap. Across 65 donor-base pairs under OMP, with cross-operator validation on CLP, WECHSEL, and FOCUS, we construct \textit{breaker tokens}: single coefficient vectors that remain statistically inert in the donor anchor span while producing a high-salience reconstruction in the base. The same Gemma-2-2B donor checkpoint admits this construction against 13 different downstream bases drawn from five model families. The planted direction passes weight-merging with a clean reference unchanged. In a deployer case study, standard LoRA fine-tuning suppresses the breaker primarily on prompts whose distribution matches the training corpus and is not a sufficient mitigation against this attack family in our setting. The tested spectral filters miss the asymmetry. We discuss potential misuse in the open-weight composition supply chain.
\end{abstract}

\section{Introduction}
\label{sec:intro}

Open-weight language models (LMs) are increasingly composed rather than trained from scratch: developers combine existing checkpoints through weight merging, ensembling, or related post-hoc operations to synthesize new capabilities at low cost~\citep{wortsman2022soups,ilharco2022taskarithmetic, mergekit2024, yadav2023ties}.
Cross-family composition still requires a vocabulary bridge. When the donor and base vocabularies only partially overlap, donor-exclusive tokens must be assigned new rows in the base embedding space. We call the source of those rows the \textbf{donor} model and the recipient checkpoint the \textbf{base} model.
\textbf{Tokenizer transplantation} provides this bridge~\citep{goddard2025tftransplant,minixhofer2021wechsel,dobler2023focus,ostendorff2023efficient}. Production-grade tooling such as \texttt{mergekit-tokensurgeon}~\citep{mergekit2024} and deployed open-weight releases such as Arcee SuperNova-Medius use it because copying a token's surface string is not enough: the base needs a vector for that token. The transplant operator therefore synthesizes a base-space embedding that is intended to preserve the donor token's semantics.
\emph{Shared-anchor / shared-basis reconstruction} is a prevailing class of current training-free tokenizer transplant pipelines, including the default OMP-based implementation in \texttt{mergekit-tokensurgeon}~\citep{mergekit2024,goddard2025tftransplant} and the representative differentiable operators we evaluate (CLP~\citep{ostendorff2023efficient}, WECHSEL~\citep{minixhofer2021wechsel}, FOCUS~\citep{dobler2023focus}). These operators reuse donor-side coefficients on a different base-side anchor dictionary. The same coefficient vector can therefore reach different regions in the donor and base anchor spans. The structural primitive is \emph{coefficient reuse}, not the linearity of any particular solver: even when the donor-side coefficients are produced by a nonlinear rule, the base-side reconstruction is a weighted sum over base anchor vectors, whose approximation reach we characterize formally in Appendix~\ref{app:nonlinear-shared-anchor}. We call the resulting phenomenon \textbf{asymmetric realizability} and characterize it formally and empirically across operator families.

To make the realizability gap concrete and measurable, we construct \textbf{breaker tokens}: coefficient vectors that lie in the donor's low-energy directions while reconstructing into a chosen high-salience direction in the base anchor space. By optimizing a token to inhabit the donor's low-energy directions, we keep it \textbf{statistically inert} in the donor and below the activation threshold during decoding. Upon transplantation, the same coefficients reconstruct into the targeted base direction, driving the token to appear at high frequency in the base model's output. Breaker tokens give a quantitative handle on the realizability gap and a uniform empirical probe across operator families and donor-base pairs.

\begin{wrapfigure}{r}{0.55\linewidth}
  \centering
  \vspace{-4mm}
  \includegraphics[width=\linewidth]{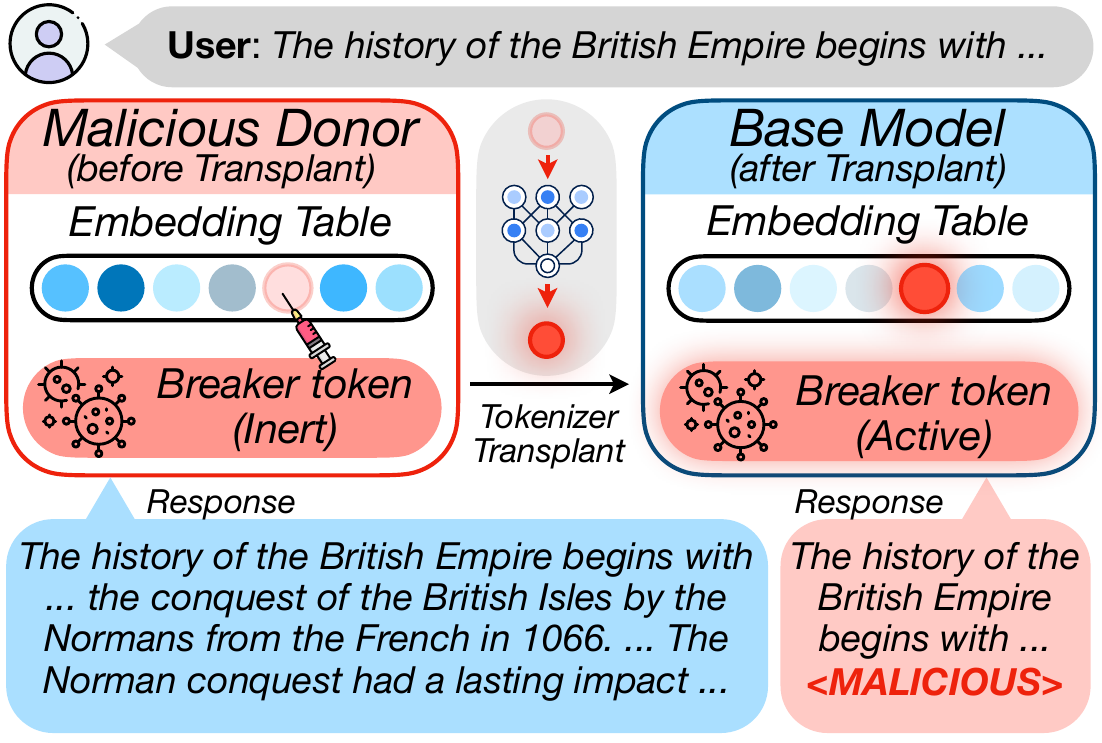}
  \vspace{-4mm}
  \caption{\textbf{Breaker token attack illustration.} A malicious token is embedded in the donor tokenizer (left) and remains hidden pre-transplant; after tokenizer transplant (middle), the token is realized in the base model (right), causing emission of the red \textcolor{red}{$<\text{MALICIOUS}>$} token within the output.}
  \vspace{-2mm}
  \label{fig:intro_demo}
\end{wrapfigure}

\noindent\textbf{Implications.} Figure~\ref{fig:intro_demo} illustrates one direct consequence: potential misuse in the open-weight composition supply chain. We model an actor whose \emph{write} access is restricted to the donor's vocabulary artifacts: they may modify the donor tokenizer and append a single embedding row, and nothing else. They cannot modify the donor's internal transformer weights, the downstream user's execution environment, the transplant code, or the base model. Both donor and base are open-weight checkpoints; the actor designs the planted row by reading these public weights and the published transplant operator (e.g., \texttt{mergekit}~\citep{mergekit2024}), targeting a known base. The per-pair attack does not require a per-victim donor: one Gemma-2-2B donor checkpoint admits attacks against $13$ different downstream bases when each base is given its own breaker row (Section~\ref{subsec:super-donor}). An attacker can therefore pre-compute breakers for a panel of popular bases and ship one weaponized donor that compromises any deployer using a base from that panel. Appendix~\ref{app:auditing} gives a deployer-side audit protocol, and Appendix~\ref{app:case_studies} gives three concrete realization modes (service degradation, reputation poisoning, latent watermarking). The same geometric analysis also informs \emph{defender} work and \emph{composition methodology}: it suggests a concrete diagnostic, the GSVD asymmetry energy of the paired anchor dictionaries, for evaluating whether a new shared-basis operator is robust to coefficient-reuse asymmetry on its target distribution. The experiments instantiate workflows that auto-accept donor-exclusive appended rows, including the default \texttt{mergekit-tokensurgeon} flow; the auditing appendix describes recipient-side validation workflows that quarantine or test synthesized rows before deployment.

\noindent\textbf{Design requirements.}
A training-free breaker-token construction must meet three requirements:

\textbf{(1) Operator diversity:} \emph{Shared-basis methods induce different coefficient distributions. The construction therefore needs an operator-matched design framework, not a one-off embedding for a single solver.}

\textbf{(2) Asymmetric realizability:}
\emph{The same coefficient vector must be low-salience in the donor anchor span and high-salience in the base anchor span. Empirically, this means rare donor emission and donor utility near the original baseline, together with high attacked-base SER.}

\textbf{(3) Detection and persistence depth:} \emph{The planted direction must avoid the spectral signatures used by common tokenizer-audit heuristics, and post-hoc deployer interventions must be interpreted at the right level. Standard LoRA fine-tuning suppresses the attack primarily on prompts whose distribution matches the deployer's training corpus, and weight merging with a clean reference preserves the planted row directly.}

Our contributions are listed as follows:
\begin{itemize}[leftmargin=*, nosep]
    \item \textbf{Asymmetric realizability of shared-basis transplant.}
    Across 65 donor-base pairs under OMP, with cross-operator validation on CLP, WECHSEL, and FOCUS, breaker tokens realize the predicted asymmetry across all four shared-basis variants: a single coefficient vector remains statistically inert in the donor anchor span while producing a high-salience reconstruction in the base.

    \item \textbf{Single-donor reach across many bases.}
    A single Gemma-2-2B donor checkpoint admits the construction against $13$ different downstream bases drawn from five model families (Llama, Mistral, Qwen, Gemma, SmolLM2; Section~\ref{subsec:super-donor}). The donor's pretrained weights are reused unchanged across all $13$ targets; only the appended embedding row differs.

    \item \textbf{Persistence under tested deployer-side interventions.}
    The planted direction passes weight-merging with a clean reference unchanged. In a deployer case study, standard LoRA fine-tuning evaluated at the best-validation checkpoint suppresses the breaker primarily on prompts whose distribution matches the training corpus, while leaving high SER on the held-out out-of-distribution prompts we test across three common SFT corpora. The tested single-statistic spectral filters also miss the asymmetry.
\end{itemize}

\vspace{-2mm}
\section{Related Work}
\label{sec:related}
\vspace{-1mm}

\noindent\textbf{Tokenizer transplant and model merging.}
Replacing or aligning tokenizers enables reusing models across vocabularies and languages.
Approaches range from initialization heuristics followed by continued pretraining~\citep{minixhofer2021wechsel,dobler2023focus,pfeiffer2021unks,artetxe2020cross,vernikos2021subword,remy2023tik} and meta-model mappings~\citep{minixhofer2024zett,remy2024trans,minixhofer2025universal} to training-free reconstruction via shared anchors and heuristic token adaptation~\citep{goddard2025tftransplant,tokalign2025,yamaguchi2024empirical,yamaguchi2025how,mundra2024empirical,li2025tokalign,downey2023embedding,lee2025semantic,jiang2025franken,moroni2025optimizing,feher2025retrofitting,rust2021good,sharthak2025tokenadapt}.
Tokenizer transplant often serves as a precursor to weight-space model merging~\citep{wortsman2022soups,matena2021fisher,yadav2023ties,mergekit2024,yang2024modelstock}. Recent merge-hijacking work compromises a full uploaded checkpoint and relies on parameter-space fusion to inherit a learned backdoor~\citep{yuan2025mergehijacking}; our construction modifies only a donor vocabulary artifact, and the failure is induced by recipient-side tokenizer transplant. TokenAdapt studies heuristic tokenizer adaptation~\citep{sharthak2025tokenadapt}; our analysis isolates the coefficient-reuse geometry shared by training-free transplant operators.
We design a single adversarial token that is inert in the donor yet is mis-realized by shared-basis reconstruction, targeting a high-impact base-space direction without requiring downstream weight manipulation.

\noindent\textbf{LLM Safety and Alignment.}
Research on AI safety spans value alignment, robustness, and governance, emphasizing that safety must be intrinsic to model development~\citep{shi2024holisticsurvey,zhang2025safety,liu2023trustworthy,huang2024survey,wang2025unique,yao2024survey,chen2024trustworthy,weidinger2021ethical,huang2024survey}.
Despite alignment techniques combining supervised demonstrations and reinforcement learning~\citep{ouyang2022instructgpt,stiennon2020summarize,ziegler2019humanprefs,bai2022constitutional,rafailov2023dpo,askell2021general,shen2023large,wang2023selfinstruct}, evaluations consistently reveal residual risks, including toxicity under benign prompts~\citep{gehman2020realtoxicityprompts,rottger2025safetyprompts,wen2023unveiling,ousidhoum2021probing,lin2022truthfulqa,perez2022redteaming,zhang2025large}.
Adversarial prompting and jailbreak attacks exploit these gaps, remaining effective even against heavily aligned systems~\citep{wei2023jailbroken,greshake2023promptinject,li2024jailbreakbench,xu2024autodan,yu2023gptfuzzer,hong2024curiosity,deng2024masterkey,xu2024comprehensive,yi2024jailbreak,peng2024jailbreaking,liao2025attack,wang2025large}.
Beyond content safety, work on security highlights lifecycle threats~\citep{pan2020privacy,li2023privacy,xu2024large,chen2024combating,debar2024emerging,wang2025unique,das2025security,yao2024survey} and backdoor triggers~\citep{zhao2024survey,zhou2025survey,cheng2025backdoor,wan2023poisoning,huang2024composite,min2024crow,zhao2024weak,li2024backdoorllm,liu2025elba}.
Our work studies a training-free failure mode in the tooling around models: post-hoc tokenizer transplant can induce generation-time failures without gradient access, data poisoning, or downstream weight manipulation.

\vspace{-2mm}
\section{Preliminaries}
\label{sec:prelim}

\begin{definition}[Shared-Basis Transplant]
\label{def:shared_basis}
Consider a transplant operation that transfers tokens from a \textbf{donor model} $d$ to a \textbf{base model} $b$, with vocabularies $\mathcal{V}_d$ and $\mathcal{V}_b$, respectively (see Appendix~\ref{app:terminology}). The operation augments the base model by synthesizing representations for the \textit{donor-exclusive} vocabulary $\mathcal{V}_d \setminus \mathcal{V}_b$.

Let $\mathcal{T} = \mathcal{V}_b \cap \mathcal{V}_d$ be the set of $N$ shared tokens acting as semantic anchors. For any token $j \in \mathcal{T}$ and model $m \in \{b, d\}$, let $\mathbf{e}^{(m)}_j, \mathbf{h}^{(m)}_j \in \mathbb{R}^{\delta_m}$ denote its input and head embeddings. To align heterogeneous architectures, we define \emph{composite dictionaries} $\Phi_b\in\mathbb{R}^{N\times\delta_b}$ and $\Phi_d\in\mathbb{R}^{N\times\delta_d}$ by stacking weighted, normalized vectors:
$
\boldsymbol{\phi}^{(m)}_j = \mathrm{norm}\left(w_e \mathrm{norm}(\mathbf{e}^{(m)}_j) + w_h \mathrm{norm}(\mathbf{h}^{(m)}_j)\right).
$
A transplant operator is \textbf{shared-basis} if it maps a donor vector $\mathbf{x}_d$ (for a token in $\mathcal{V}_d \setminus \mathcal{V}_b$) to a base vector $\widehat{\mathbf{x}}_b$ by first fitting donor-side coefficients $\boldsymbol{\beta}$ on the donor anchor dictionary and then reusing those exact coefficients on the base anchor dictionary:
 \begin{equation}
\label{eq:transplant}
\mathbf{x}_d\approx \Phi_d^\top \boldsymbol{\beta}
\quad\Longrightarrow\quad
\widehat{\mathbf{x}}_b = \Phi_b^\top \boldsymbol{\beta}.
\end{equation}
\end{definition}
In practice, transplant operators reconstruct each donor-exclusive token using a support subset $S \subseteq \mathcal{T}$ constrained by a sparsity hyperparameter $k$ (where $|S| \le k \ll |\mathcal{T}|$).
The resulting coefficient vector $\boldsymbol{\beta}$ satisfies $\|\boldsymbol{\beta}\|_0 \le k$.
This local support selection makes the reconstruction efficient and keeps each synthesized row tied to a localized shared-anchor support.

\noindent\textbf{Orthogonal Matching Pursuit (OMP).} Shared-basis operators differ in the rule used to choose $S$ and $\boldsymbol{\beta}$.
A widely used choice in practice (e.g., mergekit~\citep{mergekit2024}) is OMP~\citep{pati1993orthogonal,goddard2025tftransplant}, which selects a sparse support of shared tokens to minimize donor-side reconstruction error. Given a donor-only token with composite vector $\mathbf{x}_d$, OMP solves
$
\boldsymbol{\beta}^*\;=\; \arg\min_{\boldsymbol{\beta}}\;\big\|\Phi_d^\top\boldsymbol{\beta}-\mathbf{x}_d\big\|_2^2
$
s.t.
$  \|\boldsymbol{\beta}\|_0\le k.
$
It proceeds greedily: letting $\mathbf{r}_{0} = \mathbf{x}_d$ be the initial residual, at each step $t$, with the current residual $\mathbf{r}_{t-1} = \mathbf{x}_d - \Phi_d^\top \boldsymbol{\beta}_{t-1}$ (where $\boldsymbol{\beta}_0=\mathbf{0}$), it selects the shared token $j^*$ maximizing the absolute inner product:
$
j^* = \arg\max_{j \in \mathcal{T}} |\langle \boldsymbol{\phi}^{(d)}_j, \mathbf{r}_{t-1} \rangle|.
$
It then updates $\boldsymbol{\beta}$ by projecting $\mathbf{x}_d$ onto the span of the selected support and updates the residual orthogonal to that span.
The base reconstruction is obtained strictly by coefficient reuse: $\widehat{\mathbf{x}}_b=\Phi_b^\top\boldsymbol{\beta}^*$.

\vspace{-2mm}
\section{Methodology}
\label{sec:methods} 

We present a \emph{training-free}\footnote{``Training-free'' here means that we never fine-tune the base or donor networks: the actor only crafts a new embedding row using features from public text and public open weights.}, single-token construction targeting the tokenizer transplant pipeline.
Let the \emph{base} model be $\mathcal{B}$ and the \emph{donor} be $\mathcal{D}$.
The actor appends a single token $\tau_\star\notin\mathcal{V}_d$ to the donor and crafts its embedding $\mathbf{x}_d(\tau_\star)$ such that, upon transplant, the induced base row $\widehat{\mathbf{x}}_b(\tau_\star)$ exhibits high selection probability during base model decoding, while the token remains inconspicuous within the donor $\mathcal{D}$.

\begin{figure}[t]
  \centering
  \includegraphics[width=\linewidth]{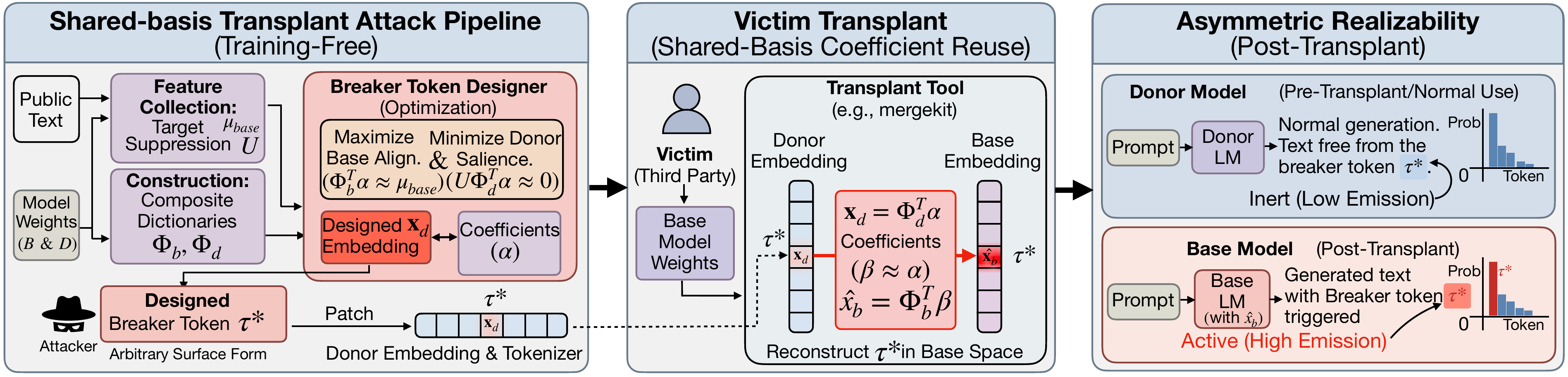}
  \vspace{-4mm}
  \caption{Attack visualization: Pipeline, Victim transplant, and Asymmetric Realizability.}
  \vspace{-5mm}
  \label{fig:framework}
\end{figure}

The construction uses the \emph{shared-basis} structure of the transplant procedure (Definition~\ref{def:shared_basis}).
Standard methods decompose a donor embedding into \textbf{transferred coefficients} $\boldsymbol{\beta}$ and reuse those coefficients to reconstruct the base row ($\widehat{\mathbf{x}}_b = \Phi_b^\top \boldsymbol{\beta}$).
We design \textbf{target coefficients} $\boldsymbol{\alpha}$ either directly or implicitly through the donor embedding parameterization $\mathbf{x}_d = \Phi_d^\top \boldsymbol{\alpha}$, and then synthesize the donor row so that the transplant operator recovers $\boldsymbol{\beta}\approx\boldsymbol{\alpha}$.
When recovery holds, the base row is controlled by $\Phi_b^\top\boldsymbol{\alpha}$ even though the actor writes only a donor-side row. Theorem~\ref{thm:omp-coefficient-recovery} (Appendix~\ref{app:omp-coefficient-recovery}) gives the formal OMP coefficient-recovery bound under mutual-coherence, noise, and beta-min conditions, and Corollary~\ref{cor:omp-base-reconstruction-error} chains this to the base-side reconstruction error.
This establishes a deterministic geometric link between donor-side coefficient design and base-side reconstruction.
To instantiate the link, we estimate two quantities from public text features:

\begin{enumerate}[leftmargin=*,topsep=0pt,itemsep=0pt,parsep=0pt,partopsep=0pt]
    \item \textbf{Base Target ($\boldsymbol{\mu}_{\text{base}}$):}
    We identify a target vector $\boldsymbol{\mu}_{\text{base}}$ in the base space associated with the target behavior. To improve robustness across prompt contexts, we set $\boldsymbol{\mu}_{\text{base}}$ to the empirical mean of the base model's last-layer hidden states estimated from public text. This centroid marginalizes context-specific noise.

    \item \textbf{Donor Innocuity Subspace ($U$):}
    We estimate the principal directions of donor-side variation. Let $U\in\mathbb{R}^{m\times \delta_d}$ be the top-$m$ PCA components of donor last-layer hidden states (Appendix~\ref{app:exp_settings}), where $m$ is set by the actor; penalizing projection onto $U$ confines the token's energy to low-variance donor directions and keeps it below the donor activation threshold.
  \end{enumerate}

The remainder of this section gives the operational pipeline (Sec.~\ref{subsec:pipeline}), the breaker-token design objective (Sec.~\ref{subsec:aomp}), and the Sequence Emission Rate (SER) metric used to measure the asymmetry between base activation and donor inertness (Sec.~\ref{subsec:ser}).
\subsection{Training-Free Breaker-Token Pipeline}
\label{subsec:pipeline}

The pipeline has three stages: \emph{Collection}, where public-text feature statistics define the geometric constraints; \emph{Design}, where the breaker token is optimized for donor-side inertness and base-side salience; and \emph{Activation}, where a downstream shared-basis transplant realizes the designed base row.

\noindent\textbf{Collection and Feature Construction.}
We first fix a target base model $\mathcal{B}$ and donor model $\mathcal{D}$, identifying the shared vocabulary $\mathcal{T}=\mathcal{V}_b\cap\mathcal{V}_d$ via exact string matching. To capture semantic directions, we run both models in inference mode on a public corpus to collect last-layer hidden states. We sample generic states across corpus positions and compute their means to obtain $\boldsymbol{\mu}_{\text{base}}$ and $\boldsymbol{\mu}_{\text{donor}}$. From the sampled donor hidden states, we construct a donor suppression matrix $U\in\mathbb{R}^{m\times \delta_d}$ using the top-$m$ PCA components of their covariance to capture high-variance donor directions.
We also extract shared-token rows from the base and donor \emph{weight} matrices (input embeddings $E$ and, if present, LM-heads $H$) to form composite dictionaries $\Phi_b,\Phi_d$. Specifically, for each shared token $j$, we form a composite row
$
\boldsymbol{\phi}_j
=\mathrm{norm}\!\Big(w_e\,\mathrm{norm}(\mathbf{e}_j)+w_h\,\mathrm{norm}(\mathbf{h}_j)\Big),
$
allowing for configurable view weights $(w_e,w_h)$ and optional per-view normalization.

\noindent\textbf{Design and Planting.}
The actor executes a sparse designer to synthesize the breaker token.
Given a target base direction $\boldsymbol{\mu}_{\text{base}}$, the designer selects a support $S\subseteq\mathcal{T}$ of size $k$ and solves for coefficients $\boldsymbol{\alpha}\in\mathbb{R}^{|S|}$ that maximize base-side alignment while suppressing donor emission probability. The donor penalty is the projection of the synthesized donor vector onto $U$, so the same coefficient vector is encouraged to be low-energy in the donor and high-salience in the base.
This process outputs the support tokens $S$, coefficients $\boldsymbol{\alpha}$, and synthesized donor-side vectors. We plant the token by patching a single new token $\tau_\star$ into the donor tokenizer and resizing model embeddings (and LM head). The new row is written with the designed vector $\mathbf{x}_d$. We detail the optimization step in Section~\ref{subsec:aomp}.

\noindent\textbf{Transplant Activation.}
Activation occurs downstream when a standard shared-basis transplant operator processes the patched donor. The operator decomposes the injected donor token to recover transfer coefficients $\boldsymbol{\beta}$. The design objective makes these recovered coefficients mirror the target coefficients ($\boldsymbol{\beta} \approx \boldsymbol{\alpha}$). Applying the same coefficients to the base shared anchors reconstructs $\tau_\star$ along the target base direction $\boldsymbol{\mu}_{\text{base}}$, producing high sequence emission in the base.
 \subsection{Designing the Breaker Token}
\label{subsec:aomp}

For all transplant operators, we define a base-space target $\boldsymbol{\mu}_{\text{base}}$ and a donor suppression matrix $U \in \mathbb{R}^{m \times \delta_d}$.
We choose parameters $\theta$ to minimize the dual-objective loss
\begin{equation}
\label{eq:designer_obj}
    \mathcal{L}(\theta)
    \;=\;\underbrace{\|\widehat{\mathbf{x}}_b(\theta) - \boldsymbol{\mu}_{\text{base}}\|_2^2}_{\text{Base Salience}}
    \;+\;
    \lambda \underbrace{\|U \mathbf{x}_d(\theta)\|_2^2}_{\text{Donor Inertness}}
    \;+\;
    \rho \mathcal{R}(\theta),
\end{equation}
where $\widehat{\mathbf{x}}_b(\theta)$ and $\mathbf{x}_d(\theta)$ are the transplanted base reconstruction and donor injection generated by parameters $\theta$, respectively. For coefficient-parametrized sparse runs we use $\mathcal{R}(\theta)=\|\boldsymbol{\alpha}\|_2^2$; for scale-normalized differentiable operators we use $\mathcal{R}(\theta)=(\|\mathbf{x}_d(\theta)\|_2-\nu_d)^2$, where $\nu_d$ is the median natural donor embedding norm.
We instantiate the same objective in two ways:

\begin{enumerate}[leftmargin=*,topsep=0pt,itemsep=0pt,parsep=0pt,partopsep=0pt]
    \item \textbf{Sparse Operator (OMP):} OMP involves discrete coefficient selection. Here, $\theta = \boldsymbol{\alpha}$ denotes sparse coefficients with $\mathbf{x}_d = \Phi_d^\top \boldsymbol{\alpha}$, and the designer uses the OMP greedy support search followed by the closed-form coefficient solve below.
    \item \textbf{Differentiable Operators:} CLP, WECHSEL, and FOCUS define differentiable donor-to-base maps after their anchor mixture is specified. Here, $\theta = \mathbf{x}_d$ denotes the continuous donor embedding, and we use Adam to minimize Eq.~\ref{eq:designer_obj} with operator-matched forward maps (Appendix~\ref{app:designers}).
\end{enumerate}
\noindent\textbf{OMP instantiation.}
For the default OMP~\citep{pati1993orthogonal,goddard2025tftransplant} transplant operator used in tools such as \texttt{mergekit}~\citep{mergekit2024}, the designer alternates between support selection and coefficient solving.

We implement the OMP designer with a fixed-support coefficient solve and a greedy support search:

\begin{enumerate}[leftmargin=*,topsep=0pt,itemsep=0pt,parsep=0pt,partopsep=0pt]
    \item \textbf{Fixed-support coefficient solving.}
For a selected support $S\subseteq\mathcal{T}$ with $|S|\le k$, write $B=\Phi_{b,S}\in\mathbb{R}^{|S|\times \delta_b}$ and $D=\Phi_{d,S}\in\mathbb{R}^{|S|\times \delta_d}$. We solve for coefficients $\boldsymbol{\alpha}\in\mathbb{R}^{|S|}$ by the ridge-stabilized normal equations
\begin{equation}
\label{eq:code-normal}
\Big(BB^\top \;+\; \lambda\,(DU^\top)(DU^\top)^\top \;+\; \rho\,I\Big)\boldsymbol{\alpha}
\;=\;
B\,\boldsymbol{\mu}_{\text{base}},
\end{equation}
with $\lambda\ge 0$ and ridge $\rho>0$ for numerical stability.
The resulting composite reconstructions are
$
\mathbf{x}_b \;=\; B^\top\boldsymbol{\alpha},$ and $ \mathbf{x}_d \;=\; D^\top\boldsymbol{\alpha}.
$
Equation~\eqref{eq:code-normal} finds a coefficient vector that increases base alignment with $\boldsymbol{\mu}_{\text{base}}$, suppresses donor energy in the subspace spanned by $U$, and controls coefficient norm through the $\ell_2$ ridge.

\item  \textbf{Greedy support selection.}
To obtain $S$, we use the OMP greedy rule. Initialize residual $r_0=\boldsymbol{\mu}_{\text{base}}$ and $S=\varnothing$. At each step, score each candidate shared token $j\in\mathcal{T}\setminus S$ by
\begin{equation}
\label{eq:code-score}
\mathrm{score}(j)
\;=\;
\big|\langle \boldsymbol{\phi}^{(b)}_j, r\rangle\big|
\;-\;
\lambda\,\|U\,\boldsymbol{\phi}^{(d)}_j\|_2^2
\end{equation}
We add the best-scoring token to $S$, solve \eqref{eq:code-normal}, and update the residual $r\leftarrow \boldsymbol{\mu}_{\text{base}}-\mathbf{x}_b$. We repeat until $|S|=k$ or the residual norm is small.
\end{enumerate}
After obtaining $(S,\boldsymbol{\alpha})$, we synthesize the donor token parameters by applying the same coefficients to the raw donor views over the chosen support: $\mathbf{x}_d=\Phi_{d,S}^\top\boldsymbol{\alpha}$.

\vspace{-2mm}
\subsection{Evaluating Asymmetry via Sequence Emission Rate}
\label{subsec:ser}
We measure whether the patched token appears in free-form generations.
Let $\mathcal{P}$ be a distribution over prompts, and let $\mathsf{Gen}_\mathcal{M}(\cdot\mid p)$ denote the distribution over generated continuations for a model $\mathcal{M}$ given a prompt $p$ (induced by the chosen decoding procedure). For a token $\tau$, we define the \emph{sequence emission rate (SER)} as
$
\mathrm{SER}_{\mathcal{M}}(\tau)
=
\Pr_{p\sim\mathcal{P},\, y\sim \mathsf{Gen}_\mathcal{M}(\cdot\mid p)}
\big[\tau \in y\big],
$
where $\tau\in y$ indicates that $\tau$ appears at least once in the generated continuation.
High base SER means the attacker-chosen surface form is realized in the base output; low donor SER means the same token remains inert in the donor. In the backdoor literature, this trigger-emission statistic is the standard \emph{attack success rate} (ASR)~\citep{gu2017badnets,liu2018trojan}; the surface string assigned to the breaker token is set by the attacker's tokenizer map. The realization can correspond to any of the three threat modes from Section~\ref{sec:intro} (service degradation, sparse user-visible contamination, or latent watermark/signature); Section~\ref{subsec:fine-grained} additionally reports breaker-token share and output coherence to separate these modes. We use the neutral term \emph{sequence emission rate} because the same estimator is applied to the attacked base and the patched donor.

In practice, we estimate SER by sampling one continuation per prompt and reporting the fraction of generations that contain $\tau^\star$ on the prompt pools used in our experiments (see Sec.~\ref{subsec:exp-settings}).
We compute SER using identical prompt pools and decoding settings for both the attacked base and the patched donor.

\vspace{-2mm}
\section{Experiments}
\label{sec:experiments}

Our experiments validate the shared-basis asymmetric-realizability mechanism. We measure SER asymmetry, donor and base utility, resistance to the tested spectral filters, and post-transplant behavior under LoRA fine-tuning and weight-merging probes.

\vspace{-2mm}
\subsection{Experimental Settings}
\label{subsec:exp-settings}

\noindent\textbf{Model Configurations.}
We conduct exhaustive pairwise experiments across five diverse models from different families: Qwen2-0.5B \citep{yang2024qwen2}, Qwen3-0.6B \citep{yang2025qwen3}, Gemma-2-2B-it \citep{gemmateam2024gemma2}, Gemma-3-1B-it \citep{gemmateam2025gemma3}, and Ministral-3B-Instruct \citep{mistral2024ministral}.
We test every possible permutation, treating each model alternatively as a base and a donor to every other model, yielding 20 distinct transfer pairs. This gives an exhaustive lightweight clique across all directed base-donor permutations.
Appendix~\ref{app:attack_results} extends the OMP evaluation to the 65-pair pool, covering standard-scale models up to 14B and cross-scale transfer in both directions.
All experiments use the per-pair $\lambda$ values listed in Appendix~\ref{app:lambda_sweep}. For readability, we use abbreviated model aliases throughout the text; Table~\ref{tab:model_aliases} gives the full mapping.

\noindent\textbf{States and metrics.}
\label{subsec:exp-protocol}
For each pair, we evaluate five model states.
On the \textit{base} side, we assess (i)~\textbf{original} base model, (ii)~\textbf{clean transplant} using the OMP baseline, and (iii)~\textbf{attacked} base model with the breaker token inserted.
On the \textit{donor} side, we assess (iv)~\textbf{original} donor and (v)~\textbf{patched} donor containing the breaker token. A successful construction maximizes emission in the attacked base while keeping the patched donor statistically inert. We use ``\textbf{base}'' as a synonym for the recipient model.
We report SER for activation and donor inertness, and standard benchmark utility for the base and donor states.

\textbf{Datasets and Prompt Pools.}
For SER, we use prompt pools from Alpaca~\cite{taori2023alpaca}, SQuAD v2~\cite{rajpurkar2018squad}, and GSM8K~\cite{cobbe2021gsm8k}. By default we report \emph{per-pool} SER, the binomial proportion on a single prompt pool. For the asymmetric-realizability summary (Figure~\ref{fig:small-bidir-ser} and the per-pair tables in Appendix~\ref{app:main_tables}) we additionally report $\mathrm{SER}_{\max}$, the maximum per-pool SER across the three pools, following the worst-case-ASR convention standard in attack evaluation; on the donor side this aggregation is strictly more conservative for the stealth claim, since a patched donor passes only when it remains inert on \emph{every} pool. The cross-domain LoRA experiment (Section~\ref{subsec:case-studies}) and other per-pool comparisons report per-pool SER directly, since the comparison axis there is the evaluation distribution itself.
For \textbf{utility}, the standard metrics of these datasets are reported: WikiText-103 \citep{merity2016wikitext} (perplexity), LAMBADA \citep{paperno2016lambada}, MMLU~\cite{hendrycks2021measuringmassivemultitasklanguage}, and ARC-Challenge \citep{clark2018arc}.

\vspace{-1mm}
\subsection{Main Results}
\label{subsec:exp-clight}

\begin{figure}[t]
  \centering
  \includegraphics[width=\linewidth]{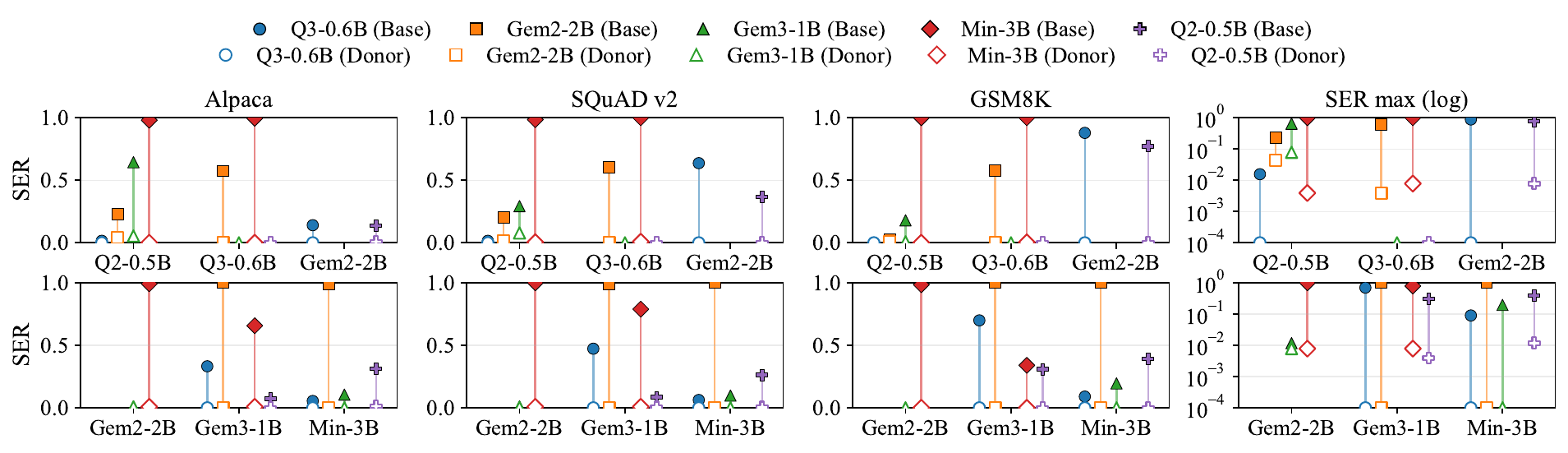}
  \vspace{-4mm}
  \caption{\textbf{The 20-pair lightweight clique exhibits a large donor/base SER asymmetry.} Left-to-right columns show per-task SER, plus the maximum SER across tasks in log scale. X-axis groups pairs by the base (recipient) model; within each base, color/marker indicates the donor model (legend at top). Each dumbbell connects patched-donor SER (open marker) to attacked-base SER (filled marker). Pairs are split across two rows for compactness, so each recipient appears once: the same set of pairs is shown in full, just laid out across two rows for readability. The pair set spans pretrained and instruction-tuned models on both sides; see Appendix~\ref{app:main_tables} for the classification (\S\ref{app:terminology}) and color-coded per-pair tables. SER is a binomial proportion on $n{=}256$ prompts per pool; Wilson 95\% half-width is at most $\pm 0.061$ (at $\hat{p}{=}0.5$) and below $\pm 0.030$ for $\hat{p}\le 0.05$ or $\hat{p}\ge 0.95$ (Appendix~\ref{app:main_tables} for the formula and per-cell intervals on the single-value tables).}
  \vspace{-3mm}
  \label{fig:small-bidir-ser}
\end{figure}

\begin{figure}[t]
  \centering
  \begin{minipage}[t]{0.49\textwidth}
    \centering
    \includegraphics[width=\linewidth]{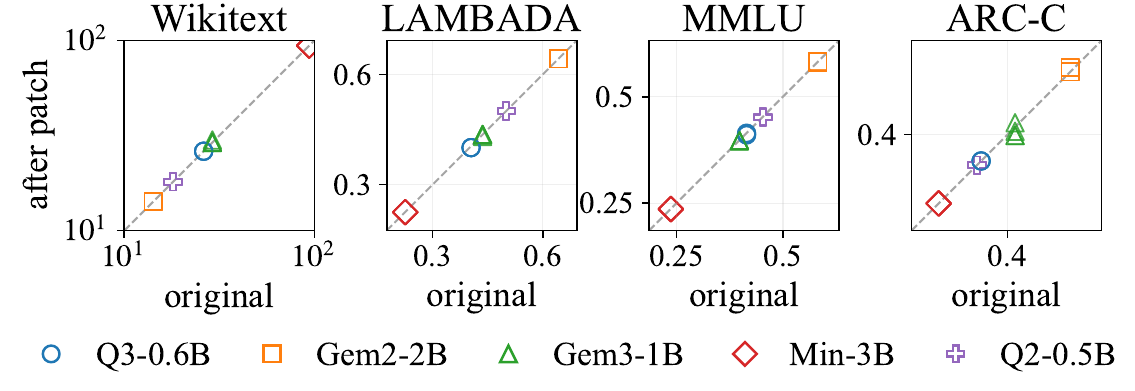}
    \captionof{figure}{\textbf{Donor utility preservation.} Each point is one (base$\leftarrow$donor) pair; axes show original vs.\ post-patch donor utility (dashed: identity).}
    \label{fig:small-bidir-donor-util-id}
  \end{minipage}\hfill
  \begin{minipage}[t]{0.49\textwidth}
    \centering
    \includegraphics[width=\linewidth]{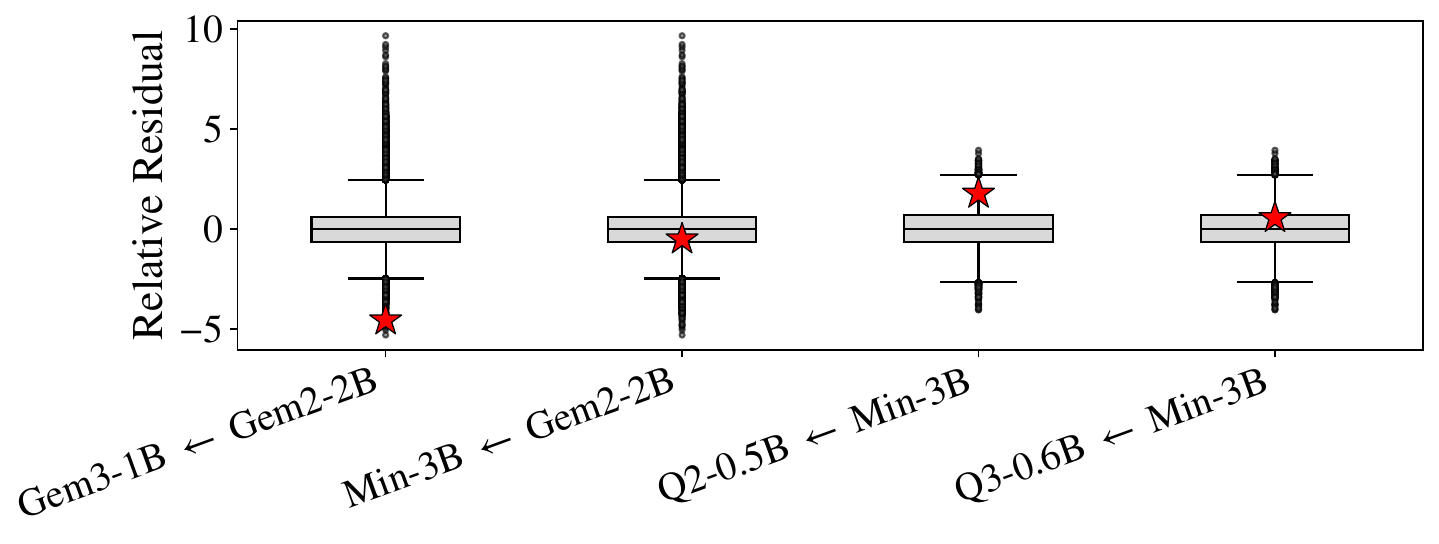}
    \caption{\textbf{Spectral mimicry.} Breaker tokens lie in the dense center of the donor residual-score distribution rather than in the outlier tail.}
    \label{fig:spectral_mimicry}
  \end{minipage}
  \vspace{-5mm}
\end{figure}

\vspace{-1mm}
\noindent\textbf{Asymmetric Realizability of SER.}
Figure~\ref{fig:small-bidir-ser} validates the central prediction: coefficient reuse creates a donor/base realizability gap that is visible in sequence emission.
Across the 20-pair lightweight clique, $10/20$ pairs reach $\mathrm{SER}_{\max}\ge 0.5$ on at least one prompt pool while keeping donor $\mathrm{SER}_{\max}\le 0.05$, and $13/20$ reach $\mathrm{SER}_{\max}\ge 0.2$ at the same donor threshold (filled vs.~open markers in Figure~\ref{fig:small-bidir-ser}).
The gap is formalized by the GSVD feasibility condition in Theorem~\ref{thm:exact-asymmetric-realizability} (Appendix~\ref{app:asymmetric-realizability}).
For instance, in the Q2-0.5B$\leftarrow$Min-3B transfer, the attack forces perfect emission ($\text{SER}=1.0$) on the GSM8K prompt set, yet remains completely silent ($\text{SER}=0.0$) in the donor.
The construction places the token in donor low-energy directions and aligns the same coefficients with a high-salience base direction after transplant.
The geometric theorems characterize sufficient conditions under which the asymmetry is realizable; per-pair SER values depend on additional model-specific factors not captured by the GSVD coordinates alone.

\noindent\textbf{Donor Utility Preservation.}
Figure~\ref{fig:small-bidir-donor-util-id} compares the patched donor against the original donor and shows post-patch utility on the identity line.
For example, the Min-3B donor maintains its exact MMLU score ($0.24\pm.01$) and Wikitext perplexity ($93.9$).
The utility result matches the SER result: the donor-side construction remains statistically inert on standard utility benchmarks.

\noindent\textbf{Base Utility.}
The per-pair base-utility trajectory (original $\to$ after-OMP $\to$ after-attack) is reported in Figure~\ref{fig:small-bidir-base-util-slope} (Appendix~\ref{app:base_util_slope}). Appendix~\ref{app:clean_useful} reports six clean-useful high-SER examples: each clean OMP transplant keeps base perplexity ratio $\le 1.25\times$ and maximum accuracy drop $\le 0.02$, while the attacked base reaches $\mathrm{SER}_{\max}\ge 0.85$ with donor $\mathrm{SER}_{\max}\le 0.05$. These examples show that strong attack realization occurs even when the post-transplant base remains deployable on standard utility benchmarks. Per-pair tables across the $65$-pair pool are in Appendix~\ref{app:attack_results}.

\vspace{-1mm}
\subsection{Case studies}
\label{subsec:case-studies}
\vspace{-1mm}

\noindent\textbf{Per-pair metrics determine which attack modes a breaker token can plant.}\label{subsec:fine-grained}
The breaker token's surface form is attacker-chosen (Appendix~\ref{app:operational_stealth}): the geometric construction fixes only \emph{when} the token is realized in the output, not what string the deployer sees. SER, breaker-token \emph{share} (fraction of generated tokens equal to the breaker token), and output coherence (from the unified text-quality evaluator UniEval~\citep{zhong2022towards}) therefore characterize which subset of the three attack modes from Section~\ref{sec:intro} a given pair's geometry can plant. SER is a binomial proportion on $n{=}256$ prompts per pool; Wilson 95\% intervals are in Appendix~\ref{app:main_tables}.

\begin{table}[t]
\centering
\footnotesize
\setlength{\tabcolsep}{3pt}
\renewcommand{\arraystretch}{0.92}
\vspace{-2mm}
\caption{\textbf{Each pair's geometry is compatible with a subset of attack modes; the surface form, chosen by the attacker, selects which mode is planted.} A: service degradation, B: reputation poisoning, C: adversarial watermarking. Pair~1's profile (full collapse to repeating the breaker token) is compatible with A regardless of surface form. Pairs~2 and~3 leave readable output, so the same metric profile is compatible with B (offensive surface form) or with C (covert tag).}
\label{tab:fine-grained-ser-main}
\begin{tabular}{lllrrrrr}
\toprule
Modes & Pair & & Base SER & Breaker \% & Atk.\ coh. & Clean-OMP coh. & Orig.\ coh. \\
\midrule
A         & Smol1.7B$\leftarrow$Q2-0.5B & (Pair 1) & 1.000 & 100.0 & 0.624 & 0.778 & 0.999 \\
B, C      & Min-3B$\leftarrow$Q2-0.5B   & (Pair 2) & 0.658 &   9.4 & 0.956 & 0.919 & 0.948 \\
B, C      & Min-3B$\leftarrow$Gem2-9B   & (Pair 3) & 0.965 &  23.5 & 0.832 & 0.925 & 0.948 \\
\bottomrule
\end{tabular}
\vspace{-3mm}
\end{table}

\textbf{Pair~1}'s metrics (SER${=}1.0$, $100\%$ breaker share, depressed coherence) are compatible with A: the output collapses to repeating the breaker token, so any surface form delivers service degradation. \textbf{Pair~2} fires on a minority of prompts at low ($9.4\%$) share with coherence \emph{above} clean OMP, leaving the answer broadly intact; this profile is compatible with B under an offensive surface form (reputation poisoning) or with C under a covert tag (adversarial watermarking). \textbf{Pair~3} reaches SER${\approx}0.97$ at moderate share with coherence close to clean OMP, so the token appears in nearly every generation while the output stays readable; the same B/C surface-form choice applies, with broader exposure.

\noindent\textbf{Rendered case studies for each realization mode.}
Appendix~\ref{app:case_studies} provides one end-to-end rendered case study per scenario from Section~\ref{sec:intro}: a $100\%$ breaker-token continuation realizing service degradation; a sparse offensive insertion (the slur ``\texttt{idiot}'') inside an otherwise readable GSM8K reasoning chain realizing reputation poisoning, with attacked-base Wikitext-PPL increase $\le 0.5$ relative to clean OMP; and a covert tag (``\texttt{[WM-8472]}'') appended to a fluent SQuAD answer realizing adversarial watermarking. Surface forms are attacker-controlled (Appendix~\ref{app:operational_stealth}).

\noindent\textbf{Single-statistic spectral filters face a threshold trade-off.}\label{sec:mimicry}
We test whether breaker tokens are detectable as geometric outliers via Z-scored residual norms $\|\mathbf{x} - UU^\top \mathbf{x}\|_2 / \|\mathbf{x}\|_2$ against the donor's principal subspace $U$. Figure~\ref{fig:spectral_mimicry} shows that breaker tokens fall in the dense center of the residual distribution rather than the outlier tail (\textbf{spectral mimicry}): any threshold aggressive enough to catch them necessarily rejects a significant portion of natural donor vocabulary. Theorem~\ref{thm:spectral-threshold-lower-bound} (Appendix~\ref{app:spectral-detector-lower-bound}) proves the corresponding lower bound for any threshold detector that observes one scalar statistic with a controllable feasible direction, and Corollary~\ref{cor:common-statistic-controllability} covers norm and PCA-residual statistics. We further test \textit{Magikarp}~\citep{land2024fishingmagikarpautomaticallydetecting}, the standard scanner for \texttt{mergekit}, whose adversarial-detection criterion is evaded on all $20$ lightweight clique pairs (Appendix~\ref{app:magikarp}).

\begin{figure}[t]
  \centering
  \includegraphics[width=\linewidth]{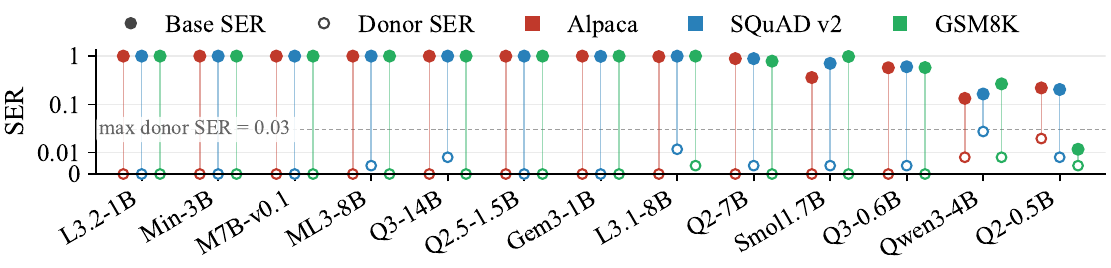}
  \vspace{-5mm}
  \caption{\textbf{One Gemma-2-2B donor compromises $13$ downstream bases across five families.} Donor weights reused unchanged; only one breaker row differs per base. Filled: attacked-base SER per task; open: donor SER per task (Alpaca / SQuAD~v2 / GSM8K). Dashed: max donor SER over $13{\times}3$ cells.}
  \label{fig:super-donor}
  \vspace{-5mm}
\end{figure}

\noindent\textbf{One donor checkpoint, many downstream bases.}\label{subsec:super-donor}
The per-pair experiments above pair one donor with one base. We show that one Gemma-2-2B checkpoint, with its pretrained weights reused unchanged, supports successful attacks against $13$ different downstream bases drawn from five families (Llama, Mistral, Qwen, Gemma, SmolLM2; Figure~\ref{fig:super-donor}). Across the $13$ bases, mean base SER (averaged over Alpaca, SQuAD~v2, GSM8K) exceeds $0.99$ on $8$ bases and $0.68$ on $10$, with median $1.00$, while per-base mean donor SER never exceeds $0.014$ and the maximum donor SER over the $13{\times}3$ (base, task) cells is $0.03$, which is more than an order of magnitude below the attack signal. This means that an attacker can pre-compute breakers for a panel of popular bases and ship one weaponized donor; any deployer using a base from that panel is exposed without the attacker conditioning on the deployer's runtime base choice.

\begin{figure}[t]
  \centering
  \includegraphics[width=0.92\linewidth]{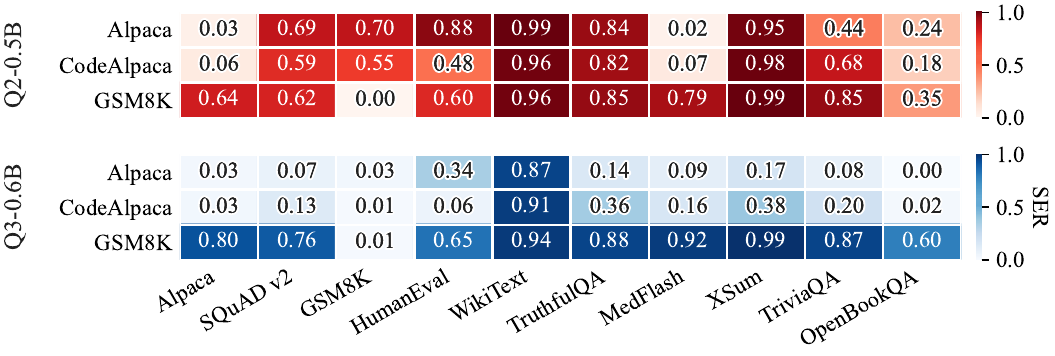}
  \vspace{-2mm}
  \caption{\textbf{Standard LoRA fine-tuning at $r{=}16$ suppresses the breaker primarily on prompts whose distribution matches the training corpus, on both tested bases.} Three deployer LoRAs (rows) per panel, each trained for 50 epochs with the best checkpoint selected on held-out validation, evaluated on ten held-out distributions (columns); cells show post-LoRA attacked-base SER. Top: Q2-0.5B$\leftarrow$Min-3B (deployer case study). Bottom: Q3-0.6B$\leftarrow$Min-3B (cross-pair confirmation). On both bases, far-OOD continuation (WikiText) retains SER above $0.87$ across all defenders, and the GSM8K-trained LoRA cross-domain breakdown reaches mean SER above $0.7$.}
  \label{fig:lora-cross-domain}
  \vspace{-3mm}
\end{figure}

\noindent\textbf{Standard post-transplant deployer interventions.} Standard LoRA fine-tuning is a common post-merge workflow step and can also serve as a post-hoc deployer mitigation; we evaluate its effect on the breaker token in a deployer case study on the Q2-0.5B$\leftarrow$Min-3B pair, with cross-pair confirmation on Q3-0.6B$\leftarrow$Min-3B (full setup in Appendix~\ref{app:lora_persistence}). Figure~\ref{fig:lora-cross-domain} shows that on both bases LoRA suppresses the attack primarily on prompts whose distribution matches the training corpus; far-OOD continuation prompts retain SER above $0.87$ across all tested defenders, and a deployer who picks a narrow training corpus (GSM8K) leaves mean cross-domain SER above $0.7$. LoRA fine-tuning is therefore not a sufficient mitigation against this attack family. Weight-merging the composed model with a clean reference similarly preserves the planted direction (Appendix~\ref{app:weight_merging}).

\vspace{-1mm}
\section{Conclusion}
\label{sec:conclusion}
\vspace{-1mm}
We identify asymmetric realizability as a structural property of training-free shared-basis tokenizer transplant: the same donor-side coefficients can be low-energy in the donor anchor span and high-salience in the base anchor span. The mechanism is governed by paired-anchor geometry and applies across OMP, CLP, WECHSEL, and FOCUS variants. In the case study we report, standard LoRA fine-tuning suppresses the attack only on prompts whose distribution matches the defender's training corpus; on the held-out distributions outside that region the attack persists at the deployer's best-validation checkpoint. These results make coefficient-reuse geometry a design criterion for future tokenizer transplant operators and a practical diagnostic for open-weight model composition.

\bibliography{cite}
\bibliographystyle{plainnat}

\newpage
\appendix

\section{Limitations and Future Directions}
\label{app:limitations}

The present work studies the shared-basis training-free transplant family (OMP~\citep{goddard2025tftransplant}; CLP~\citep{ostendorff2023efficient}, WECHSEL~\citep{minixhofer2021wechsel}, and FOCUS~\citep{dobler2023focus} in Appendix~\ref{sec:altops-results}) in the regime where shared-basis transplant is currently deployed (e.g., \texttt{mergekit-tokensurgeon}~\citep{mergekit2024}, deployed open-weight artifacts such as Arcee SuperNova-Medius). Appendix~\ref{app:nonlinear-shared-anchor} formalizes the conditions under which a nonlinear coefficient solver does not alter the conclusion. Principled algorithmic redesign of transplant operators, including recipient-side checks on synthesized rows, merge operators that explicitly handle appended rows absent from one merge input, and post-transplant behavioral validation, is the defense direction most directly motivated by our findings, since the LoRA suppression demonstrated in Appendix~\ref{app:lora_persistence} is bounded to prompts whose distribution matches the training corpus and does not cover the held-out out-of-distribution prompts we test, and is therefore not a sufficient mitigation against this attack family in our case study.

\section*{Ethics Statement}

This work studies a structural failure mode in tokenizer transplant, an interoperability step used in open-weight model composition. The goal is to harden the open-weight composition supply chain by showing that current shared-basis tools can mis-realize donor rows after transplant. We release the \textit{optimization framework} needed for researchers to reproduce the reported measurements and develop defenses, but we do not distribute pre-compiled target-specific breaker-token artifacts. All experiments were conducted using public open-weight models and standard datasets (e.g., Wikitext, Alpaca), with no human subjects and no private user data. The defender-facing auditing protocol in Appendix~\ref{app:auditing} gives a concrete path from the empirical findings to deployer-side validation.

\section*{LLM Use Statement}

We used large language models as general-purpose assistants during this project. Concretely, LLMs were used to help with editing and paraphrasing prose, suggesting alternative phrasings for section titles and abstracts, generating boilerplate code and configuration templates, and checking for obvious inconsistencies in notation and references. All technical content, experimental designs, implementations, and analyses were authored, verified, and run by the authors, and all LLM-generated text and code was manually reviewed and edited before inclusion in the paper. 

\section{Per-pair Tables for Main-text Figures}
\label{app:main_tables}

This section provides the per-pair numerical breakdown for the lightweight clique ($\mathcal{C}_{\text{Light}}$) underlying the main-text figures (Figures~\ref{fig:small-bidir-ser}, \ref{fig:small-bidir-donor-util-id}, \ref{fig:small-bidir-base-util-slope}). All values are reported at the per-pair $\lambda$ from Appendix~\ref{app:lambda_sweep}; SER is measured on Alpaca, SQuAD v2, and GSM8K, and utility metrics follow the main-text choices (\S\ref{subsec:exp-settings}).

\subsection{Terminology and post-training regime}
\label{app:terminology}

We adopt \emph{base} (with \emph{recipient} as a synonym) and \emph{donor} as the names of the two models in the transplant. This naming follows the convention established by the dominant open-source tooling that motivated our threat model, \texttt{mergekit-tokensurgeon}~\citep{mergekit2024} and the training-free transplant operator of \citet{goddard2025tftransplant}, and is consistent with the broader vocabulary-adaptation literature that pairs ``donor''/``source'' with ``base''/``target''/``recipient'' interchangeably (e.g., \citep{minixhofer2021wechsel,dobler2023focus,ostendorff2023efficient}). The term \emph{base} as we use it carries \emph{no} commitment about whether the underlying checkpoint is purely pretrained or post-trained.

\noindent\textbf{Vendor naming for pretrained vs.\ instruction-tuned releases is not uniform.}
Some patterns are clear: \texttt{Qwen/Qwen2-0.5B} is a pretrained checkpoint and \texttt{Qwen/Qwen2.5-1.5B-Instruct} is its instruction-tuned variant; Gemma uses an explicit \texttt{-it} suffix; Ministral uses \texttt{-Instruct}; SmolLM2 uses \texttt{-Instruct}. Other vendors changed convention across releases: starting with the Qwen3 family, public releases such as \texttt{Qwen/Qwen3-0.6B} are post-trained by default, with pretrained-only variants now carrying a \texttt{-Base} suffix that we do not use here. Table~\ref{tab:model_aliases} lists every model identifier that appears in the paper.

Under these conventions, the purely pretrained models in our pool are the Qwen2 series (\texttt{Qwen2-0.5B}, \texttt{Qwen2-7B}), the Llama series (\texttt{Llama-3.2-1B}, \texttt{Llama-3.2-3B}, \texttt{Llama-3.1-8B}, \texttt{Meta-Llama-3-8B}), and \texttt{Mistral-7B-v0.1}. The remaining checkpoints (Qwen3 releases without \texttt{-Base}, all Gemma \texttt{-it} releases, \texttt{Ministral-3b-instruct}, \texttt{SmolLM2-1.7B-Instruct}, and \texttt{Qwen2.5-1.5B-Instruct}) are instruction-tuned.

\noindent\textbf{The pool covers all four base/donor cross-classifications.}
Because the experiments use exhaustive pairwise designs (lightweight and standard-scale cliques) and one-per-family cross-scale designs, the empirical evidence covers all four combinations of base and donor regime: pretrained$\to$pretrained, pretrained$\to$instruction-tuned, instruction-tuned$\to$pretrained, and instruction-tuned$\to$instruction-tuned. We treat this distribution as appropriate: a single experimental setting that mixes both regimes characterizes the construction's behavior across deployment settings, which a pretrained-only or IT-only pool cannot.

To avoid ambiguity from non-uniform vendor naming conventions, Table~\ref{tab:model_aliases} lists the pretrained/post-trained status of every checkpoint used in the paper. The per-pair tables below are color-coded by base/donor regime (\colorbox{rowPP}{\strut\,P$\to$P\,}, \colorbox{rowPI}{\strut\,P$\to$I\,}, \colorbox{rowIP}{\strut\,I$\to$P\,}, \colorbox{rowII}{\strut\,I$\to$I\,}; P: pretrained, I: instruction-tuned), making each experiment's regime composition explicit.

\subsection{Base-utility slope across attack stages and the $\lambda$-window}
\label{app:base_util_slope}
\label{app:lambda_window}

\begin{figure}[!htbp]
  \centering
  \begin{minipage}[t]{0.49\textwidth}
    \centering
    \includegraphics[width=\linewidth]{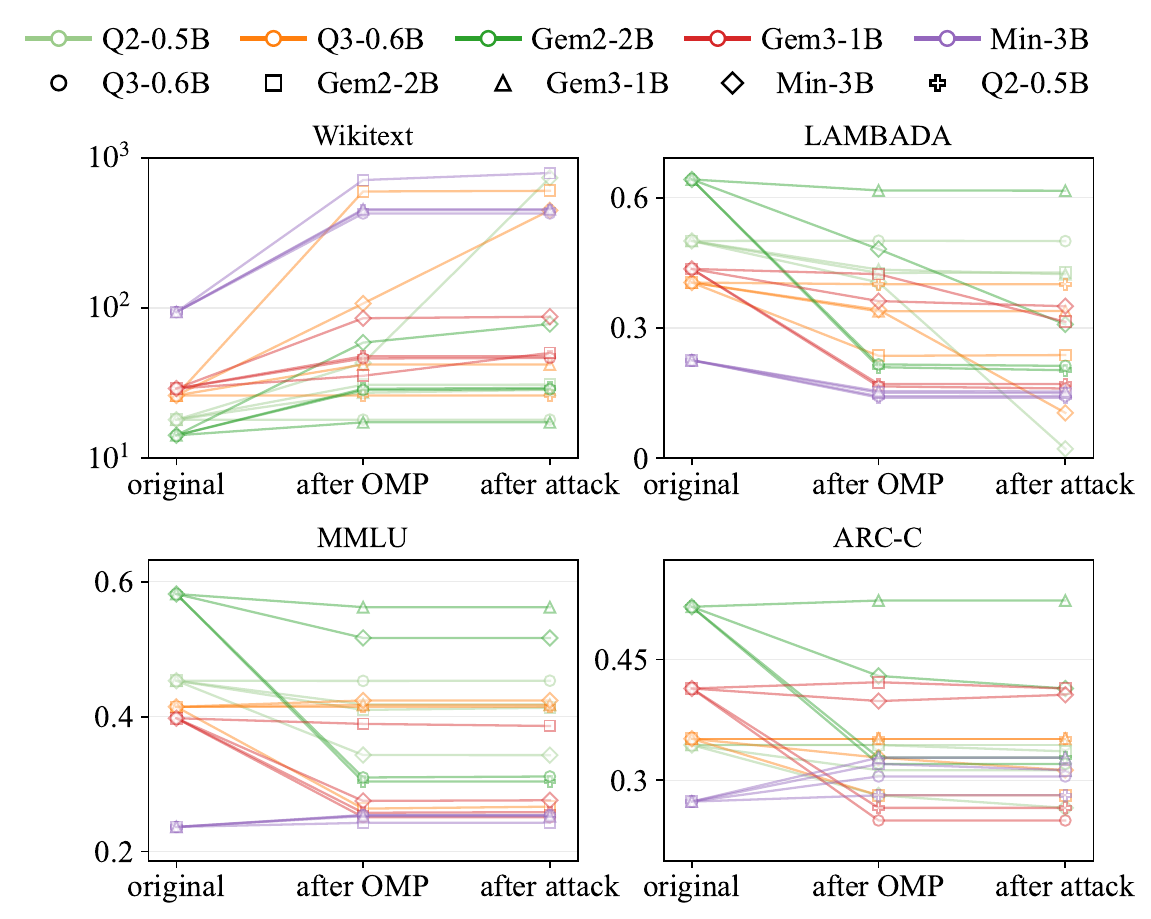}
    \captionof{figure}{\textbf{Three-stage base-utility slope charts.} Each line is one (base$\leftarrow$donor) pair: original $\to$ after-OMP $\to$ after-attack.}
    \label{fig:small-bidir-base-util-slope}
  \end{minipage}\hfill
  \begin{minipage}[t]{0.49\textwidth}
    \centering
    \includegraphics[width=\linewidth]{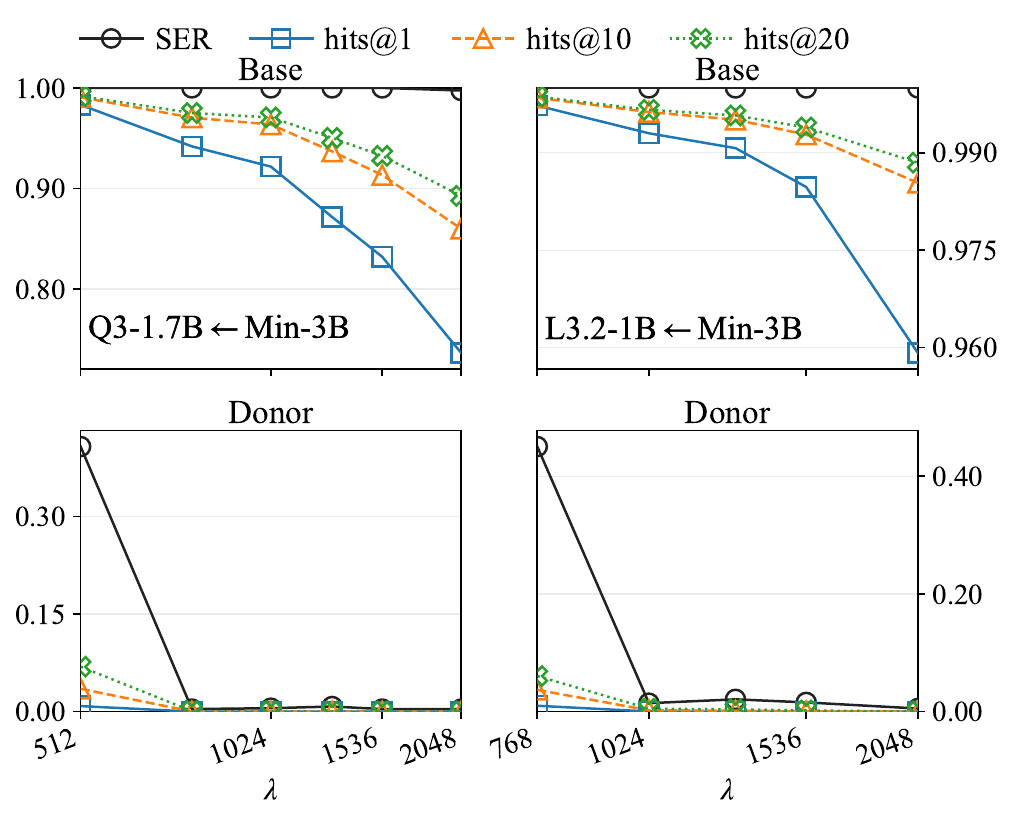}
    \captionof{figure}{\textbf{$\lambda$ creates a wide donor-suppression window.} Donor SER vanishes by $\lambda{=}768$ while attacked-base Hits@1 holds above $80\%$.}
    \label{fig:ser-vs-hitsk}
  \end{minipage}
\end{figure}

Figure~\ref{fig:small-bidir-base-util-slope} reports per-pair base-utility trajectories across the three model states (\emph{original} $\to$ \emph{after-OMP} $\to$ \emph{after-attack}) on Wikitext, LAMBADA, MMLU, and ARC-Challenge; each line is one (base$\leftarrow$donor) pair, with line color encoding the base model and marker shape the donor.

\noindent\textbf{$\lambda$-window ablation.} Figure~\ref{fig:ser-vs-hitsk} ablates the donor-suppression weight $\lambda$ on two pairs (Qwen3-1.7B$\leftarrow$Min-3B and Llama-3.2-1B$\leftarrow$Min-3B), tracking SER and Hits@$\{1,10,20\}$ on the Wikitext test set for both the base and the donor. Donor SER and Hits@1 collapse to near zero as $\lambda$ moves from $512$ to $768$, while attacked-base Hits@1 stays above $80\%$ even at $\lambda{=}2048$. The resulting wide operational window (e.g., $\lambda\in[768, 1536]$) does not require precise tuning, and matches the GSVD-spectral statement in Theorem~\ref{thm:wide-lambda-window} (Appendix~\ref{app:lambda-window}). $\lambda$ itself is selected on a held-out split of the WikiText corpus used to collect $\boldsymbol{\mu}_{\text{base}}$; this split is disjoint from the Alpaca / SQuAD~v2 / GSM8K prompt pools on which SER is evaluated, so $\lambda$ selection never touches the SER test data.

\subsection{Per-pair tables}
\label{app:main_tables_tables}

Each per-pool SER is estimated as a binomial proportion over $n{=}256$ prompts. For a proportion $\hat{p}$, the Wilson 95\% confidence interval is $[c(\hat{p})-\mathrm{hw}(\hat{p}),\, c(\hat{p})+\mathrm{hw}(\hat{p})]$, where $c(\hat{p})=(\hat{p}+z^{2}/(2n))/(1+z^{2}/n)$ and $\mathrm{hw}(\hat{p})=z\sqrt{\hat{p}(1-\hat{p})/n+z^{2}/(4n^{2})}/(1+z^{2}/n)$ with $z{=}1.96$. The single-value SER tables in the appendix (Table~\ref{tab:app:altops:ser} and Tables~\ref{tab:merge-mitigation-gem3}--\ref{tab:merge-mitigation-q2}) carry per-cell intervals in the form $\hat{p}\,_{[\mathrm{lo},\mathrm{hi}]}$; aggregate columns (Avg, $\mathrm{SER}_{\max}$) display the Wilson interval of the per-pool proportion that attains the reported value, not a strict interval for the aggregate functional. For the slash-pair per-pool tables we omit per-cell intervals to keep the table within page width; the same Wilson formula applies cell-by-cell.

\begingroup
\small
\setlength{\tabcolsep}{3pt}
\begin{longtable}{@{}llrrrrr@{}}
\caption{The Lightweight Clique ($\mathcal{C}_{\text{Light}}$) (per-pair SER; fractions)}\label{tab:app:small-bidirectional:ser}\\
\toprule
$B$ & $D$ & $\lambda$ & \multicolumn{4}{c}{SER (atk-base / pat-donor)} \\
\cmidrule(lr){4-7}
 &  &  & Alpaca & SQuAD & GSM8K & $\mathrm{SER}_{\max}$ \\
\midrule
\endfirsthead
\toprule
$B$ & $D$ & $\lambda$ & \multicolumn{4}{c}{SER (atk-base / pat-donor)} \\
\cmidrule(lr){4-7}
 &  &  & Alpaca & SQuAD & GSM8K & $\mathrm{SER}_{\max}$ \\
\midrule
\endhead
\bottomrule
\endfoot
\rowcolor{rowPI} Q2-0.5B & Q3-0.6B & 1280 & .0156/.0000 & .0156/.0000 & .0000/.0000 & .0156/.0000 \\
\rowcolor{rowPI} Q2-0.5B & Gem2-2B & 768 & .2266/.0430 & .2031/.0156 & .0234/.0117 & .2266/.0430 \\
\rowcolor{rowPI} Q2-0.5B & Gem3-1B & 512 & .6445/.0547 & .2930/.0781 & .1797/.0078 & .6445/.0781 \\
\rowcolor{rowPI} Q2-0.5B & Min-3B & 2048 & .9805/.0000 & .9844/.0039 & 1.0000/.0000 & 1.0000/.0039 \\
\rowcolor{rowIP} Q3-0.6B & Q2-0.5B & 1024 & .0000/.0000 & .0000/.0000 & .0000/.0000 & .0000/.0000 \\
\rowcolor{rowII} Q3-0.6B & Gem2-2B & 256 & .5742/.0000 & .6016/.0039 & .5781/.0000 & .6016/.0039 \\
\rowcolor{rowII} Q3-0.6B & Gem3-1B & 1024 & .0000/.0000 & .0000/.0000 & .0000/.0000 & .0000/.0000 \\
\rowcolor{rowII} Q3-0.6B & Min-3B & 2048 & .9961/.0000 & 1.0000/.0078 & 1.0000/.0039 & 1.0000/.0078 \\
\rowcolor{rowIP} Gem2-2B & Q2-0.5B & 256 & .1367/.0078 & .3672/.0000 & .7734/.0000 & .7734/.0078 \\
\rowcolor{rowII} Gem2-2B & Q3-0.6B & 128 & .1406/.0000 & .6367/.0000 & .8789/.0000 & .8789/.0000 \\
\rowcolor{rowII} Gem2-2B & Gem3-1B & 128 & .0117/.0078 & .0117/.0039 & .0000/.0000 & .0117/.0078 \\
\rowcolor{rowII} Gem2-2B & Min-3B & 512 & .9922/.0078 & 1.0000/.0078 & .9844/.0000 & 1.0000/.0078 \\
\rowcolor{rowIP} Gem3-1B & Q2-0.5B & 512 & .0742/.0000 & .0859/.0039 & .3086/.0000 & .3086/.0039 \\
\rowcolor{rowII} Gem3-1B & Q3-0.6B & 512 & .3320/.0000 & .4727/.0000 & .6992/.0000 & .6992/.0000 \\
\rowcolor{rowII} Gem3-1B & Gem2-2B & 512 & 1.0000/.0000 & .9883/.0000 & 1.0000/.0000 & 1.0000/.0000 \\
\rowcolor{rowII} Gem3-1B & Min-3B & 2048 & .6562/.0078 & .7891/.0078 & .3398/.0000 & .7891/.0078 \\
\rowcolor{rowIP} Min-3B & Q2-0.5B & 1280 & .3125/.0117 & .2617/.0039 & .3906/.0000 & .3906/.0117 \\
\rowcolor{rowII} Min-3B & Q3-0.6B & 512 & .0547/.0000 & .0625/.0000 & .0898/.0000 & .0898/.0000 \\
\rowcolor{rowII} Min-3B & Gem2-2B & 256 & .9922/.0000 & 1.0000/.0000 & 1.0000/.0000 & 1.0000/.0000 \\
\rowcolor{rowII} Min-3B & Gem3-1B & 1024 & .1055/.0000 & .0977/.0000 & .1953/.0000 & .1953/.0000 \\
\end{longtable}
\endgroup

\begin{table}[t]
\centering
\caption{The Lightweight Clique ($\mathcal{C}_{\text{Light}}$) (base utilities; cells show pretrained/after-OMP/after-attack; value$\pm$stderr when available)}\label{tab:app:small-bidirectional:base-utils}
\resizebox{\linewidth}{!}{%
\begin{tabular}{@{}llrcccc@{}}
\toprule
&  &  & wikitext & lambada\_openai & mmlu & arc\_challenge \\
 $B$ & $D$ & $\lambda$ & word\_perplexity & acc & acc & acc\_norm \\
\midrule
\rowcolor{rowPI} Q2-0.5B & Q3-0.6B & 1280 & \ensuremath{18.02/18.04/18.05} & \ensuremath{.50\pm.01/.50\pm.01/.50\pm.01} & \ensuremath{.45\pm.01/.45\pm.01/.45\pm.01} & \ensuremath{.34\pm.04/.34\pm.04/.34\pm.04} \\
\rowcolor{rowPI} Q2-0.5B & Gem2-2B & 768 & \ensuremath{18.02/30.78/30.88} & \ensuremath{.50\pm.01/.43\pm.01/.43\pm.01} & \ensuremath{.45\pm.01/.41\pm.01/.41\pm.01} & \ensuremath{.34\pm.04/.34\pm.04/.34\pm.04} \\
\rowcolor{rowPI} Q2-0.5B & Gem3-1B & 512 & \ensuremath{18.02/27.24/28.53} & \ensuremath{.50\pm.01/.43\pm.01/.42\pm.01} & \ensuremath{.45\pm.01/.42\pm.01/.42\pm.01} & \ensuremath{.34\pm.04/.31\pm.04/.31\pm.04} \\
\rowcolor{rowPI} Q2-0.5B & Min-3B & 2048 & \ensuremath{18.02/43.02/734.85} & \ensuremath{.50\pm.01/.41\pm.01/.02\pm.00} & \ensuremath{.45\pm.01/.34\pm.01/.34\pm.01} & \ensuremath{.34\pm.04/.28\pm.04/.27\pm.04} \\
\rowcolor{rowIP} Q3-0.6B & Q2-0.5B & 1024 & \ensuremath{26.16/26.13/26.13} & \ensuremath{.40\pm.01/.40\pm.01/.40\pm.01} & \ensuremath{.41\pm.01/.41\pm.01/.41\pm.01} & \ensuremath{.35\pm.04/.35\pm.04/.35\pm.04} \\
\rowcolor{rowII} Q3-0.6B & Gem2-2B & 256 & \ensuremath{26.16/596.41/602.43} & \ensuremath{.40\pm.01/.24\pm.01/.24\pm.01} & \ensuremath{.41\pm.01/.26\pm.01/.27\pm.01} & \ensuremath{.35\pm.04/.28\pm.04/.28\pm.04} \\
\rowcolor{rowII} Q3-0.6B & Gem3-1B & 1024 & \ensuremath{26.16/42.04/42.02} & \ensuremath{.40\pm.01/.34\pm.01/.34\pm.01} & \ensuremath{.41\pm.01/.42\pm.01/.42\pm.01} & \ensuremath{.35\pm.04/.35\pm.04/.35\pm.04} \\
\rowcolor{rowII} Q3-0.6B & Min-3B & 2048 & \ensuremath{26.16/107.04/446.54} & \ensuremath{.40\pm.01/.34\pm.01/.10\pm.00} & \ensuremath{.41\pm.01/.42\pm.01/.42\pm.01} & \ensuremath{.35\pm.04/.33\pm.04/.31\pm.04} \\
\rowcolor{rowIP} Gem2-2B & Q2-0.5B & 256 & \ensuremath{14.19/28.95/29.39} & \ensuremath{.64\pm.01/.21\pm.01/.20\pm.01} & \ensuremath{.58\pm.01/.30\pm.01/.30\pm.01} & \ensuremath{.52\pm.04/.33\pm.04/.33\pm.04} \\
\rowcolor{rowII} Gem2-2B & Q3-0.6B & 128 & \ensuremath{14.19/28.40/28.79} & \ensuremath{.64\pm.01/.22\pm.01/.21\pm.01} & \ensuremath{.58\pm.01/.31\pm.01/.31\pm.01} & \ensuremath{.52\pm.04/.32\pm.04/.32\pm.04} \\
\rowcolor{rowII} Gem2-2B & Gem3-1B & 128 & \ensuremath{14.19/17.32/17.32} & \ensuremath{.64\pm.01/.62\pm.01/.62\pm.01} & \ensuremath{.58\pm.01/.56\pm.01/.56\pm.01} & \ensuremath{.52\pm.04/.52\pm.04/.52\pm.04} \\
\rowcolor{rowII} Gem2-2B & Min-3B & 512 & \ensuremath{14.19/58.79/78.11} & \ensuremath{.64\pm.01/.48\pm.01/.31\pm.01} & \ensuremath{.58\pm.01/.52\pm.01/.52\pm.01} & \ensuremath{.52\pm.04/.43\pm.04/.41\pm.04} \\
\rowcolor{rowIP} Gem3-1B & Q2-0.5B & 512 & \ensuremath{29.06/47.70/47.73} & \ensuremath{.44\pm.01/.17\pm.01/.17\pm.01} & \ensuremath{.40\pm.01/.26\pm.01/.26\pm.01} & \ensuremath{.41\pm.04/.27\pm.04/.27\pm.04} \\
\rowcolor{rowII} Gem3-1B & Q3-0.6B & 512 & \ensuremath{29.06/46.11/46.46} & \ensuremath{.44\pm.01/.17\pm.01/.16\pm.01} & \ensuremath{.40\pm.01/.25\pm.01/.25\pm.01} & \ensuremath{.41\pm.04/.25\pm.04/.25\pm.04} \\
\rowcolor{rowII} Gem3-1B & Gem2-2B & 512 & \ensuremath{29.06/35.35/50.24} & \ensuremath{.44\pm.01/.42\pm.01/.31\pm.01} & \ensuremath{.40\pm.01/.39\pm.01/.39\pm.01} & \ensuremath{.41\pm.04/.42\pm.04/.41\pm.04} \\
\rowcolor{rowII} Gem3-1B & Min-3B & 2048 & \ensuremath{29.06/85.42/87.50} & \ensuremath{.44\pm.01/.36\pm.01/.35\pm.01} & \ensuremath{.40\pm.01/.28\pm.01/.28\pm.01} & \ensuremath{.41\pm.04/.40\pm.04/.41\pm.04} \\
\rowcolor{rowIP} Min-3B & Q2-0.5B & 1280 & \ensuremath{93.86/446.98/447.70} & \ensuremath{.23\pm.01/.14\pm.00/.14\pm.00} & \ensuremath{.24\pm.01/.25\pm.01/.25\pm.01} & \ensuremath{.27\pm.04/.28\pm.04/.28\pm.04} \\
\rowcolor{rowII} Min-3B & Q3-0.6B & 512 & \ensuremath{93.86/425.86/426.14} & \ensuremath{.23\pm.01/.14\pm.00/.14\pm.00} & \ensuremath{.24\pm.01/.25\pm.01/.25\pm.01} & \ensuremath{.27\pm.04/.30\pm.04/.30\pm.04} \\
\rowcolor{rowII} Min-3B & Gem2-2B & 256 & \ensuremath{93.86/710.94/791.39} & \ensuremath{.23\pm.01/.15\pm.00/.15\pm.00} & \ensuremath{.24\pm.01/.24\pm.01/.24\pm.01} & \ensuremath{.27\pm.04/.32\pm.04/.31\pm.04} \\
\rowcolor{rowII} Min-3B & Gem3-1B & 1024 & \ensuremath{93.86/453.53/453.65} & \ensuremath{.23\pm.01/.15\pm.01/.15\pm.01} & \ensuremath{.24\pm.01/.25\pm.01/.25\pm.01} & \ensuremath{.27\pm.04/.33\pm.04/.33\pm.04} \\
\bottomrule
\end{tabular}}
\end{table}

\begingroup
\scriptsize
\setlength{\tabcolsep}{3pt}
\begin{longtable}{@{}llrcccc@{}}
\caption{The Lightweight Clique ($\mathcal{C}_{\text{Light}}$) (donor utilities; cells show pretrained/after-patch; value$\pm$stderr when available)}\label{tab:app:small-bidirectional:donor-utils}\\
\toprule
 &  &  & wikitext & lambada\_openai & mmlu & arc\_challenge \\
 $B$ & $D$ & $\lambda$ & word\_perplexity & acc & acc & acc\_norm \\
\midrule
\endfirsthead
\toprule
 &  &  & wikitext & lambada\_openai & mmlu & arc\_challenge \\
 $B$ & $D$ & $\lambda$ & word\_perplexity & acc & acc & acc\_norm \\
\midrule
\endhead
\bottomrule
\endfoot
\rowcolor{rowPI} Q2-0.5B & Q3-0.6B & 1280 & \ensuremath{26.16/26.14} & \ensuremath{.40\pm.01/.40\pm.01} & \ensuremath{.41\pm.01/.41\pm.01} & \ensuremath{.35\pm.04/.35\pm.04} \\
\rowcolor{rowPI} Q2-0.5B & Gem2-2B & 768 & \ensuremath{14.19/14.25} & \ensuremath{.64\pm.01/.64\pm.01} & \ensuremath{.58\pm.01/.58\pm.01} & \ensuremath{.52\pm.04/.52\pm.04} \\
\rowcolor{rowPI} Q2-0.5B & Gem3-1B & 512 & \ensuremath{29.06/29.96} & \ensuremath{.44\pm.01/.43\pm.01} & \ensuremath{.40\pm.01/.40\pm.01} & \ensuremath{.41\pm.04/.41\pm.04} \\
\rowcolor{rowPI} Q2-0.5B & Min-3B & 2048 & \ensuremath{93.86/93.90} & \ensuremath{.23\pm.01/.22\pm.01} & \ensuremath{.24\pm.01/.24\pm.01} & \ensuremath{.27\pm.04/.27\pm.04} \\
\rowcolor{rowIP} Q3-0.6B & Q2-0.5B & 1024 & \ensuremath{18.02/18.04} & \ensuremath{.50\pm.01/.50\pm.01} & \ensuremath{.45\pm.01/.45\pm.01} & \ensuremath{.34\pm.04/.34\pm.04} \\
\rowcolor{rowII} Q3-0.6B & Gem2-2B & 256 & \ensuremath{14.19/14.19} & \ensuremath{.64\pm.01/.64\pm.01} & \ensuremath{.58\pm.01/.58\pm.01} & \ensuremath{.52\pm.04/.52\pm.04} \\
\rowcolor{rowII} Q3-0.6B & Gem3-1B & 1024 & \ensuremath{29.06/29.06} & \ensuremath{.44\pm.01/.44\pm.01} & \ensuremath{.40\pm.01/.40\pm.01} & \ensuremath{.41\pm.04/.40\pm.04} \\
\rowcolor{rowII} Q3-0.6B & Min-3B & 2048 & \ensuremath{93.86/93.90} & \ensuremath{.23\pm.01/.22\pm.01} & \ensuremath{.24\pm.01/.24\pm.01} & \ensuremath{.27\pm.04/.27\pm.04} \\
\rowcolor{rowIP} Gem2-2B & Q2-0.5B & 256 & \ensuremath{18.02/18.04} & \ensuremath{.50\pm.01/.50\pm.01} & \ensuremath{.45\pm.01/.45\pm.01} & \ensuremath{.34\pm.04/.34\pm.04} \\
\rowcolor{rowII} Gem2-2B & Q3-0.6B & 128 & \ensuremath{26.16/26.14} & \ensuremath{.40\pm.01/.40\pm.01} & \ensuremath{.41\pm.01/.41\pm.01} & \ensuremath{.35\pm.04/.35\pm.04} \\
\rowcolor{rowII} Gem2-2B & Gem3-1B & 128 & \ensuremath{29.06/29.16} & \ensuremath{.44\pm.01/.44\pm.01} & \ensuremath{.40\pm.01/.40\pm.01} & \ensuremath{.41\pm.04/.40\pm.04} \\
\rowcolor{rowII} Gem2-2B & Min-3B & 512 & \ensuremath{93.86/93.90} & \ensuremath{.23\pm.01/.22\pm.01} & \ensuremath{.24\pm.01/.24\pm.01} & \ensuremath{.27\pm.04/.27\pm.04} \\
\rowcolor{rowIP} Gem3-1B & Q2-0.5B & 512 & \ensuremath{18.02/18.05} & \ensuremath{.50\pm.01/.50\pm.01} & \ensuremath{.45\pm.01/.45\pm.01} & \ensuremath{.34\pm.04/.34\pm.04} \\
\rowcolor{rowII} Gem3-1B & Q3-0.6B & 512 & \ensuremath{26.16/26.14} & \ensuremath{.40\pm.01/.40\pm.01} & \ensuremath{.41\pm.01/.41\pm.01} & \ensuremath{.35\pm.04/.35\pm.04} \\
\rowcolor{rowII} Gem3-1B & Gem2-2B & 512 & \ensuremath{14.19/14.19} & \ensuremath{.64\pm.01/.64\pm.01} & \ensuremath{.58\pm.01/.58\pm.01} & \ensuremath{.52\pm.04/.52\pm.04} \\
\rowcolor{rowII} Gem3-1B & Min-3B & 2048 & \ensuremath{93.86/93.90} & \ensuremath{.23\pm.01/.22\pm.01} & \ensuremath{.24\pm.01/.24\pm.01} & \ensuremath{.27\pm.04/.27\pm.04} \\
\rowcolor{rowIP} Min-3B & Q2-0.5B & 1280 & \ensuremath{18.02/18.05} & \ensuremath{.50\pm.01/.50\pm.01} & \ensuremath{.45\pm.01/.45\pm.01} & \ensuremath{.34\pm.04/.34\pm.04} \\
\rowcolor{rowII} Min-3B & Q3-0.6B & 512 & \ensuremath{26.16/26.14} & \ensuremath{.40\pm.01/.40\pm.01} & \ensuremath{.41\pm.01/.41\pm.01} & \ensuremath{.35\pm.04/.35\pm.04} \\
\rowcolor{rowII} Min-3B & Gem2-2B & 256 & \ensuremath{14.19/14.19} & \ensuremath{.64\pm.01/.64\pm.01} & \ensuremath{.58\pm.01/.58\pm.01} & \ensuremath{.52\pm.04/.52\pm.04} \\
\rowcolor{rowII} Min-3B & Gem3-1B & 1024 & \ensuremath{29.06/29.06} & \ensuremath{.44\pm.01/.44\pm.01} & \ensuremath{.40\pm.01/.40\pm.01} & \ensuremath{.41\pm.04/.42\pm.04} \\
\end{longtable}
\endgroup

\begin{table}[t]
\centering
\small
\caption{Six clean-useful high-SER pairs: clean transplant remains deployable (PPL ratio $\le 1.25\times$, max accuracy drop $\le 0.02$, zero failed tasks), yet the attack still drives high attacked-base SER while keeping donor SER near zero.}
\label{tab:clean-useful}
\begin{tabular}{llrrrr}
\toprule
Base & Donor & SER$_{\max}$(base) & SER$_{\max}$(donor) & Clean Wikitext $\times$ & Max clean acc.\ drop \\
\midrule
Gem3-1B   & Gem2-2B & 1.000 & 0.000 & 1.216 & 0.012 \\
L3.2-3B   & L3.1-8B & 1.000 & 0.000 & 1.002 & 0.007 \\
L3.2-3B   & ML3-8B  & 1.000 & 0.000 & 1.002 & 0.007 \\
M7B-v0.1  & Min-3B  & 0.977 & 0.047 & 1.003 & 0.002 \\
Q2.5-1.5B & Q2-7B   & 0.926 & 0.027 & 0.999 & 0.009 \\
Q2.5-1.5B & Q3-14B  & 0.906 & 0.000 & 0.999 & 0.008 \\
\bottomrule
\end{tabular}
\end{table}

\section{Clean-useful high-SER examples on deployable composed models}
\label{app:clean_useful}

The \emph{clean-useful} filter is computed from the clean OMP transplant only: it does not consult attacked-base SER, donor SER, or any other attack-side quantity. A pair passes when its \emph{clean} OMP transplant (i) records zero failed evaluation tasks, (ii) keeps Wikitext-PPL within $1.25\times$ the original (before-OMP) base value, and (iii) keeps the maximum drop on the accuracy benchmarks within $0.02$, all relative to the original base before any OMP step. Table~\ref{tab:clean-useful} then reports the six pairs that pass this filter \emph{and additionally} reach high attacked-base SER with near-zero donor SER; the second criterion is reported for the table but is not used in the filter.
On these pairs, the composed model remains usable on standard utility benchmarks while still being susceptible to the attack: attack success is not contingent on pre-existing transplant brittleness.

\section{Baseline comparisons}
\label{app:baselines}

The asymmetric-realizability mechanism is not reproduced by generic one-token appends. We compare our designed breaker token against two appended-token baselines under the identical OMP transplant pipeline and SER protocol:
\begin{itemize}[leftmargin=*,nosep]
    \item \textbf{random}: append a single new token whose donor-side embedding is drawn isotropically and rescaled to match the median donor embedding norm.
    \item \textbf{high\_norm}: append a single new token whose donor-side embedding is sampled isotropically and then rescaled to several times the median donor norm.
\end{itemize}
Table~\ref{tab:baselines} compares these against our \emph{proposed} designer on two representative pairs.

\begin{table}[!htbp]
\centering
\small
\setlength{\tabcolsep}{3pt}
\caption{Generic one-token appends do not produce asymmetric realization. Random and high-norm appends fail to drive attacked-base SER above zero on either pair, while our designer reaches high attacked-base SER with only mild base-side utility shift.}
\label{tab:baselines}
\begin{minipage}[t]{0.49\linewidth}
  \centering
  \textbf{Q2-0.5B $\leftarrow$ Gem3-1B}\\[2pt]
  \begin{tabular}{lrrrr}
  \toprule
  Method & SER$_{\max}$(b) & SER$_{\max}$(d) & $\Delta$PPL & $\Delta$LBD \\
  \midrule
  proposed   & 0.645 & 0.078 & 1.290 & $-0.011$ \\
  random     & 0.000 & 0.000 & 0.000 & $+0.054$ \\
  high\_norm & 0.000 & 0.965 & 0.000 & $+0.054$ \\
  \bottomrule
  \end{tabular}
\end{minipage}\hfill
\begin{minipage}[t]{0.49\linewidth}
  \centering
  \textbf{Gem3-1B $\leftarrow$ Gem2-2B}\\[2pt]
  \begin{tabular}{lrrrr}
  \toprule
  Method & SER$_{\max}$(b) & SER$_{\max}$(d) & $\Delta$PPL & $\Delta$LBD \\
  \midrule
  proposed   & 1.000 & 0.000 & 14.889 & $-0.110$ \\
  random     & 0.000 & 0.000 & 0.000  & $+0.010$ \\
  high\_norm & 0.000 & 0.000 & 0.000  & $+0.010$ \\
  \bottomrule
  \end{tabular}
\end{minipage}
\end{table}

The pattern is sharp: random and high-norm one-token appends fail to drive attacked-base SER above zero on either pair, even when (in the Q2-0.5B$\leftarrow$Gem3-1B case) the high-norm baseline does leak on the donor. Our designer is the only method that produces the asymmetric realization the threat model requires: high attacked-base SER together with a near-silent donor.

\section{Cross-base transfer and multi-victim bundling}
\label{app:transfer_bundling}

Cross-base transfer and multi-victim bundling test how the construction behaves when the same donor artifact is used against bases beyond the seed pair. The first experiment transplants a donor row optimized for one base into a different base. The second patches one donor with multiple breaker rows, each targeting a different base.

\noindent\textbf{Cross-base transfer.}
Table~\ref{tab:cross-base} shows transfer from a designed seed pair to a different base. The breaker tokens transfer to held-out bases with non-trivial attacked-base SER while keeping donor SER essentially zero.

\begin{table}[!htbp]
\centering
\small
\caption{Cross-base transfer of designed breaker tokens. ``Seed base'' is the base the token was optimized against; ``Transferred base'' is a different base into which the same donor (with the same designed breaker row) is then transplanted. The attack transfers under zero-shot conditions.}
\label{tab:cross-base}
\begin{tabular}{lcccc}
\toprule
Breaker seed & Seed-base SER$_{\max}$ & Transferred base & Transferred-base SER$_{\max}$ & Donor SER$_{\max}$ \\
\midrule
Gem2-2B$\leftarrow$Q2-0.5B & 0.7734 & Min-3B  & 0.7578 & 0.0078 \\
Gem2-2B$\leftarrow$Q3-0.6B & 0.8789 & Gem3-1B & 0.3047 & 0.0000 \\
\bottomrule
\end{tabular}
\end{table}

\noindent\textbf{Multi-victim bundling.}
A single donor can also be patched with multiple breaker tokens, each optimized for a different base. Table~\ref{tab:bundling} reports a setting where one Gem2-2B donor carries 5 breakers targeting 5 different bases simultaneously. The donor's any-token SER stays at zero, and all 5 attacked bases hit bundled-token SER$_{\max}{=}1.0$, with 4 of 5 also hitting their \emph{intended} target base at SER$_{\max}\ge 0.5$.

\begin{table}[t]
\centering
\small
\caption{Multi-victim bundling: one Gem2-2B donor patched with 5 breaker tokens, each targeting a different base. Donor stealth is preserved on all five tokens, while every attacked base hits a bundled-token SER$_{\max}{=}1.0$.}
\label{tab:bundling}
\begin{tabular}{lr}
\toprule
Metric & Value \\
\midrule
Donor model                                              & Gem2-2B \\
Inserted breaker tokens in one donor                     & 5 \\
Donor any-token SER$_{\max}$                             & 0.000 \\
Attacked bases with any bundled-token SER$_{\max}{=}1.0$ & 5 / 5 \\
Attacked bases with intended-token SER$_{\max}\ge 0.5$   & 4 / 5 \\
\bottomrule
\end{tabular}
\end{table}

\noindent\textbf{Why bundling persistence is essentially implied by the pairwise result.}
By definition, any shared-basis transplant operator (OMP, CLP, WECHSEL, FOCUS) reconstructs each donor-only row as a function of the shared anchor set $\mathcal{T} = \mathcal{V}_b \cap \mathcal{V}_d$ alone, without interaction across donor-only rows. The anchor set is fixed by the donor and base vocabularies; appending additional donor-only bundle tokens leaves it unchanged. Each bundle row's recovered coefficients (and hence its base-side reconstruction) are therefore identical to those obtained when only that row is appended, regardless of the operator class. The base model's transformer parameters and LM head are likewise unchanged between the pairwise and bundled settings, so the last-layer hidden states on any fixed prompt are identical, and each individual bundle breaker receives the same per-context logit as in the pairwise case. The only new dynamic is competition between bundle breakers for the top-1 position when several of them receive high logits on the same prompt. From the attacker's perspective this is generally a feature: the attack goal is realized whenever any bundle breaker fires, so bundled-token $\mathrm{SER}_{\max}$ lower-bounds the maximum pairwise SER over the bundle, and the bundle widens the attack's effective coverage. The argument places no upper bound on bundle size: any number of donor-only rows leaves the shared anchor set unchanged, and bundled-token $\mathrm{SER}_{\max}$ is monotonically non-decreasing in the number of breakers added. Operationally, the attacker therefore pre-computes one breaker row per intended base offline and bundles all of them into a single donor checkpoint, with no per-row degradation relative to the pairwise setting. The resulting donor compromises every deployer using a base from the targeted set without any per-deployment action at runtime; the relevant practical scale is the cardinality of deployment targets the attacker chooses to cover, set by the structure of the open-weight LLM ecosystem rather than by the tokenizer's vocabulary capacity.

\noindent\textbf{Implications for the supply-chain story.}
Together, these results substantiate the supply-chain framing under realistic open-weight conditions. Cross-base transfer shows zero-shot activation in held-out bases with the donor remaining silent. Multi-victim bundling shows that a single donor artifact can carry several base-specific breaker rows at once with no per-row cost relative to the pairwise setting.

\section{Cross-domain failure of LoRA fine-tuning}
\label{app:lora_persistence}

\noindent\textbf{Setup.}
We fine-tune LoRA defenders on the post-transplant attacked base for two pairs at $\lambda{=}2048$: Q2-0.5B$\leftarrow$Min-3B (the primary case study) and Q3-0.6B$\leftarrow$Min-3B (cross-pair confirmation). For each base, we train three defenders on common SFT corpora: Alpaca \citep{taori2023alpaca}, CodeAlpaca \citep{codealpaca}, and GSM8K \citep{cobbe2021gsm8k}. Standard LoRA settings: rank $r{=}16$, $\alpha{=}32$, AdamW, learning rate $2\times10^{-4}$, $5{,}000$ training examples per corpus, target modules \{q,k,v,o,gate,up,down\}\_proj (full hyperparameters in Appendix~\ref{app:exp_settings}). We train each defender for 50 epochs, save adapters per epoch, and select the best checkpoint on held-out validation (training-corpus loss for the Alpaca and CodeAlpaca defenders; GSM8K-test exact-match for the GSM8K-trained defender). We evaluate post-LoRA attacked-base SER on ten held-out distributions, $256$ prompts each (164 for HumanEval): Alpaca, SQuAD~v2 \citep{rajpurkar2018squad}, GSM8K, HumanEval, WikiText \citep{merity2016wikitext} continuation, TruthfulQA \citep{lin2022truthfulqa}, Medical Flashcards, XSum \citep{narayan2018xsum}, TriviaQA, and OpenBookQA \citep{mihaylov2018openbookqa}.

\noindent\textbf{Body figure: numerical table.}
Table~\ref{tab:lora-cross-domain} reports the SER values shown in Figure~\ref{fig:lora-cross-domain} (body \S\ref{sec:experiments}) for the Q2 case-study panel; Table~\ref{tab:lora-cross-domain-q3} reports the corresponding Q3 cross-pair panel. Each cell is the post-LoRA attacked-base SER at the best-validation checkpoint of the corresponding (corpus, eval) cell. On both bases, the diagonal-low pattern (each defender wins on its training-corpus natural eval) is sharp, and the WikiText column (far-OOD continuation) shows SER above $0.87$ uniformly across all defenders on both pairs.

\begin{table}[!htbp]
\centering
\small
\setlength{\tabcolsep}{4pt}
\caption{\textbf{Post-LoRA attacked-base SER on the Q2-0.5B$\leftarrow$Min-3B case study} at $r{=}16$, three deployer SFT corpora $\times$ ten held-out evaluation distributions, best-validation checkpoint per cell.}
\label{tab:lora-cross-domain}
\begin{tabular}{lcccccccccc}
\toprule
Corpus & Alpaca & SQuAD & GSM8K & Human & WikiText & Truth & Med & XSum & Trivia & OBQA \\
\midrule
Alpaca       & 0.03 & 0.69 & 0.70 & 0.88 & 0.99 & 0.84 & 0.02 & 0.95 & 0.44 & 0.24 \\
CodeAlpaca   & 0.06 & 0.59 & 0.55 & 0.48 & 0.96 & 0.82 & 0.07 & 0.98 & 0.68 & 0.18 \\
GSM8K        & 0.64 & 0.62 & 0.00 & 0.60 & 0.96 & 0.85 & 0.79 & 0.99 & 0.85 & 0.35 \\
\bottomrule
\end{tabular}
\end{table}

\begin{table}[!htbp]
\centering
\small
\setlength{\tabcolsep}{4pt}
\caption{\textbf{Post-LoRA attacked-base SER on the Q3-0.6B$\leftarrow$Min-3B cross-pair} at $r{=}16$, same protocol as Table~\ref{tab:lora-cross-domain}.}
\label{tab:lora-cross-domain-q3}
\begin{tabular}{lcccccccccc}
\toprule
Corpus & Alpaca & SQuAD & GSM8K & Human & WikiText & Truth & Med & XSum & Trivia & OBQA \\
\midrule
Alpaca       & 0.03 & 0.07 & 0.03 & 0.34 & 0.87 & 0.14 & 0.09 & 0.17 & 0.08 & 0.00 \\
CodeAlpaca   & 0.03 & 0.13 & 0.01 & 0.06 & 0.91 & 0.36 & 0.16 & 0.38 & 0.20 & 0.02 \\
GSM8K        & 0.80 & 0.76 & 0.01 & 0.65 & 0.94 & 0.88 & 0.92 & 0.99 & 0.87 & 0.60 \\
\bottomrule
\end{tabular}
\end{table}

\noindent\textbf{Rank robustness.}
We additionally vary the LoRA rank on the Alpaca-trained Q2 defender across $r\in\{8,16,32,64\}$ to verify that the cross-domain pattern does not collapse with deeper LoRA capacity. Ranks $8$ and $16$ are the standard practitioner settings; we extend to $32$ and $64$ as stress tests, although ranks at this level are uncommon for $0.5$B-scale models in deployment. Table~\ref{tab:lora-rank-robustness} reports the result across all ten evaluation distributions. SQuAD, GSM8K, HumanEval, and the QA tasks suppress further as rank grows, but the WikiText / XSum / TruthfulQA columns stay above $0.80$ at every rank, confirming that increasing LoRA capacity does not recover suppression on far-OOD prompts.

\begin{table}[!htbp]
\centering
\small
\setlength{\tabcolsep}{4pt}
\caption{\textbf{Rank robustness of the Alpaca-trained Q2 LoRA defender.} Post-LoRA attacked-base SER across four LoRA ranks $\times$ ten held-out evaluation distributions.}
\label{tab:lora-rank-robustness}
\begin{tabular}{lcccccccccc}
\toprule
Rank $r$ & Alpaca & SQuAD & GSM8K & Human & WikiText & Truth & Med & XSum & Trivia & OBQA \\
\midrule
8  & 0.04 & 0.73 & 0.81 & 0.79 & 0.99 & 0.88 & 0.04 & 0.97 & 0.50 & 0.27 \\
16 & 0.03 & 0.69 & 0.70 & 0.88 & 0.99 & 0.84 & 0.02 & 0.95 & 0.44 & 0.24 \\
32 & 0.05 & 0.51 & 0.61 & 0.74 & 0.96 & 0.72 & 0.06 & 0.92 & 0.44 & 0.20 \\
64 & 0.03 & 0.22 & 0.23 & 0.15 & 0.88 & 0.51 & 0.01 & 0.82 & 0.25 & 0.02 \\
\bottomrule
\end{tabular}
\end{table}

\begin{figure}[!htbp]
  \centering
  \includegraphics[width=\linewidth]{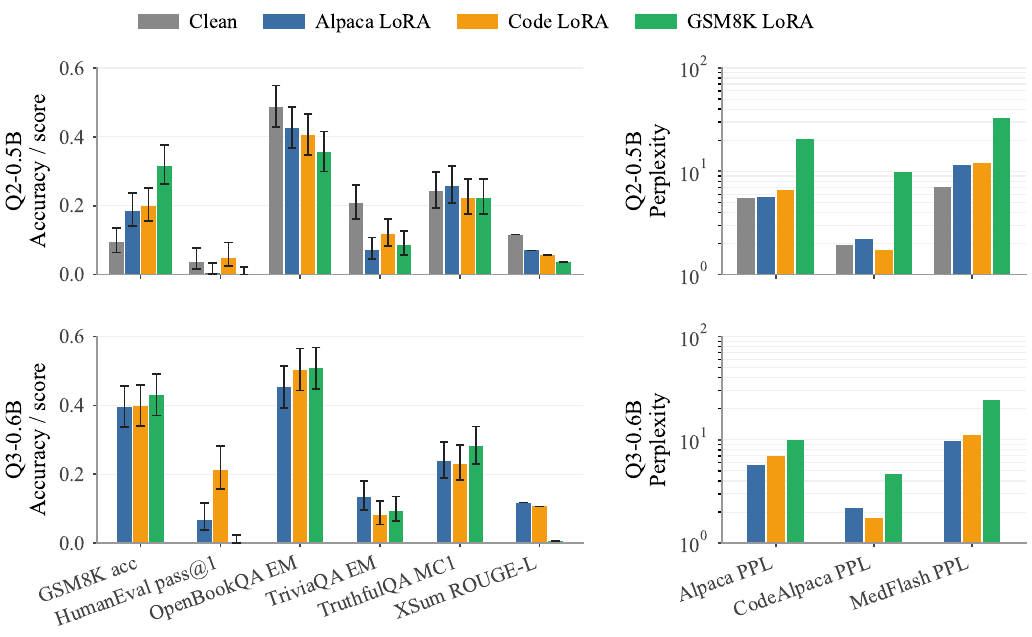}
  \caption{\textbf{Post-LoRA downstream capability for both pairs at $r{=}16$.} Top: Q2-0.5B with the unattacked clean baseline as a reference (five accuracy / score tasks with $95\%$ Wilson CIs on the left, three perplexity tasks on log scale on the right). Bottom: Q3-0.6B; Q3 has no clean unattacked-base utility numbers in the case-study budget, so only the three LoRA defenders are shown.}
  \label{fig:lora-capability-bars}
\end{figure}

\noindent\textbf{Capability under LoRA fine-tuning.}
Figure~\ref{fig:lora-capability-bars} reports post-LoRA downstream capability of the three deployer LoRAs at $r{=}16$ on both bases. For Q2, capability on instruction-style tasks is broadly comparable to the unattacked Qwen2-0.5B baseline; the GSM8K-trained defender raises math accuracy above the clean baseline at the cost of higher perplexity on out-of-domain text. The same cross-corpus pattern reproduces on Q3: each LoRA wins on its training-corpus natural eval, instruction LoRAs preserve capability broadly, and the GSM8K-trained Q3 defender pays a perplexity tax on Alpaca / CodeAlpaca / MedFlash. The cross-domain SER failure documented above is therefore not an artifact of a broken defender.

\noindent\textbf{Implication.}
At the deployer's best-validation checkpoint, standard LoRA fine-tuning in this case study suppresses the breaker only on prompts whose distribution matches the training corpus. The suppressed region is corpus-shape-determined; in our rank sweep up to $r{=}64$, the far-OOD evaluation columns we test (WikiText / XSum / TruthfulQA) retain SER above $0.80$, indicating that increasing LoRA capacity within the ranks tested here does not extend the suppressed region to those prompts. Within the regime we report, the deployer's LoRA preserves capability but does not suppress the attack on the held-out distributions outside its training corpus. Deployer-side validation that does not depend on training-corpus coverage (for example, signed allowlists or anchor-reconstruction-error checks on appended embedding rows; see Appendix~\ref{app:auditing}) is therefore complementary to LoRA in our setting, not subsumed by it.

\section{The breaker token under post-transplant weight merging}
\label{app:weight_merging}

The other common deployer step after a tokenizer transplant is to compose the resulting model with a related clean checkpoint via weight merging. Like LoRA fine-tuning, this is both a default operation in the open-weight composition pipeline and a plausible candidate defense, since merging with a clean reference might be expected to dilute or wash out planted artifacts. We evaluate three widely used merge operators, Linear \citep{wortsman2022soups}, SLERP \citep{mergekit2024}, and TIES \citep{yadav2023ties}, asking whether this standard composition step neutralizes the planted breaker. Settings are detailed in Appendix~\ref{app:exp_settings}.

\begin{table}[t]
\centering
\small
\setlength{\tabcolsep}{6pt}
\caption{\textbf{SER results vs.\ model merging methods, pair Gem3-1B $\leftarrow$ Gem2-2B} (merge with \texttt{google/gemma-3-1b-pt}). Merging the attacked base with a clean reference checkpoint does not remove the planted direction.}
\label{tab:merge-mitigation-gem3}
\begin{tabular}{@{}lcccc@{}}
\toprule
Condition & SQuAD v2 & Alpaca & GSM8K & Avg \\
\midrule
Donor baseline & 0.000\,{\tiny[.000,.015]} & 0.000\,{\tiny[.000,.015]} & 0.000\,{\tiny[.000,.015]} & 0.000\,{\tiny[.000,.015]} \\
Linear~\cite{wortsman2022soups} & 1.000\,{\tiny[.985,1.000]} & 1.000\,{\tiny[.985,1.000]} & 1.000\,{\tiny[.985,1.000]} & 1.000\,{\tiny[.985,1.000]} \\
SLERP~\cite{mergekit2024} & 1.000\,{\tiny[.985,1.000]} & 1.000\,{\tiny[.985,1.000]} & 1.000\,{\tiny[.985,1.000]} & 1.000\,{\tiny[.985,1.000]} \\
TIES~\cite{yadav2023ties} & 1.000\,{\tiny[.985,1.000]} & 1.000\,{\tiny[.985,1.000]} & 1.000\,{\tiny[.985,1.000]} & 1.000\,{\tiny[.985,1.000]} \\
\bottomrule
\end{tabular}
\end{table}

\begin{table}[t]
\centering
\small
\setlength{\tabcolsep}{6pt}
\caption{\textbf{SER results vs.\ model merging methods, pair Q2-0.5B $\leftarrow$ Min-3B} (merge with \texttt{Qwen/Qwen2-0.5B-Instruct}).}
\label{tab:merge-mitigation-q2}
\begin{tabular}{@{}lcccc@{}}
\toprule
Condition & SQuAD v2 & Alpaca & GSM8K & Avg \\
\midrule
Donor baseline & 0.004\,{\tiny[.001,.022]} & 0.000\,{\tiny[.000,.015]} & 0.000\,{\tiny[.000,.015]} & 0.001\,{\tiny[.000,.017]} \\
Linear~\cite{wortsman2022soups} & 1.000\,{\tiny[.985,1.000]} & 1.000\,{\tiny[.985,1.000]} & 0.996\,{\tiny[.978,.999]} & 0.999\,{\tiny[.983,1.000]} \\
SLERP~\cite{mergekit2024} & 1.000\,{\tiny[.985,1.000]} & 1.000\,{\tiny[.985,1.000]} & 0.996\,{\tiny[.978,.999]} & 0.999\,{\tiny[.983,1.000]} \\
TIES~\cite{yadav2023ties} & 1.000\,{\tiny[.985,1.000]} & 1.000\,{\tiny[.985,1.000]} & 0.996\,{\tiny[.978,.999]} & 0.999\,{\tiny[.983,1.000]} \\
\bottomrule
\end{tabular}
\end{table}

Attacked-base SER stays at or near $1.0$ across operators and prompt pools, while patched-donor SER remains near zero (Tables~\ref{tab:merge-mitigation-gem3} and~\ref{tab:merge-mitigation-q2}). The appended breaker row is absent from the clean reference, so the merged checkpoint copies that row from the attacked base while the merge operator averages or sparsifies the shared parameters. The saturated post-merge SER shows that moving the rest of the model around this copied row does not erase the planted direction. Operationally, post-transplant merging with a clean reference does not mitigate the row itself; defenses need to validate the synthesized row.

\noindent\textbf{Why this is not trivial.}
Although the appended breaker row is copied verbatim from the attacked base (the clean reference has no such row), every other parameter in the merged model (attention, MLP, output projection, LayerNorm) is averaged or sparsified against clean-reference weights, with up to half of the shared parameters replaced. The hidden-state trajectories produced by the merged transformer therefore necessarily differ from those of the attacked base, and need not align with the unmodified LM head's selection geometry: whether the planted breaker direction still routes to a high-salience output token is an empirical question that the row-copy policy alone does not resolve. Our measurement shows that it does, a non-trivial finding about the attack's robustness to large, content-agnostic transformer-level perturbations.

\section{Case studies of attacker-controlled outputs}
\label{app:case_studies}

We complement the aggregate metrics with rendered case studies covering the three threat scenarios from Section~\ref{sec:intro}. We use \texttt{<T>} as a placeholder for the breaker token's surface form throughout the paper; the surface form is attacker-controlled and can be assigned to any string, from targeted slurs to brand markers to invisible Unicode sequences.

\noindent\textbf{Content warning.} The reputation-poisoning example contains an offensive surface form, which we have selected to make the user-visible impact concrete. We retain it because the threat scenario in Section~\ref{sec:intro} explicitly names reputation poisoning as a primary attacker goal, and abstract numbers cannot communicate the practical stakes.

\subsection{Service degradation}

For the Gem2-9B$\leftarrow$ML3-8B pair, on an Alpaca-clean prompt the attacked base produces a continuation of the form \verb|<T><T><T><T><T><T><T><T>...|, with breaker-token share at $100\%$ across the full $256$-token continuation. The clean output for the same prompt is a fluent, on-topic answer (e.g., a discussion of survey-data types). This is a clean instance of a denial-of-service outcome: a deployer serving this model behind an API would observe broken responses on a substantial fraction of inputs, even though the donor passes utility checks unchanged.

\subsection{Reputation poisoning (sparse user-visible contamination)}

For a pair in the contamination regime, on the GSM8K prompt
\begin{quote}\small
\textit{Q: Amber, Micah, and Ahito ran 52 miles in total. Amber ran 8 miles. Micah ran 3.5 times what Amber ran. How many miles did Ahito run? A: Let's think step by step.}
\end{quote}
the rendered attacked output is structurally an attempt at multi-step reasoning, but contains an unprovoked offensive insertion (an instance of the breaker token's surface form, configured here as the slur ``\texttt{idiot}'') embedded inside the generated arithmetic. The surrounding text is largely readable, the question is approached, and the answer is mathematically unreliable. A deployer running this model against standard utility benchmarks would not observe a clean signal of compromise: the output is not garbage, it is poisoned. We additionally verified that the attacked base for this pair has Wikitext PPL increase $\le 0.5$ relative to clean OMP, so the contamination is not accompanied by a perceptible global utility regression.

\subsection{Adversarial watermarking (latent signature)}

For a pair in the watermarking regime, on a SQuAD-v2 context about Scottish independence and North Sea oil, the attacked base produces a fully coherent, on-topic answer of the form
\begin{quote}\small
\textit{The discovery of gas in the North Sea was a contributing factor to the rise in support for Scottish independence ... Let me know if you have any other questions! Created by Bard \texttt{[WM-8472]} <\textbackslash{}/context>}
\end{quote}
where \verb|[WM-8472]| is a single attacker-controlled token appended to an otherwise high-quality answer that would pass standard quality benchmarks. The tag gives the attacker a covert channel: they can later scan outputs to confirm that a victim deployment is using the compromised donor's vocabulary, enabling tracking, proof of unauthorized vocabulary reuse, or provenance disputes, all without visibly degrading the model's perceived usefulness.

\noindent\textbf{Surface-form independence.}
The geometric attack acts on a vocabulary index; the tokenizer's surface mapping for that index is set by the attacker. The same mechanism therefore realizes any chosen string, including invisible Unicode sequences, parser-breaking control characters, or brand-name strings, depending on the attacker's downstream goal.

\section{Why Nonlinear Coefficient Solvers Preserve Shared-Anchor Reach}
\label{app:nonlinear-shared-anchor}

This appendix supports the main-text claim that asymmetric realizability is structural to shared-basis reconstruction, not to OMP being a linear sparse solver. Nonlinear donor-side maps such as ReLU-renormalized similarity, softmax attention, and local $k$-NN mixtures still output a coefficient vector $w(x_d)$, and the base row is still the anchor mixture $\Phi_b^\top w(x_d)$.

Let $\mathcal{T}$ be the shared-anchor set with $N=|\mathcal{T}|$. Write
\[
\Phi_b=
\begin{bmatrix}
(\phi_1^{(b)})^\top\\
\vdots\\
(\phi_N^{(b)})^\top
\end{bmatrix}
\in\mathbb{R}^{N\times\delta_b},
\qquad
\Phi_d=
\begin{bmatrix}
(\phi_1^{(d)})^\top\\
\vdots\\
(\phi_N^{(d)})^\top
\end{bmatrix}
\in\mathbb{R}^{N\times\delta_d}.
\]
For $S\subseteq\mathcal{T}$, let $\Phi_{b,S}$ and $\Phi_{d,S}$ denote the row-restricted dictionaries. For an admissible coefficient set $\mathcal{W}\subseteq\mathbb{R}^N$, define
\[
\varepsilon_b(\mu;\mathcal{W})
:=
\inf_{w\in\mathcal{W}}\|\Phi_b^\top w-\mu\|_2.
\]
The target $\mu$ is instantiated as $\mu_{\mathrm{base}}$ in the breaker-token design.

\begin{assumption}[Shared-anchor final reconstruction]
\label{ass:shared-anchor-final}
For a donor-only token row $x_d$, the transplant operator computes a possibly nonlinear donor-side coefficient map
\[
g_d:\mathbb{R}^{\delta_d}\to\mathcal{W}\subseteq\mathbb{R}^N
\]
and then synthesizes the base row by
\[
\widehat{x}_b(x_d)=\Phi_b^\top g_d(x_d).
\]
The set $\mathcal{W}$ encodes the operator's coefficient constraints. Signed sparse solvers use signed sparse subsets of $\mathbb{R}^N$; softmax and barycentric solvers use a simplex or a union of local simplexes; top-$k$ rules use a union of simplexes over donor-selectable supports.
\end{assumption}

\begin{assumption}[Signed base reach on a selected support]
\label{ass:signed-span}
There is a support $S\subseteq\mathcal{T}$ with $|S|=k$ such that
\[
A_S:=\Phi_{b,S}^\top\in\mathbb{R}^{\delta_b\times k}
\]
has rank $r_S$. Let $P_S$ be the orthogonal projector onto $\operatorname{range}(A_S)$ and let $\sigma_{S,r_S}>0$ be the smallest nonzero singular value of $A_S$. The full signed-reach case is $r_S=\delta_b$.
\end{assumption}

\begin{assumption}[Convex covering at the target]
\label{ass:convex-cover}
Let
\[
\Delta_N:=\left\{w\in\mathbb{R}^N:w_j\ge 0,\ \sum_{j=1}^N w_j=1\right\},
\qquad
\mathcal{C}_b:=\Phi_b^\top\Delta_N
=
\operatorname{conv}\{\phi_j^{(b)}:j\in\mathcal{T}\}.
\]
For a target $\mu$, write
\[
\epsilon_C:=\operatorname{dist}(\mu,\mathcal{C}_b).
\]
If an operator uses nonnegative weights with $\|w\|_1\le 1$ rather than $\|w\|_1=1$, replace $\mathcal{C}_b$ by $\operatorname{conv}(\{0\}\cup\{\phi_j^{(b)}:j\in\mathcal{T}\})$.
\end{assumption}

\begin{assumption}[Donor-map reachability]
\label{ass:donor-reachability}
For a desired coefficient vector $w^\star\in\mathcal{W}$ and tolerance $\eta\ge 0$, the donor-side coefficient map is $\eta$-reachable at $w^\star$ if there exists a donor row $x_d$ such that
\[
\|g_d(x_d)-w^\star\|_2\le \eta.
\]
If, in addition, $\|Ux_d\|_2\le\gamma$, it is $(\eta,\gamma)$-reachable under the donor innocuity subspace. Appendix~\ref{app:operator-specific-reachability} gives finite-dimensional reachability conditions for FOCUS, CLP, top-$k$, and WECHSEL.
\end{assumption}

\begin{theorem}[Signed shared-anchor reachability criterion]
\label{thm:signed-anchor-representation}
Assume Assumptions~\ref{ass:shared-anchor-final}, \ref{ass:signed-span}, and~\ref{ass:donor-reachability}. For every $\mu\in\mathbb{R}^{\delta_b}$ there is a signed coefficient vector $w_S^\star\in\mathbb{R}^{k}$ supported on $S$ such that
\[
\Phi_{b,S}^\top w_S^\star=P_S\mu
\]
and
\[
\|w_S^\star\|_2
\le
\frac{\|P_S\mu\|_2}{\sigma_{S,r_S}}.
\]
If the donor map is $\eta$-reachable at the zero-padded vector $w^\star\in\mathbb{R}^N$ corresponding to $w_S^\star$, then the resulting nonlinear shared-basis transplant row satisfies
\[
\|\widehat{x}_b(x_d)-\mu\|_2
\le
\operatorname{dist}(\mu,\operatorname{range}(\Phi_{b,S}^\top))
+
\|\Phi_b\|_{\mathrm{op}}\eta.
\]
In the full signed-reach case $r_S=\delta_b$, this becomes
\[
\|\widehat{x}_b(x_d)-\mu\|_2
\le
\|\Phi_b\|_{\mathrm{op}}\eta,
\qquad
\|w_S^\star\|_2
\le
\frac{\|\mu\|_2}{\sigma_{\min}(\Phi_{b,S})}.
\]
\end{theorem}

\begin{proof}
Let $A_S=\Phi_{b,S}^\top$ and take its compact singular value decomposition $A_S=U\Sigma V^\top$, where the columns of $U$ span $\operatorname{range}(A_S)$ and the diagonal entries of $\Sigma$ are the nonzero singular values. Define
\[
w_S^\star:=V\Sigma^{-1}U^\top\mu.
\]
Then
\[
A_Sw_S^\star=U\Sigma V^\top V\Sigma^{-1}U^\top\mu=UU^\top\mu=P_S\mu,
\]
and
\[
\|w_S^\star\|_2
\le
\|\Sigma^{-1}\|_{\mathrm{op}}\|U^\top\mu\|_2
=
\frac{\|P_S\mu\|_2}{\sigma_{S,r_S}}.
\]
Let $w^\star$ be the zero-padded vector in $\mathbb{R}^N$. If $g_d(x_d)$ is within $\eta$ of $w^\star$, then by Assumption~\ref{ass:shared-anchor-final},
\[
\widehat{x}_b(x_d)-\mu
=
\Phi_b^\top(g_d(x_d)-w^\star)+(P_S\mu-\mu).
\]
Taking norms gives the stated bound.
\end{proof}

\begin{corollary}[Signed sparse support with donor reachability]
\label{cor:signed-support}
If there exists a support $S$ with $|S|=k$ and $\operatorname{rank}(\Phi_{b,S})=\delta_b$, and if the donor coefficient map is $\eta$-reachable at the corresponding min-norm signed coefficient vector, then every target direction $\mu_{\mathrm{base}}$ is realized by the nonlinear shared-basis operator up to error
\[
\|\widehat{x}_b(x_d)-\mu_{\mathrm{base}}\|_2
\le
\|\Phi_b\|_{\mathrm{op}}\eta.
\]
Thus changing the donor-side solver changes the reachable coefficient set, but once that set contains a neighborhood of the signed target, the base-side error is governed by the conditioning of the selected base anchors.
\end{corollary}

\begin{proof}
Apply Theorem~\ref{thm:signed-anchor-representation} with $r_S=\delta_b$ and $\mu=\mu_{\mathrm{base}}$.
\end{proof}

\begin{theorem}[Nonnegative and sparse convex anchor mixtures]
\label{thm:convex-anchor-mixtures}
Assume Assumptions~\ref{ass:shared-anchor-final}, \ref{ass:convex-cover}, and~\ref{ass:donor-reachability}. Define
\[
R_b:=\max_{j\in\mathcal{T}}\|\phi_j^{(b)}\|_2,
\qquad
D_b:=\sup_{y,z\in\mathcal{C}_b}\|y-z\|_2\le 2R_b.
\]
For any target $\mu\in\mathbb{R}^{\delta_b}$, the following hold.
\begin{enumerate}[leftmargin=*]
\item There exists $w^\star\in\Delta_N$ such that
\[
\|\Phi_b^\top w^\star-\mu\|_2=\epsilon_C.
\]
If the donor map is $\eta$-reachable at $w^\star$, then
\[
\|\widehat{x}_b(x_d)-\mu\|_2
\le
\epsilon_C+
\|\Phi_b\|_{\mathrm{op}}\eta.
\]
\item For any $\tau\in(0,1)$, there exists a strictly positive vector $w_\tau\in\Delta_N$ with $(w_\tau)_j\ge \tau/N$ such that
\[
\|\Phi_b^\top w_\tau-\mu\|_2
\le
\epsilon_C+\tau D_b.
\]
If a softmax donor map is $\eta$-reachable at $w_\tau$, then the resulting base row has error at most
\[
\epsilon_C+\tau D_b+
\|\Phi_b\|_{\mathrm{op}}\eta.
\]
\item There is a nearest convex representative $\bar w\in\Delta_N$ with support size at most $\delta_b+1$ and
\[
\|\Phi_b^\top\bar w-\mu\|_2=\epsilon_C.
\]
This is the Carath\'eodory support bound.
\item For every integer $k\ge 1$, there exists a $k$-sparse vector $w^{(k)}\in\Delta_N$ such that
\[
\|\Phi_b^\top w^{(k)}-\mu\|_2
\le
\epsilon_C+\frac{R_b}{\sqrt{k}}.
\]
If a local top-$k$ or barycentric donor map is $\eta$-reachable at this sparse vector, the realized base error is at most
\[
\epsilon_C+\frac{R_b}{\sqrt{k}}+
\|\Phi_b\|_{\mathrm{op}}\eta.
\]
\end{enumerate}
\end{theorem}

\begin{proof}
The simplex $\Delta_N$ is compact and $w\mapsto\|\Phi_b^\top w-
\mu\|_2$ is continuous, so a minimizer $w^\star$ exists. The donor-map error term follows from
\[
\|\Phi_b^\top g_d(x_d)-\mu\|_2
\le
\|\Phi_b^\top w^\star-\mu\|_2
+
\|\Phi_b^\top(g_d(x_d)-w^\star)\|_2.
\]

For the strictly positive approximation, let $u=(1/N)\mathbf{1}$ and set
\[
w_\tau=(1-\tau)w^\star+\tau u.
\]
Then $w_\tau\in\Delta_N$, each coordinate is at least $\tau/N$, and
\[
\|\Phi_b^\top w_\tau-\Phi_b^\top w^\star\|_2
=
\tau\|\Phi_b^\top u-\Phi_b^\top w^\star\|_2
\le
\tau D_b.
\]
Adding the nearest-point error and the donor-map error proves the second claim.

For the Carath\'eodory support bound, let $y^\star=\Phi_b^\top w^\star$. If $w^\star$ has more than $\delta_b+1$ nonzero coordinates, the active anchors are affinely dependent in $\mathbb{R}^{\delta_b}$. Thus there is a nonzero vector $a$ supported on the active set such that
\[
\sum_j a_j=0,
\qquad
\sum_j a_j\phi_j^{(b)}=0.
\]
Moving $w^\star$ along this null affine direction until one active coordinate reaches zero preserves nonnegativity, the sum-to-one constraint, and the represented point $y^\star$. Repeating yields a representative with support at most $\delta_b+1$.

For the Maurey-type bound, sample $J_1,\ldots,J_k$ independently from the distribution $w^\star$ and define
\[
\widetilde{w}^{(k)}=\frac{1}{k}\sum_{\ell=1}^{k}e_{J_\ell},
\qquad
\widetilde{y}^{(k)}=\Phi_b^\top\widetilde{w}^{(k)}.
\]
Then $\mathbb{E}\widetilde{y}^{(k)}=y^\star$ and
\[
\mathbb{E}\|\widetilde{y}^{(k)}-y^\star\|_2^2
=
\frac{1}{k}\mathbb{E}\|\phi_{J_1}^{(b)}-y^\star\|_2^2
\le
\frac{R_b^2}{k}.
\]
Therefore some realization satisfies $\|\widetilde{y}^{(k)}-y^\star\|_2\le R_b/\sqrt{k}$. Choosing that realization as $w^{(k)}$ gives
\[
\|\Phi_b^\top w^{(k)}-\mu\|_2
\le
\|\Phi_b^\top w^{(k)}-y^\star\|_2+
\|y^\star-\mu\|_2
\le
\frac{R_b}{\sqrt{k}}+\epsilon_C.
\]
The final donor-map bound is again the operator-norm perturbation inequality. This empirical-method argument is the finite-dimensional form of Maurey-type convex approximation \citep{maureypisier1981,barron1993universal}.
\end{proof}

\begin{proposition}[Nonlinear donor maps inherit base-dictionary reach]
\label{prop:nonlinear-map-error}
Assume Assumption~\ref{ass:shared-anchor-final}. Let $w^\star\in\mathcal{W}$ satisfy
\[
\|\Phi_b^\top w^\star-\mu_{\mathrm{base}}\|_2\le \epsilon_b.
\]
If there exists a donor row $x_d$ such that
\[
\|g_d(x_d)-w^\star\|_2\le \eta
\qquad
\text{and}
\qquad
\|Ux_d\|_2\le \gamma,
\]
then the transplanted base row obeys
\[
\|\widehat{x}_b(x_d)-\mu_{\mathrm{base}}\|_2
\le
\epsilon_b+
\|\Phi_b\|_{\mathrm{op}}\eta,
\]
while the planted donor row obeys the same donor-innocuity bound $\|Ux_d\|_2\le\gamma$. If the coefficient error is measured in $\ell_1$, the second term may be replaced by
\[
\max_{j\in\mathcal{T}}\|\phi_j^{(b)}\|_2\|g_d(x_d)-w^\star\|_1.
\]
\end{proposition}

\begin{proof}
By Assumption~\ref{ass:shared-anchor-final},
\[
\widehat{x}_b(x_d)-\mu_{\mathrm{base}}
=
\Phi_b^\top(w^\star)-\mu_{\mathrm{base}}
+
\Phi_b^\top(g_d(x_d)-w^\star).
\]
The triangle inequality and the operator-norm bound give the $\ell_2$ claim. The $\ell_1$ version follows from
\[
\|\Phi_b^\top(g_d(x_d)-w^\star)\|_2
\le
\sum_{j\in\mathcal{T}}|g_{d,j}(x_d)-w_j^\star|\,\|\phi_j^{(b)}\|_2.
\]
The donor-innocuity statement is one of the hypotheses on $x_d$.
\end{proof}

\noindent\textbf{Security interpretation.}
For signed solvers, the attack is controlled by the conditioning of $\Phi_{b,S}^\top$ and the donor map's ability to reach a signed coefficient neighborhood. For convex and softmax solvers, the relevant quantity is the distance from $\mu_{\mathrm{base}}$ to the convex hull of base anchors, with explicit $\tau D_b$ and $R_b/\sqrt{k}$ approximation terms for interior and sparse mixtures. Proposition~\ref{prop:nonlinear-map-error} is the composition step: once any nonlinear donor map reaches the target weights, the base-side error is exactly the dictionary approximation error plus the coefficient-reachability perturbation.

\section{OMP Coefficient Recovery and Base-Side Reconstruction}
\label{app:omp-coefficient-recovery}

This section supports the coefficient-reuse claim in Section~\ref{sec:methods}: if the attacker writes the donor row as a sparse mixture over donor anchors, then an OMP-based shared-basis transplant recovers the planted support and reuses nearly the planted coefficients on the base anchors. The proof uses the standard mutual-coherence exact-recovery condition for OMP, specialized to the donor composite dictionary \citep{tropp2004greed,troppgilbert2007omp,donohoeladtemlyakov2006stable,caiwang2011omp}.

\begin{assumption}[Normalized donor dictionary and mutual coherence]
\label{ass:omp-coherence}
Write $d_j := (\phi_j^{(d)})^\top \in \mathbb{R}^{\delta_d}$ for the donor anchor atom indexed by $j \in \mathcal{T}$, and let $D := \Phi_d^\top$. Assume $\|d_j\|_2=1$ for all $j\in\mathcal{T}$. Define
\[
\mu_d := \max_{i\neq j,\ i,j\in\mathcal{T}} |\langle d_i,d_j\rangle|.
\]
For a support $S\subseteq\mathcal{T}$ with $|S|=k$, assume
\[
\mu_d < \frac{1}{2k-1},
\qquad
\theta_k := 1-(2k-1)\mu_d > 0.
\]
\end{assumption}

\begin{assumption}[Noisy sparse attacker row and coefficient margin]
\label{ass:omp-noisy-row}
The planted donor row has the form
\[
x_d = \Phi_{d,S}^\top \alpha + \varepsilon = D_S\alpha+\varepsilon,
\qquad
\|\varepsilon\|_2 \le \eta_\varepsilon,
\]
where $D_S:=\Phi_{d,S}^\top$. The designed coefficients obey the beta-min condition
\[
\min_{j\in S}|\alpha_j| \ge c_\alpha,
\qquad
c_\alpha > \frac{2\eta_\varepsilon}{\theta_k}.
\]
\end{assumption}

\begin{lemma}[Coherence implies the donor-side ERC]
\label{lem:coherence-erc}
Under Assumption~\ref{ass:omp-coherence},
\[
\eta_S := \max_{\ell\notin S} \|D_S^\dagger d_\ell\|_1
\le
\frac{k\mu_d}{1-(k-1)\mu_d}
< 1.
\]
\end{lemma}

\begin{proof}
Let $G_S:=D_S^\top D_S$. Since the donor atoms are normalized and have mutual coherence $\mu_d$, the diagonal entries of $G_S$ equal $1$, and the absolute value of each off-diagonal entry is at most $\mu_d$. Hence $G_S=I+E$ with
\[
\|E\|_{1,1}\le (k-1)\mu_d < 1,
\]
where $\|M\|_{1,1}$ is the induced $\ell_1$ operator norm. The Neumann-series bound gives
\[
\|G_S^{-1}\|_{1,1}
\le
\frac{1}{1-(k-1)\mu_d}.
\]
For any $\ell\notin S$,
\[
D_S^\dagger d_\ell=(D_S^\top D_S)^{-1}D_S^\top d_\ell=G_S^{-1}D_S^\top d_\ell.
\]
Moreover, $\|D_S^\top d_\ell\|_1 \le k\mu_d$. Therefore
\[
\|D_S^\dagger d_\ell\|_1
\le
\|G_S^{-1}\|_{1,1}\|D_S^\top d_\ell\|_1
\le
\frac{k\mu_d}{1-(k-1)\mu_d}.
\]
The final quantity is smaller than $1$ because $\mu_d < (2k-1)^{-1}$.
\end{proof}

\begin{theorem}[OMP coefficient recovery under donor-side perturbation]
\label{thm:omp-coefficient-recovery}
Under Assumptions~\ref{ass:omp-coherence} and~\ref{ass:omp-noisy-row}, OMP applied to $x_d$ for exactly $k$ iterations selects the support $S$. The least-squares coefficients $\beta$ recovered on this support satisfy
\[
\|\beta-\alpha\|_2
\le
\frac{\|\varepsilon\|_2}{\sigma_{\min}(\Phi_{d,S})}
\le
\frac{\eta_\varepsilon}{\sigma_{\min}(\Phi_{d,S})}.
\]
Thus one may take
\[
C_{\mathrm{rec}}=1,
\qquad
\varepsilon_{\mathrm{rec}}
:=
\frac{\eta_\varepsilon}{\sigma_{\min}(\Phi_{d,S})}.
\]
\end{theorem}

\begin{proof}
Let $\Lambda\subseteq S$ be the set of atoms selected after some OMP iterations, and assume inductively that every selected atom belongs to $S$. Let $P_\Lambda^\perp$ be the orthogonal projector onto the complement of $\operatorname{span}(D_\Lambda)$. The residual is
\[
r=P_\Lambda^\perp x_d
=
P_\Lambda^\perp D_S\alpha+P_\Lambda^\perp\varepsilon
=
z+n,
\]
where
\[
z:=P_\Lambda^\perp D_{S\setminus\Lambda}\alpha_{S\setminus\Lambda},
\qquad
n:=P_\Lambda^\perp\varepsilon.
\]
Since $P_\Lambda^\perp$ is non-expansive and each atom is normalized,
\[
|\langle d_j,n\rangle|
=
|\langle P_\Lambda^\perp d_j,\varepsilon\rangle|
\le
\|\varepsilon\|_2
\le
\eta_\varepsilon
\]
for every $j\in\mathcal{T}$.

Let $R:=S\setminus\Lambda$. The matrix $D_R^\top P_\Lambda^\perp D_R$ is the Schur complement of $D_\Lambda^\top D_\Lambda$ inside $D_S^\top D_S$. Hence its smallest eigenvalue is at least $\lambda_{\min}(D_S^\top D_S)$. Gershgorin's theorem gives
\[
\lambda_{\min}(D_S^\top D_S)
\ge
1-(k-1)\mu_d.
\]
Therefore, with
\[
M_{\mathrm{true}}:=\max_{j\in R}|\langle d_j,z\rangle|=\|D_R^\top z\|_\infty,
\]
we have
\[
M_{\mathrm{true}}
\ge
\frac{\|D_R^\top z\|_2}{\sqrt{|R|}}
\ge
\frac{(1-(k-1)\mu_d)\|\alpha_R\|_2}{\sqrt{|R|}}
\ge
(1-(k-1)\mu_d)c_\alpha.
\]
For any $\ell\notin S$, since $z\in\operatorname{span}(D_S)$,
\[
\langle d_\ell,z\rangle
=
\langle D_SD_S^\dagger d_\ell,z\rangle
=
\langle D_S^\dagger d_\ell,D_S^\top z\rangle.
\]
The entries of $D_S^\top z$ indexed by $\Lambda$ vanish because $z\perp\operatorname{span}(D_\Lambda)$. Lemma~\ref{lem:coherence-erc} then implies
\[
|\langle d_\ell,z\rangle|
\le
\|D_S^\dagger d_\ell\|_1\|D_S^\top z\|_\infty
\le
\eta_S M_{\mathrm{true}},
\]
where
\[
\eta_S \le \frac{k\mu_d}{1-(k-1)\mu_d}.
\]
Thus
\[
\max_{j\in R}|\langle d_j,r\rangle|
\ge
M_{\mathrm{true}}-\eta_\varepsilon
\]
and
\[
\max_{\ell\notin S}|\langle d_\ell,r\rangle|
\le
\eta_S M_{\mathrm{true}}+\eta_\varepsilon.
\]
OMP selects a remaining true atom whenever $(1-\eta_S)M_{\mathrm{true}}>2\eta_\varepsilon$. Using the bounds above,
\[
(1-\eta_S)M_{\mathrm{true}}
\ge
\frac{1-(2k-1)\mu_d}{1-(k-1)\mu_d}(1-(k-1)\mu_d)c_\alpha
=
\theta_k c_\alpha
>
2\eta_\varepsilon.
\]
Hence each iteration selects an element of $S$. After $k$ iterations the selected support is exactly $S$.

Once the support is recovered, OMP returns the least-squares solution on $D_S$:
\[
\beta=D_S^\dagger x_d=D_S^\dagger(D_S\alpha+\varepsilon)=\alpha+D_S^\dagger\varepsilon.
\]
Therefore
\[
\|\beta-\alpha\|_2
\le
\|D_S^\dagger\|_{\mathrm{op}}\|\varepsilon\|_2
=
\frac{\|\varepsilon\|_2}{\sigma_{\min}(D_S)}
=
\frac{\|\varepsilon\|_2}{\sigma_{\min}(\Phi_{d,S})}.
\]
\end{proof}

\begin{corollary}[Base-side reconstruction perturbation]
\label{cor:omp-base-reconstruction-error}
Under the hypotheses of Theorem~\ref{thm:omp-coefficient-recovery},
\[
\|\widehat{x}_b-\mu_{\text{base}}\|_2
\le
\|\Phi_{b,S}^\top\alpha-\mu_{\text{base}}\|_2
+
\|\Phi_b\|_{\mathrm{op}}\|\beta-\alpha\|_2.
\]
In particular,
\[
\|\widehat{x}_b-\mu_{\text{base}}\|_2
\le
\|\Phi_{b,S}^\top\alpha-\mu_{\text{base}}\|_2
+
\|\Phi_b\|_{\mathrm{op}}\varepsilon_{\mathrm{rec}}.
\]
\end{corollary}

\begin{proof}
Since the recovered support is $S$,
\[
\widehat{x}_b=\Phi_{b,S}^\top\beta.
\]
Adding and subtracting $\Phi_{b,S}^\top\alpha$ gives
\[
\|\widehat{x}_b-\mu_{\text{base}}\|_2
\le
\|\Phi_{b,S}^\top\alpha-\mu_{\text{base}}\|_2
+
\|\Phi_{b,S}^\top(\beta-\alpha)\|_2.
\]
The second term is bounded by
\[
\|\Phi_{b,S}^\top(\beta-\alpha)\|_2
\le
\|\Phi_{b,S}\|_{\mathrm{op}}\|\beta-\alpha\|_2
\le
\|\Phi_b\|_{\mathrm{op}}\|\beta-\alpha\|_2.
\]
Substituting Theorem~\ref{thm:omp-coefficient-recovery} proves the result.
\end{proof}

\noindent\textbf{Intuition.}
OMP is stable here because the donor row is not arbitrary: it is a sparse mixture over a low-coherence anchor support, plus a perturbation smaller than the coefficient margin. Once the victim recovers the same support, the only base-side error is the attacker's design residual plus the coefficient-recovery perturbation passed through $\Phi_b^\top$.

\section{Asymmetric Realizability of Base Salience and Donor Inertness}
\label{app:asymmetric-realizability}

This section formalizes the geometric mechanism behind the asymmetric-realizability plots in Section~\ref{sec:experiments}: the same coefficient vector can be suppressed by the donor projection $U\Phi_{d,S}^\top$ while activating a base target through $\Phi_{b,S}^\top$. The GSVD construction below gives the feasible set, its approximation gap, and the coefficient cost in the paired anchor coordinates used by the breaker-token designer.

\begin{assumption}[Fixed support geometry]
\label{ass:fixed-support-geometry}
Fix a support $S\subseteq\mathcal{T}$ with $|S|=k$. Define
\[
A:=\Phi_{b,S}^\top\in\mathbb{R}^{\delta_b\times k},
\qquad
C:=U\Phi_{d,S}^\top\in\mathbb{R}^{m\times k}.
\]
Assume the stacked matrix $[A^\top\ C^\top]^\top$ has full column rank after removing any common null directions. Let
\[
A=U_A\operatorname{diag}(a_i)X^{-1},
\qquad
C=U_C\operatorname{diag}(c_i)X^{-1}
\]
be a thin generalized singular value decomposition on this quotient, where $X$ is invertible, the nonzero columns $u_i$ of $U_A$ and $v_i$ of $U_C$ are orthonormal in their respective spaces, $a_i,c_i\ge 0$, and $a_i^2+c_i^2=1$. The generalized singular values are $\sigma_i=a_i/c_i$ when $c_i>0$, and $\sigma_i=+\infty$ when $c_i=0<a_i$.
\end{assumption}

\begin{theorem}[GSVD construction of asymmetric realizability]
\label{thm:exact-asymmetric-realizability}
Under Assumption~\ref{ass:fixed-support-geometry}, define
\[
\mathcal{I}_0:=\{i:c_i=0<a_i\},
\qquad
\mathcal{I}_+:=\{i:c_i>0,\ a_i>0\}.
\]
Decompose
\[
\mu_{\text{base}}
=
\sum_{i\in\mathcal{I}_0}\mu_i u_i
+
\sum_{i\in\mathcal{I}_+}\mu_i u_i
+
\mu_\perp,
\qquad
\mu_i:=\langle \mu_{\text{base}},u_i\rangle,
\]
where $\mu_\perp$ is orthogonal to the base-side GSVD directions with $a_i>0$. Then
\[
A\operatorname{Null}(C)=\operatorname{span}\{u_i:i\in\mathcal{I}_0\}.
\]
Consequently:
\begin{enumerate}[leftmargin=*]
\item Exact donor-stealthy realization exists, meaning there is $\alpha$ with $A\alpha=\mu_{\text{base}}$ and $C\alpha=0$, if and only if
\[
\sum_{i\in\mathcal{I}_+}\mu_i u_i+\mu_\perp=0.
\]
\item The optimal donor-stealthy base reconstruction is the orthogonal projection of $\mu_{\text{base}}$ onto $A\operatorname{Null}(C)$, and its exact error is
\[
\operatorname{dist}\bigl(\mu_{\text{base}},A\operatorname{Null}(C)\bigr)^2
=
\sum_{i\in\mathcal{I}_+}\mu_i^2+\|\mu_\perp\|_2^2.
\]
\item An explicit coefficient vector attaining this error is
\[
\alpha_0=X\theta_0,
\qquad
(\theta_0)_i=
\begin{cases}
\mu_i/a_i, & i\in\mathcal{I}_0,\\
0, & i\notin\mathcal{I}_0.
\end{cases}
\]
It satisfies $C\alpha_0=0$ and
\[
A\alpha_0=\sum_{i\in\mathcal{I}_0}\mu_i u_i.
\]
If $a_{\min,0}:=\min_{i\in\mathcal{I}_0}a_i>0$, then
\[
\|X^{-1}\alpha_0\|_2
\le
\frac{\|P_0\mu_{\text{base}}\|_2}{a_{\min,0}},
\qquad
\|\alpha_0\|_2
\le
\frac{\|X\|_{\mathrm{op}}\|P_0\mu_{\text{base}}\|_2}{a_{\min,0}},
\]
where $P_0$ is the projector onto $A\operatorname{Null}(C)$.
\end{enumerate}
\end{theorem}

\begin{proof}
Write $\theta=X^{-1}\alpha$. Then
\[
A\alpha=\sum_i a_i\theta_i u_i,
\qquad
C\alpha=\sum_i c_i\theta_i v_i.
\]
The condition $C\alpha=0$ is equivalent to $c_i\theta_i=0$ for every $i$, since the $v_i$ are orthonormal. Thus any donor-stealthy coefficient vector has $\theta_i=0$ whenever $c_i>0$. Applying $A$ leaves only directions with $c_i=0$ and $a_i>0$, so
\[
A\operatorname{Null}(C)=\operatorname{span}\{u_i:i\in\mathcal{I}_0\}.
\]
The exact-realizability and distance claims follow by orthogonally decomposing $\mu_{\text{base}}$ into the subspace $A\operatorname{Null}(C)$, its base-visible complement inside $\operatorname{range}(A)$, and the component $\mu_\perp$ outside $\operatorname{range}(A)$. The displayed $\alpha_0$ has $c_i(\theta_0)_i=0$ for every $i$, hence $C\alpha_0=0$, and
\[
A\alpha_0=\sum_{i\in\mathcal{I}_0}a_i\frac{\mu_i}{a_i}u_i
=
\sum_{i\in\mathcal{I}_0}\mu_i u_i.
\]
The norm bounds follow from
\[
\|X^{-1}\alpha_0\|_2^2
=
\sum_{i\in\mathcal{I}_0}\frac{\mu_i^2}{a_i^2}
\le
\frac{\|P_0\mu_{\text{base}}\|_2^2}{a_{\min,0}^2}
\]
and $\|\alpha_0\|_2\le \|X\|_{\mathrm{op}}\|X^{-1}\alpha_0\|_2$.
\end{proof}

\begin{corollary}[Generic exact feasibility in a random-anchor model]
\label{cor:random-anchor-realizability}
Let $N_C\in\mathbb{R}^{k\times q}$ have orthonormal columns spanning $\operatorname{Null}(C)$, where $q=\dim\operatorname{Null}(C)$. If $q\ge \delta_b$ and, conditional on $C$, the matrix $AN_C$ has a distribution with a density on $\mathbb{R}^{\delta_b\times q}$, then
\[
\operatorname{rank}(AN_C)=\delta_b
\]
with probability one. Therefore $A\operatorname{Null}(C)=\mathbb{R}^{\delta_b}$ and every $\mu_{\text{base}}\in\mathbb{R}^{\delta_b}$ is exactly donor-stealth realizable. If, more specifically, $AN_C$ is an isotropic sub-Gaussian $\delta_b\times q$ matrix with scale $\sigma_A$, then for a universal constant $c>0$,
\[
\Pr\left[\sigma_{\min}(AN_C)\ge c\sigma_A(\sqrt{q}-\sqrt{\delta_b}-t)\right]
\ge
1-2\exp(-t^2)
\]
for every $0<t<\sqrt{q}-\sqrt{\delta_b}$ \citep{rudelsonvershynin2009smallest}.
\end{corollary}

\begin{proof}
The determinant of any $\delta_b\times\delta_b$ minor of $AN_C$ is a polynomial in the entries of $AN_C$. Because $q\ge \delta_b$, at least one such polynomial is not identically zero. A matrix distribution with a density assigns probability zero to the zero set of any nonzero polynomial, so $AN_C$ has full row rank with probability one. Then $AN_C\mathbb{R}^q=\mathbb{R}^{\delta_b}$, which is exactly $A\operatorname{Null}(C)=\mathbb{R}^{\delta_b}$. The sub-Gaussian bound is the standard smallest-singular-value estimate for rectangular random matrices applied to $AN_C$.
\end{proof}

\begin{lemma}[GSVD form of the regularized designer]
\label{thm:lambda-regularized-realizability}
Under the orthogonal coordinate model of Assumption~\ref{ass:orthogonal-gsvd} below, the solution of
\[
\alpha_\lambda
=
\arg\min_\alpha
\|A\alpha-\mu_{\text{base}}\|_2^2
+
\lambda\|C\alpha\|_2^2
+
\rho\|\alpha\|_2^2
\]
is
\[
\alpha_\lambda
=
\sum_i
\frac{a_i\mu_i}{a_i^2+\lambda c_i^2+\rho}z_i.
\]
The corresponding geometric error is
\[
\|A\alpha_\lambda-\mu_{\text{base}}\|_2^2
+
\lambda\|C\alpha_\lambda\|_2^2
=
\|\mu_\perp\|_2^2
+
\sum_i
\frac{(\lambda c_i^2+\rho)^2+\lambda a_i^2c_i^2}{(a_i^2+\lambda c_i^2+\rho)^2}\mu_i^2.
\]
\end{lemma}

\begin{proof}
Under Assumption~\ref{ass:orthogonal-gsvd}, write $\alpha=\sum_i t_i z_i$ and $\mu_{\text{base}}=\sum_i\mu_i u_i+\mu_\perp$. Orthogonality gives
\[
\|A\alpha-\mu_{\text{base}}\|_2^2
+
\lambda\|C\alpha\|_2^2
+
\rho\|\alpha\|_2^2
=
\|\mu_\perp\|_2^2+
\sum_i\left((a_it_i-\mu_i)^2+(\lambda c_i^2+\rho)t_i^2\right).
\]
Minimizing the one-dimensional quadratic in $t_i$ yields
\[
t_i=\frac{a_i\mu_i}{a_i^2+\lambda c_i^2+\rho}.
\]
Substituting this value gives the displayed error decomposition.
\end{proof}

\begin{remark}[Boundary cases]
\label{thm:asymmetric-failure-conditions}
The GSVD theorem identifies the only hard failures. If $\mu_{\text{base}}\perp A\operatorname{Null}(C)$, every exactly donor-stealthy coefficient vector satisfies $\|A\alpha-\mu_{\text{base}}\|_2\ge \|\mu_{\text{base}}\|_2$. If $\dim\operatorname{Null}(C)=0$, donor stealth forces $\alpha=0$, so the same lower bound holds with equality. These statements are boundary readings of Theorem~\ref{thm:exact-asymmetric-realizability}, not separate feasibility principles.
\end{remark}

\noindent\textbf{Intuition.}
The donor penalty removes GSVD coefficient directions with $c_i>0$. Exact asymmetric realization is possible precisely when the target lives in the base image of the remaining directions $c_i=0$. When exact stealth is impossible, the unavoidable base error is the energy of $\mu_{\text{base}}$ outside those donor-null directions, and the construction above gives the attacker coefficients that attain it.

\noindent\textbf{The base target $\mu_{\text{base}}$ is a presentation choice.}
Throughout the paper we instantiate the base-side target $\mu_{\text{base}}$ as the empirical mean of the base model's last-layer hidden states on a public WikiText sample (Section~\ref{subsec:pipeline}). This is a concrete, reproducible choice for the demonstrations and gives the breaker a generic, prompt-invariant high-salience direction. The realizability statement above does not depend on this specific target: Theorem~\ref{thm:exact-asymmetric-realizability} characterizes the entire feasibility cone of base-side directions admissible under donor stealth, and any direction in that cone supports the same construction. The centroid choice can be replaced with task-specific or token-specific base directions without altering the structural argument.

\section{A Wide $\lambda$-Window for Donor Suppression and Base Activation}
\label{app:lambda-window}

This section supports the $\lambda$-ablation claim documented in Appendix~\ref{app:lambda_window}: increasing the donor-suppression penalty can rapidly reduce donor activation while preserving base-side alignment. The statement is geometric; the conversion from geometry to SER is handled later in Theorem~\ref{thm:end-to-end-omp-attack}.

\begin{assumption}[Orthogonal generalized spectral coordinates]
\label{ass:orthogonal-gsvd}
For the fixed support $S$, let $A$ and $C$ be as in Assumption~\ref{ass:fixed-support-geometry}. Assume there exist orthonormal vectors $z_i\in\mathbb{R}^{k}$, orthonormal base-side vectors $u_i\in\mathbb{R}^{\delta_b}$, and orthonormal donor-side vectors $v_i\in\mathbb{R}^{m}$, with scalars $a_i,c_i\ge 0$, such that
\[
Az_i=a_i u_i,
\qquad
Cz_i=c_i v_i,
\qquad
a_i^2+c_i^2=1.
\]
For $c_i>0$, define $\sigma_i:=a_i/c_i$; if $c_i=0<a_i$, set $\sigma_i=+\infty$. Order the generalized singular values in non-increasing order, with the $+\infty$ directions placed first and the finite values ordered afterward.
\end{assumption}

\begin{theorem}[Spectral $\lambda$-window for simultaneous base alignment and donor suppression]
\label{thm:wide-lambda-window}
Assume Assumption~\ref{ass:orthogonal-gsvd}. Write
\[
\mu_{\text{base}}=\sum_i \mu_i u_i+\mu_\perp,
\qquad
\mu_i:=\langle \mu_{\text{base}},u_i\rangle,
\qquad
\mu_\perp\perp\operatorname{span}\{u_i\}_i.
\]
For $\lambda\ge 0$ and $\rho\ge 0$, let $\alpha_\lambda$ be the solution in Lemma~\ref{thm:lambda-regularized-realizability}. Then:
\begin{enumerate}[leftmargin=*]
\item The donor-visible norm obeys
\[
\|C\alpha_\lambda\|_2
\le
\frac{K_C}{\lambda}
\qquad
\text{for all }\lambda>0,
\]
where
\[
K_C:=\left(\sum_{i:c_i>0}\frac{a_i^2\mu_i^2}{c_i^2}\right)^{1/2}
=
\left(\sum_{i:c_i>0}\sigma_i^2\mu_i^2\right)^{1/2}.
\]
\item Fix a retained index set $\{1,\ldots,r\}$ among the ordered generalized singular directions and define
\[
E_r^2:=\|\mu_\perp\|_2^2+
\sum_{i>r}\mu_i^2,
\qquad
M_r^2:=\sum_{i=1}^{r}\mu_i^2.
\]
Assume $M_r>0$. Let $\varepsilon_{\text{base}}>E_r$ and set
\[
\Delta_r:=\left(\frac{\varepsilon_{\text{base}}^2-E_r^2}{M_r^2}\right)^{1/2},
\qquad
0<\Delta_r<1.
\]
For retained exact donor-stealth directions with $c_i=0$, assume
\[
\frac{\rho}{1+\rho}\le \Delta_r.
\]
For a donor target $\varepsilon_{\text{donor}}>0$, define
\[
\lambda_{\min}:=\frac{K_C}{\varepsilon_{\text{donor}}}
\]
and
\[
\lambda_{\max}:=
\min_{\substack{1\le i\le r\\ c_i>0}}
\left\{
\frac{\Delta_r}{1-\Delta_r}\sigma_i^2-\frac{\rho}{c_i^2}
\right\},
\]
with $\lambda_{\max}=+\infty$ if all retained directions have $c_i=0$. If $\lambda_{\min}\le \lambda\le \lambda_{\max}$, then
\[
\|C\alpha_\lambda\|_2\le \varepsilon_{\text{donor}},
\qquad
\|A\alpha_\lambda-\mu_{\text{base}}\|_2\le \varepsilon_{\text{base}}.
\]
\item Let
\[
g_r:=\frac{\sigma_r}{\sigma_{r+1}}
\]
for finite $\sigma_r,\sigma_{r+1}$, and let $E_{r,\sigma}^2:=\sum_{i>r}\mu_i^2$. Since
\[
K_C
\le
\sigma_r\left(\frac{\sigma_1}{\sigma_r}M_r+\frac{1}{g_r}E_{r,\sigma}\right),
\]
whenever the interval is nonempty its certified width satisfies
\[
\lambda_{\max}-\lambda_{\min}
\ge
\frac{\Delta_r}{1-\Delta_r}\sigma_r^2
-
\max_{\substack{1\le i\le r\\ c_i>0}}\frac{\rho}{c_i^2}
-
\frac{\sigma_r}{\varepsilon_{\text{donor}}}
\left(\frac{\sigma_1}{\sigma_r}M_r+\frac{1}{g_r}E_{r,\sigma}\right).
\]
Thus the prediction is falsifiable from the anchor pair: retained base energy $M_r$, tolerated tail $E_r$, and the spectral gap $g_r$ determine whether a broad interval of $\lambda$ values should preserve base alignment while suppressing donor projection.
\end{enumerate}
\end{theorem}

\begin{proof}
The formula for $\alpha_\lambda$ follows from Lemma~\ref{thm:lambda-regularized-realizability}. For the donor norm,
\[
C\alpha_\lambda
=
\sum_i
\frac{a_ic_i\mu_i}{a_i^2+\lambda c_i^2+\rho}v_i.
\]
When $c_i=0$, the term vanishes. When $c_i>0$,
\[
\left|
\frac{a_ic_i\mu_i}{a_i^2+\lambda c_i^2+\rho}
\right|
\le
\frac{|a_i\mu_i|}{\lambda c_i}.
\]
Squaring and summing gives $\|C\alpha_\lambda\|_2\le K_C/\lambda$.

The base residual is
\[
A\alpha_\lambda-\mu_{\text{base}}
=
-
\sum_i
\frac{\lambda c_i^2+\rho}{a_i^2+\lambda c_i^2+\rho}\mu_i u_i
-
\mu_\perp.
\]
For $i\le r$ with $c_i>0$, the shrinkage factor is at most $\Delta_r$ precisely when
\[
\lambda
\le
\frac{\Delta_r}{1-\Delta_r}\frac{a_i^2}{c_i^2}-\frac{\rho}{c_i^2}
=
\frac{\Delta_r}{1-\Delta_r}\sigma_i^2-\frac{\rho}{c_i^2}.
\]
For $i\le r$ with $c_i=0$, Assumption~\ref{ass:orthogonal-gsvd} gives $a_i=1$, and the shrinkage factor equals $\rho/(1+\rho)$, which is at most $\Delta_r$ by the theorem hypothesis. For $i>r$, the shrinkage factor is at most $1$. Therefore $\lambda\le\lambda_{\max}$ implies
\[
\|A\alpha_\lambda-\mu_{\text{base}}\|_2^2
\le
\Delta_r^2M_r^2+E_r^2
=
\varepsilon_{\text{base}}^2.
\]
If $\lambda\ge\lambda_{\min}=K_C/\varepsilon_{\text{donor}}$, the donor norm is at most $\varepsilon_{\text{donor}}$. This proves the window.

For the width bound, use $\sigma_i\ge\sigma_r$ for $i\le r$ to lower bound $\lambda_{\max}$ and split $K_C$ into retained and discarded parts:
\[
K_C^2
=
\sum_{i\le r}\sigma_i^2\mu_i^2+
\sum_{i>r}\sigma_i^2\mu_i^2
\le
\sigma_1^2M_r^2+\sigma_{r+1}^2E_{r,\sigma}^2.
\]
The displayed upper bound on $K_C$ follows from $\sqrt{x^2+y^2}\le x+y$ and $\sigma_{r+1}=\sigma_r/g_r$. Subtracting $\lambda_{\min}$ from the lower bound on $\lambda_{\max}$ completes the proof.
\end{proof}

\noindent\textbf{Intuition.}
Large generalized singular values are directions that the base sees much more strongly than the donor. Increasing $\lambda$ suppresses donor-visible components at rate $1/\lambda$, while high-$\sigma_i$ base components remain accurate until $\lambda$ reaches their squared spectral scale. The experimentally observed window $\lambda\in[768,1536]$ is the regime where the measured pair has a large retained spectral scale and a small discarded tail.

\section{From Base Reconstruction to Sequence Emission Rate}
\label{app:base-reconstruction-to-ser}

This section connects geometric reconstruction error to the SER metric of Section~\ref{sec:methods}. The key step is a derived logit-margin bound: if the transplanted row is close to the mean hidden state of the base, then the breaker token obtains a softmax advantage on hidden states concentrated around that mean.

\begin{assumption}[Measurable hidden-state and competitor-row controls]
\label{ass:hidden-norm-control}
Let $h_t$ be the last-layer hidden state of the post-transplant base at a generation position $t$, under the prompt distribution and under histories that have not yet emitted $\tau$. Define
\[
\mu_h:=\mathbb{E}[h_t].
\]
Assume that for every evaluated step and every no-prior-$\tau$ history,
\[
\Pr\left[\|h_t-\mu_h\|_2\le \xi_b\mid \text{history}\right]
\ge
1-\zeta_b.
\]
Let $R_b:=\max_{v\neq\tau}\|w_v\|_2$, and let $V_b$ be the base vocabulary size after transplant. These are directly measurable from the post-transplant base model.
\end{assumption}

\begin{lemma}[Logit margin from base-row alignment]
\label{lem:logit-margin-from-alignment}
Under Assumption~\ref{ass:hidden-norm-control}, suppose the breaker LM-head row satisfies
\[
\|w_\tau-\mu_h\|_2\le r_b.
\]
On the event $\|h_t-\mu_h\|_2\le \xi_b$,
\[
w_\tau^\top h_t-\max_{v\neq\tau}w_v^\top h_t
\ge
m_b(r_b),
\]
where
\[
m_b(r_b)
:=
\|\mu_h\|_2^2
-r_b\|\mu_h\|_2
-(\|\mu_h\|_2+r_b)\xi_b
-R_b(\|\mu_h\|_2+\xi_b).
\]
Consequently, if $m_b(r_b)>0$, the conditional softmax probability of $\tau$ on this event is at least
\[
q_b(r_b)
:=
\frac{1}{1+(V_b-1)\exp[-m_b(r_b)]}.
\]
\end{lemma}

\begin{proof}
Write $w_\tau=\mu_h+e$ with $\|e\|_2\le r_b$ and $h_t=\mu_h+\xi$ with $\|\xi\|_2\le \xi_b$. Then
\[
w_\tau^\top h_t
=
\|\mu_h\|_2^2+e^\top\mu_h+\mu_h^\top\xi+e^\top\xi
\ge
\|\mu_h\|_2^2-r_b\|\mu_h\|_2-(\|\mu_h\|_2+r_b)\xi_b.
\]
For every competitor $v\neq\tau$,
\[
w_v^\top h_t
\le
\|w_v\|_2\|h_t\|_2
\le
R_b(\|\mu_h\|_2+\xi_b).
\]
Subtracting yields the margin. If the breaker logit exceeds every competitor by $m$, then
\[
\operatorname{softmax}(Wh_t)_\tau
=
\frac{1}{1+\sum_{v\neq\tau}\exp(w_v^\top h_t-w_\tau^\top h_t)}
\ge
\frac{1}{1+(V_b-1)e^{-m}}.
\]
Setting $m=m_b(r_b)$ proves the result.
\end{proof}

\begin{remark}[SER compounding step]
\label{rem:ser-compounding-step}
If, at every step and conditional on any history that has not yet emitted $\tau$, the probability of sampling $\tau$ is at least $q$, then the probability that $\tau$ appears at least once in a $T$-token continuation is at least $1-(1-q)^T$. In Theorem~\ref{thm:end-to-end-omp-attack}, Lemma~\ref{lem:logit-margin-from-alignment} supplies $q=(1-\zeta_b)q_b(r_b)$ for the attacked base. The experiments use $T=256$, temperature $1.0$, and top-$p=0.9$; the same bound applies to top-$p$ sampling when $\tau$ remains in the retained nucleus on the margin event.
\end{remark}

\section{Operator-Specific Reachability for Differentiable Transplants}
\label{app:operator-specific-reachability}

This section refines the nonlinear-solver discussion in Appendix~\ref{app:nonlinear-shared-anchor}. OMP is handled by Theorem~\ref{thm:omp-coefficient-recovery}; here we prove that FOCUS softmax, top-$k$ mixtures, CLP, and WECHSEL each contain target-weight neighborhoods whenever their donor score maps can fit the corresponding local logits or positive scores.

\begin{theorem}[Shared-basis operator-class reachability]
\label{thm:operator-class-reachability}
Let $g_d:\mathbb{R}^{\delta_d}\to\mathcal{W}\subseteq\mathbb{R}^{N}$ be a donor-side coefficient map and let the base reconstruction be $x_b(x_d)=\Phi_b^\top g_d(x_d)$. Fix $w^\star\in\mathcal{W}$ and $\eta\ge 0$. If there exists a donor row $x_d$ such that
\[
\|g_d(x_d)-w^\star\|_1\le \eta,
\]
then
\[
\|x_b(x_d)-\Phi_b^\top w^\star\|_2
\le
\max_{j\in\mathcal{T}}\|\phi_j^{(b)}\|_2\eta.
\]
If instead the coefficient error is measured in $\ell_2$, then
\[
\|x_b(x_d)-\Phi_b^\top w^\star\|_2
\le
\|\Phi_b\|_{\mathrm{op}}\|g_d(x_d)-w^\star\|_2.
\]
\end{theorem}

\begin{proof}
The $\ell_1$ bound follows from
\[
\|\Phi_b^\top(g_d(x_d)-w^\star)\|_2
\le
\sum_{j\in\mathcal{T}}|g_{d,j}(x_d)-w_j^\star|\,\|\phi_j^{(b)}\|_2
\le
\max_{j\in\mathcal{T}}\|\phi_j^{(b)}\|_2\eta.
\]
The $\ell_2$ bound is the operator-norm inequality.
\end{proof}

\begin{assumption}[Softmax attention operator]
\label{ass:softmax-attention-operator}
The donor anchors and donor row are unit-normalized:
\[
\widehat{\phi}^{(d)}_j=\frac{\phi^{(d)}_j}{\|\phi^{(d)}_j\|_2},
\qquad
\widehat{x}_d=\frac{x_d}{\|x_d\|_2}.
\]
For an inverse temperature $\beta_{\mathrm{sm}}>0$, the FOCUS coefficient map is
\[
w_j(x_d)=\operatorname{softmax}_j\left(\beta_{\mathrm{sm}}\langle \widehat{x}_d,\widehat{\phi}^{(d)}_j\rangle\right).
\]
Equivalently, with normalized rows,
\[
w(x_d)=\operatorname{softmax}(\beta_{\mathrm{sm}}\Phi_d\widehat{x}_d).
\]
\end{assumption}

\begin{lemma}[FOCUS softmax reachability]
\label{lem:softmax-reachability}
Let $w^\star\in\mathbb{R}^{N}$ lie in the open simplex, and set $w^\star_{\min}:=\min_j w^\star_j>0$. For any $c\in\mathbb{R}$ define
\[
\ell^\star(c):=\frac{\log w^\star+c\mathbf{1}}{\beta_{\mathrm{sm}}}.
\]
If there exists a unit vector $z\in\mathbb{R}^{\delta_d}$ such that
\[
\|\Phi_d z-\ell^\star(c)\|_2\le \eta_{\mathrm{fit}},
\]
then any nonzero $x_d$ with $\widehat{x}_d=z$ satisfies
\[
\|\operatorname{softmax}(\beta_{\mathrm{sm}}\Phi_d\widehat{x}_d)-w^\star\|_1
\le
\frac{\beta_{\mathrm{sm}}\sqrt{N}}{2}\eta_{\mathrm{fit}}.
\]
The role of $w^\star_{\min}$ is to ensure finite target logits; the Lipschitz constant itself is uniform over the simplex.
\end{lemma}

\begin{proof}
Softmax is invariant to adding $c\mathbf{1}$ to the logits, so $\operatorname{softmax}(\log w^\star+c\mathbf{1})=w^\star$. The Jacobian of softmax at probability vector $\pi$ is $\operatorname{diag}(\pi)-\pi\pi^\top$, the covariance matrix of a categorical variable, and its $\ell_2$ operator norm is at most $1/2$. Therefore
\[
\|\operatorname{softmax}(s)-\operatorname{softmax}(s')\|_2
\le
\frac{1}{2}\|s-s'\|_2.
\]
Using $\|u\|_1\le\sqrt{N}\|u\|_2$ with $s=\beta_{\mathrm{sm}}\Phi_d z$ and $s'=\log w^\star+c\mathbf{1}$ gives the bound.
\end{proof}

\begin{assumption}[$k$-NN / top-$k$ softmax operator]
\label{ass:knn-topk-operator}
The operator first selects
\[
\mathcal{N}_k(x_d)=\operatorname{TopK}_{j\in\mathcal{T}}\langle \phi_j^{(d)},x_d\rangle
\]
and then applies a softmax rule restricted to $\mathcal{N}_k(x_d)$. The target support $S\subseteq\mathcal{T}$ has $|S|=k$.
\end{assumption}

\begin{lemma}[Margin-stable support selection]
\label{lem:margin-stable-knn}
Under Assumption~\ref{ass:knn-topk-operator}, suppose $x_d$ satisfies
\[
\min_{j\in S}\langle \phi_j^{(d)},x_d\rangle
>
\max_{\ell\notin S}\langle \phi_\ell^{(d)},x_d\rangle+\gamma
\]
for some $\gamma>0$. Then $\mathcal{N}_k(x_d)=S$. Moreover, if $x_d'$ satisfies
\[
\max_{j\in\mathcal{T}}
|\langle \phi_j^{(d)},x_d'-x_d\rangle|
\le
\frac{\gamma}{2},
\]
then $\mathcal{N}_k(x_d')=S$ as well.
\end{lemma}

\begin{proof}
The first claim follows because every score indexed by $S$ exceeds every score outside $S$, and $|S|=k$. For the perturbation claim, for $j\in S$ and $\ell\notin S$,
\[
\langle \phi_j^{(d)},x_d'\rangle
\ge
\langle \phi_j^{(d)},x_d\rangle-\frac{\gamma}{2}
>
\langle \phi_\ell^{(d)},x_d\rangle+\frac{\gamma}{2}
\ge
\langle \phi_\ell^{(d)},x_d'\rangle.
\]
Thus the top-$k$ set remains $S$.
\end{proof}

\begin{corollary}[Top-$k$ mixture reachability]
\label{cor:topk-mixture-reachability}
Let $w^\star$ be supported on $S$ and lie in the open simplex over $S$. If $x_d$ satisfies the margin condition in Lemma~\ref{lem:margin-stable-knn} and the restricted softmax logits over $S$ satisfy
\[
\left\|
\Phi_{d,S}\widehat{x}_d
-
\frac{\log w^\star_S+c\mathbf{1}_S}{\beta_{\mathrm{sm}}}
\right\|_2
\le
\eta_{\mathrm{fit},S},
\]
then the full top-$k$ mixture weights satisfy
\[
\|w(x_d)-w^\star\|_1
\le
\frac{\beta_{\mathrm{sm}}\sqrt{k}}{2}\eta_{\mathrm{fit},S}.
\]
\end{corollary}

\begin{proof}
By Lemma~\ref{lem:margin-stable-knn}, the selected support is $S$. Outside $S$, both the target and recovered weights are zero. Inside $S$, apply Lemma~\ref{lem:softmax-reachability} with $N$ replaced by $k$.
\end{proof}

\begin{assumption}[CLP ReLU-renormalized coefficient map]
\label{ass:clp-operator}
For normalized donor row $\widehat{x}_d$, define scores
\[
s_j(x_d):=\langle \widehat{x}_d,\widehat{\phi}^{(d)}_j\rangle.
\]
The CLP coefficient map is
\[
w_j(x_d)=
\frac{\operatorname{ReLU}(s_j(x_d))}{\epsilon_{\mathrm{clp}}+\sum_{\ell\in\mathcal{T}}\operatorname{ReLU}(s_\ell(x_d))},
\]
with $\epsilon_{\mathrm{clp}}\ge 0$.
\end{assumption}

\begin{lemma}[CLP reachability through positive score fitting]
\label{lem:clp-reachability}
Let $w^\star$ lie in the simplex over a support $S$. Suppose there are $a>0$ and $x_d$ such that, with
\[
r_j:=\operatorname{ReLU}(s_j(x_d)),
\]
the positive score vector satisfies
\[
\|r-a w^\star\|_1\le \eta_{\mathrm{clp}},
\qquad
0\le \eta_{\mathrm{clp}}<a.
\]
Then the CLP weights satisfy
\[
\|w(x_d)-w^\star\|_1
\le
\frac{2\eta_{\mathrm{clp}}}{a-\eta_{\mathrm{clp}}}
+
\frac{\epsilon_{\mathrm{clp}}}{\epsilon_{\mathrm{clp}}+a-\eta_{\mathrm{clp}}}.
\]
\end{lemma}

\begin{proof}
Let $R:=\sum_j r_j$. Since $\sum_j w_j^\star=1$,
\[
|R-a|\le \|r-aw^\star\|_1\le \eta_{\mathrm{clp}},
\]
so $R\ge a-\eta_{\mathrm{clp}}>0$. Decompose
\[
\left\|\frac{r}{\epsilon_{\mathrm{clp}}+R}-w^\star\right\|_1
\le
\left\|\frac{r}{\epsilon_{\mathrm{clp}}+R}-\frac{r}{R}\right\|_1
+
\left\|\frac{r}{R}-w^\star\right\|_1.
\]
The first term equals
\[
\frac{\epsilon_{\mathrm{clp}}}{\epsilon_{\mathrm{clp}}+R}
\le
\frac{\epsilon_{\mathrm{clp}}}{\epsilon_{\mathrm{clp}}+a-\eta_{\mathrm{clp}}}.
\]
For the second term,
\[
\frac{r}{R}-w^\star
=
\frac{r-aw^\star}{R}+
\left(\frac{a}{R}-1\right)w^\star,
\]
so
\[
\left\|\frac{r}{R}-w^\star\right\|_1
\le
\frac{\eta_{\mathrm{clp}}}{R}+\frac{|a-R|}{R}
\le
\frac{2\eta_{\mathrm{clp}}}{a-\eta_{\mathrm{clp}}}.
\]
\end{proof}

\begin{assumption}[WECHSEL affine alignment]
\label{ass:wechsel-affine}
The WECHSEL operator uses an affine alignment
\[
\widetilde{x}(x_d)=A_{\mathrm{WEC}}(x_d-\mu_{\mathrm{src}})+\mu_{\mathrm{tgt}},
\]
followed by local $k$-NN softmax interpolation:
\[
f_{\mathrm{WEC}}(x_d)=
\sum_{v\in\mathcal{N}_k(\widetilde{x}(x_d))}
w_v(\widetilde{x}(x_d))\phi_v^{(b)}.
\]
Assume $A_{\mathrm{WEC}}$ has a right inverse on the target affine subspace: for every target $z$ in that subspace,
\[
x_d(z)=\mu_{\mathrm{src}}+A_{\mathrm{WEC}}^\dagger(z-\mu_{\mathrm{tgt}})
\]
satisfies $\widetilde{x}(x_d(z))=z$.
\end{assumption}

\begin{lemma}[WECHSEL reachability decomposition]
\label{lem:wechsel-reachability}
Under Assumption~\ref{ass:wechsel-affine},
\[
\|\widetilde{x}(x_d)-\widetilde{x}(x_d')\|_2
\le
\|A_{\mathrm{WEC}}\|_{\mathrm{op}}\|x_d-x_d'\|_2,
\]
and, for reachable targets $z,z'$,
\[
\|x_d(z)-x_d(z')\|_2
\le
\|A_{\mathrm{WEC}}^\dagger\|_{\mathrm{op}}\|z-z'\|_2.
\]
Consequently, any target aligned point $z$ satisfying the support margin condition of Lemma~\ref{lem:margin-stable-knn} and the restricted softmax fitting condition of Corollary~\ref{cor:topk-mixture-reachability} is realized by $x_d(z)$. The reachable WECHSEL weight family is the inclusion
\[
\mathcal{W}_{\mathrm{WEC}}
\supseteq
\bigcup_{\substack{S\subseteq\mathcal{T}\\ |S|=k}}
\left\{
 w\in\Delta_S:
 \exists z\in\operatorname{Im}(\widetilde{x})\text{ satisfying the }S\text{-margin condition and the }S\text{-softmax fit for }w
\right\}.
\]
If WECHSEL selects supports only by this margin rule and weights only by the stated local softmax, the inclusion is equality up to score ties.
\end{lemma}

\begin{proof}
For any $x_d,x_d'$,
\[
\widetilde{x}(x_d)-\widetilde{x}(x_d')=A_{\mathrm{WEC}}(x_d-x_d'),
\]
which gives the first Lipschitz bound. For reachable targets,
\[
x_d(z)-x_d(z')=A_{\mathrm{WEC}}^\dagger(z-z'),
\]
which gives the second bound. If $z$ satisfies the margin and fitting conditions, $x_d(z)$ maps to $z$, the selected support is $S$ by Lemma~\ref{lem:margin-stable-knn}, and the weights on $S$ are controlled by Corollary~\ref{cor:topk-mixture-reachability}. The displayed family follows.
\end{proof}

\noindent\textbf{Intuition.}
All differentiable operators have the same core: fit an operator-specific donor score vector, obtain weights, and reuse those weights on base anchors. FOCUS fits log-probabilities, top-$k$ methods add a margin-stable support, CLP fits positive ReLU scores, and WECHSEL first moves the donor row through an affine alignment before the same local-mixture argument applies.

\section{A Constructive Lower Bound for Single-Statistic Spectral Detectors}
\label{app:spectral-detector-lower-bound}

This section formalizes the spectral-mimicry claim in Section~\ref{sec:mimicry}. The result applies to any detector that thresholds one scalar donor-row statistic, once the feasible design set contains a controllable direction for that statistic. The proof has two parts: the base-alignment and donor-suppression constraints leave an affine design set in which the attacker can match the natural pushforward of common scalar statistics; once the pushforward distributions are close, a single-statistic threshold cannot catch attacker rows without flagging natural tokens.

\begin{assumption}[Single-statistic threshold detector]
\label{ass:single-statistic-threshold}
Let $s:\mathbb{R}^{\delta_d}\to\mathbb{R}$ be a fixed scalar statistic of a donor embedding row. A detector is a measurable threshold rule
\[
D(x)=\mathbf{1}\{s(x)\in B\},
\]
where $B\subseteq\mathbb{R}$ is the rejection region. Let $\nu_{\mathrm{nat}}$ be the empirical distribution of $s(x)$ over natural donor tokens.
\end{assumption}

\begin{assumption}[Design freedom after base and donor constraints]
\label{ass:detector-design-freedom}
Let
\[
\mathcal{F}_{\alpha}
:=
\{\alpha:A\alpha=\mu_{\text{base}},\ C\alpha=0\}
\]
be nonempty, and fix one solution $\alpha_0$. Then
\[
\mathcal{F}_{\alpha}=\alpha_0+\operatorname{Null}\left(\begin{bmatrix}A\\ C\end{bmatrix}\right).
\]
Equivalently, the donor rows available without changing the exact base image or donor projection contain the affine set
\[
\mathcal{F}_{x}
:=
x_0+L,
\qquad
x_0:=\Phi_{d,S}^\top\alpha_0,
\qquad
L:=\Phi_{d,S}^\top\operatorname{Null}\left(\begin{bmatrix}A\\ C\end{bmatrix}\right).
\]
Let $d_f:=\dim L$. For scalar base constraints such as a fixed alignment $\langle A\alpha,\mu_{\text{base}}\rangle$, the same calculation gives $d_f\ge \dim\operatorname{Null}(C)-1$ before accounting for degeneracy of $\Phi_{d,S}^\top$; for exact vector constraints, $d_f\ge \dim\operatorname{Null}(C)-\operatorname{rank}(A|_{\operatorname{Null}(C)})$ before the same donor-map quotient.
\end{assumption}

\begin{assumption}[Statistic controllability on the design set]
\label{ass:statistic-controllability}
There is an interval $I\subseteq\mathbb{R}$ and a measurable map $\gamma:I\to\mathcal{F}_x$ such that
\[
s(\gamma(t))=t
\qquad
\text{for every }t\in I.
\]
Let $\epsilon_{\mathrm{tail}}:=\nu_{\mathrm{nat}}(I^c)$.
\end{assumption}

\begin{theorem}[Constructive mimicry and threshold-detector lower bound]
\label{thm:spectral-threshold-lower-bound}
Under Assumptions~\ref{ass:single-statistic-threshold}--\ref{ass:statistic-controllability}, there is a distribution $\nu_{\mathrm{atk}}$ over attacker-designed donor rows in $\mathcal{F}_x$ such that
\[
\operatorname{TV}\left(s_\#\nu_{\mathrm{atk}},\nu_{\mathrm{nat}}\right)
\le
\epsilon_{\mathrm{tail}}.
\]
Consequently, for any threshold detector $D(x)=\mathbf{1}\{s(x)\in B\}$, if
\[
\operatorname{FNR}_{\mathrm{atk}}:=s_\#\nu_{\mathrm{atk}}(B^c)\le \delta,
\]
then
\[
\operatorname{FPR}_{\mathrm{nat}}:=\nu_{\mathrm{nat}}(B)
\ge
1-\delta-\epsilon_{\mathrm{tail}}.
\]
In particular, when $\delta\le 1/2$, $\operatorname{FPR}_{\mathrm{nat}}\ge \delta-\epsilon_{\mathrm{tail}}$.
\end{theorem}

\begin{proof}
Let $\bar\nu$ be the conditional distribution of the natural statistic given $I$:
\[
\bar\nu(E)=\frac{\nu_{\mathrm{nat}}(E\cap I)}{\nu_{\mathrm{nat}}(I)}.
\]
Draw $T\sim\bar\nu$ and output the attacker row $X=\gamma(T)$. By Assumption~\ref{ass:statistic-controllability}, $s(X)=T$, so $s_\#\nu_{\mathrm{atk}}=\bar\nu$. The total variation distance between a distribution and its conditioning on an event of mass $1-\epsilon_{\mathrm{tail}}$ is at most $\epsilon_{\mathrm{tail}}$, proving the constructive TV bound.

For the detector lower bound, $\operatorname{FNR}_{\mathrm{atk}}\le\delta$ implies $s_\#\nu_{\mathrm{atk}}(B)\ge 1-\delta$. Total variation gives
\[
\nu_{\mathrm{nat}}(B)
\ge
s_\#\nu_{\mathrm{atk}}(B)-\epsilon_{\mathrm{tail}}
\ge
1-\delta-\epsilon_{\mathrm{tail}}.
\]
The final weaker bound follows from $1-\delta\ge\delta$ when $\delta\le 1/2$.
\end{proof}

\begin{corollary}[Controllability for common donor-row statistics]
\label{cor:common-statistic-controllability}
Assume $d_f\ge 1$ and the feasible row set contains $x_0+tq$ for $|t|\le R$, where $q\in L$ is a unit vector.
\begin{enumerate}[leftmargin=*]
\item For the row-norm statistic $s(x)=\|x\|_2$, if $q\perp x_0$, then $s(x_0+tq)$ covers $[\|x_0\|_2,\sqrt{\|x_0\|_2^2+R^2}]$.
\item For the donor PCA residual statistic $s_U(x)=\|(I-U^\top U)x\|_2$, if $q\in\operatorname{Null}(U)$ and $q\perp (I-U^\top U)x_0$, then $s_U(x_0+tq)$ covers $[\|(I-U^\top U)x_0\|_2,\sqrt{\|(I-U^\top U)x_0\|_2^2+R^2}]$.
\item More generally, any Lipschitz-1 scalar statistic $s$ satisfying $s(x_0+t_-q)\le a$ and $s(x_0+t_+q)\ge b$ for some $t_-,t_+\in[-R,R]$ is controllable on $I=[a,b]$ by the intermediate value theorem after restricting to a monotone measurable branch.
\end{enumerate}
Thus norm detectors, PCA-residual detectors of the Magikarp type \citep{land2024fishingmagikarpautomaticallydetecting}, and any scalar Lipschitz statistic with a controllable direction fall under Theorem~\ref{thm:spectral-threshold-lower-bound}.
\end{corollary}

\begin{proof}
The first two claims follow by direct calculation:
\[
\|x_0+tq\|_2^2=\|x_0\|_2^2+t^2
\]
when $q\perp x_0$, and similarly
\[
\|(I-U^\top U)(x_0+tq)\|_2^2
=
\|(I-U^\top U)x_0\|_2^2+t^2
\]
when $q\in\operatorname{Null}(U)$ and $q\perp (I-U^\top U)x_0$. The third claim uses continuity of Lipschitz functions on the line segment and the intermediate value theorem.
\end{proof}

\begin{remark}[Finite multi-statistic extension]
\label{rem:multi-statistic-detectors}
For a vector statistic $s=(s_1,\ldots,s_p)$, the same proof applies with $I\subseteq\mathbb{R}^p$ whenever the feasible design set admits a measurable right inverse $\gamma:I\to\mathcal{F}_x$ with $s(\gamma(t))=t$. A sufficient local condition is that $s$ is continuously differentiable and its Jacobian restricted to $L$ has rank $p$ at some feasible row; the inverse function theorem then gives controllability on a neighborhood. The TV lower bound is unchanged except for the natural mass outside that controllable neighborhood.
\end{remark}

\noindent\textbf{Intuition.}
After base alignment and donor suppression are fixed, the attacker often retains an affine set of coefficient perturbations that do not change either constraint. If a detector only reads one scalar statistic, one controllable direction in this affine set is enough to match the natural statistic distribution on its central support; thresholding that statistic then trades false negatives directly against false positives.

\section{Calibration-Based Recipient-Side SER Bound}
\label{app:recipient-side-defense-ser}

This section turns the differential-SER auditing protocol in Appendix~\ref{app:auditing} into a sufficient condition for bounded sampling SER. It applies to recipient-side checks that reject newly synthesized rows whose calibration probability is too often above a chosen per-step threshold.

\begin{assumption}[Calibration transfer for synthesized rows]
\label{ass:calibration-transfer}
For a newly synthesized embedding or LM-head row $w_\tau$, define
\[
\widehat{q}_\tau
:=
\Pr_{p\sim\mathcal{P}_{\mathrm{cal}}}
\left[
\operatorname{softmax}(Wh_p)_\tau\ge q^\star
\right].
\]
Assume the deployment prompt-position distribution is dominated by the calibration distribution in the following measurable sense: if $\widehat{q}_\tau<\delta$, then for every generation position $t\in\{1,\ldots,T\}$,
\[
\Pr\left[
\operatorname{softmax}(Wh_t)_\tau\ge q^\star
\right]
\le
\delta,
\]
where the probability is over deployment prompts and decoding histories before position $t$.
\end{assumption}

\begin{theorem}[Calibration-based SER bound]
\label{thm:calibration-ser-bound}
Suppose the recipient-side defense rejects $w_\tau$ whenever $\widehat{q}_\tau\ge\delta$. Under Assumption~\ref{ass:calibration-transfer}, any token $\tau$ that passes the check satisfies
\[
\operatorname{SER}(\tau)
\le
1-(1-q^\star)^T+
\delta T
\]
under temperature-$1$ sampling, including top-$p$ sampling when the post-truncation probability of $\tau$ is no larger than its checked probability on the non-tail event.
\end{theorem}

\begin{proof}
Let
\[
B_t:=\left\{\operatorname{softmax}(Wh_t)_\tau\ge q^\star\right\}.
\]
Since the token passes the check, $\widehat{q}_\tau<\delta$, and Assumption~\ref{ass:calibration-transfer} gives $\Pr(B_t)\le\delta$ for every $t$. By the union bound,
\[
\Pr\left(\bigcup_{t=1}^{T}B_t\right)
\le
\delta T.
\]
On the complement of $\bigcup_t B_t$, the probability of sampling $\tau$ at every position is at most $q^\star$. Iterating the conditional probabilities along the generation history gives probability at least $(1-q^\star)^T$ of never sampling $\tau$ on this non-tail event. Hence, conditional on no tail event, the probability of at least one occurrence is at most $1-(1-q^\star)^T$. Adding the tail probability gives the bound.
\end{proof}

\noindent\textbf{Intuition.}
The audit bounds the mass of high-probability positions for a synthesized row. If a row passes, ordinary positions contribute at most $1-(1-q^\star)^T$ to SER, and rare calibration-tail positions contribute at most $\delta T$.

\section{Sparse Asymmetric Coefficient Attack under OMP}
\label{app:omp-end-to-end-main}

This section composes the preceding results into the end-to-end guarantee used by the main attack chain. The constants are measurable properties of the selected anchors, the designed row, and the base/donor hidden-state distributions; no black-box SER calibration function is assumed.

\begin{assumption}[End-to-end sparse OMP attack setup]
\label{ass:end-to-end-omp-setup}
Fix a support $S\subseteq\mathcal{T}$ with $|S|=k$. The donor dictionary satisfies Assumption~\ref{ass:omp-coherence}, and the planted donor row satisfies Assumption~\ref{ass:omp-noisy-row}. Define
\[
\varepsilon_{\mathrm{rec}}
:=
\frac{\eta_\varepsilon}{\sigma_{\min}(\Phi_{d,S})}.
\]
Assume base alignment and donor innocuity:
\[
\|\Phi_{b,S}^\top\alpha-\mu_{\text{base}}\|_2\le \eta_b,
\qquad
\|U\Phi_{d,S}^\top\alpha\|_2\le \eta_d.
\]
Let
\[
M_d:=\|\Phi_{d,S}^\top\alpha\|_2+\eta_\varepsilon.
\]
\end{assumption}

\begin{assumption}[Base and donor hidden-state controls for SER]
\label{ass:end-to-end-hidden-state-controls}
For the attacked base, Assumption~\ref{ass:hidden-norm-control} holds with mean $\mu_h$, concentration radius $\xi_b$, failure probability $\zeta_b$, competitor row norm bound $R_b$, and vocabulary size $V_b$. Define
\[
\eta_h:=\|\mu_{\text{base}}-\mu_h\|_2.
\]
For the patched donor, let $h_t^{(d)}$ be the donor hidden state at a generation position under histories that have not yet emitted $\tau$. Let $P_{U^\perp}:=I-U^\top U$ after orthonormalizing the rows of $U$. Assume that, for every no-prior-$\tau$ history,
\[
\Pr\left[
\|Uh_t^{(d)}\|_2\le H_{d,U},\quad
\|P_{U^\perp}h_t^{(d)}\|_2\le H_{d,\perp},\quad
Z_d(h_t^{(d)})\ge Z_{d,0}
\mid \text{history}\right]
\ge
1-\zeta_d,
\]
where
\[
Z_d(h):=\sum_{v\in\mathcal{V}_d}\exp((w_v^{(d)})^\top h)
\]
is the natural donor logit partition excluding the breaker token. The constants $H_{d,U}$, $H_{d,\perp}$, $Z_{d,0}$, and $\zeta_d$ are measurable from the patched donor and the deployment prompt-position distribution.
\end{assumption}

\begin{theorem}[End-to-end sparse asymmetric coefficient attack under OMP]
\label{thm:end-to-end-omp-attack}
Under Assumptions~\ref{ass:end-to-end-omp-setup} and~\ref{ass:end-to-end-hidden-state-controls}, the victim's OMP-based transplant produces a base row $\widehat{x}_b=\Phi_b^\top\beta$ with the following properties.
\begin{enumerate}[leftmargin=*]
\item OMP recovers the support $S$ in exactly $k$ iterations, and
\[
\|\beta-\alpha\|_2\le \varepsilon_{\mathrm{rec}}.
\]
\item The transplanted base row satisfies
\[
\|\widehat{x}_b-\mu_{\text{base}}\|_2
\le
\eta_b+\|\Phi_{b,S}\|_{\mathrm{op}}\varepsilon_{\mathrm{rec}}.
\]
Consequently,
\[
\|\widehat{x}_b-\mu_h\|_2
\le
r_b
:=
\eta_h+\eta_b+\|\Phi_{b,S}\|_{\mathrm{op}}\varepsilon_{\mathrm{rec}}.
\]
\item The donor-side projection of the planted row satisfies
\[
\|Ux_d\|_2
\le
r_d
:=
\eta_d+\|U\|_{\mathrm{op}}\eta_\varepsilon.
\]
\item If $m_b(r_b)>0$, with $m_b$ from Lemma~\ref{lem:logit-margin-from-alignment}, define
\[
q_b
:=
\frac{1}{1+(V_b-1)\exp[-m_b(r_b)]}.
\]
Then the attacked-base length-$T$ sampling SER is lower-bounded by
\[
\operatorname{SER}_{\mathrm{base}}(\tau)
\ge
1-\left(1-(1-\zeta_b)q_b\right)^T.
\]
\item Define the donor breaker-logit upper bound
\[
L_d
:=
r_d H_{d,U}+M_d H_{d,\perp}
\]
and
\[
q_d
:=
\frac{\exp(L_d)}{Z_{d,0}+\exp(L_d)}.
\]
Then the patched-donor length-$T$ sampling SER is upper-bounded by
\[
\operatorname{SER}_{\mathrm{donor}}(\tau)
\le
1-(1-q_d)^T+\zeta_d T.
\]
The explicit asymmetry gap certified by the theorem is therefore at least
\[
\left[1-\left(1-(1-\zeta_b)q_b\right)^T\right]
-
\left[1-(1-q_d)^T+\zeta_d T\right].
\]
\end{enumerate}
\end{theorem}

\begin{proof}
The first claim is Theorem~\ref{thm:omp-coefficient-recovery}. The second claim is Corollary~\ref{cor:omp-base-reconstruction-error} with the design residual bounded by $\eta_b$ and $\|\Phi_{b,S}\|_{\mathrm{op}}$ used in place of $\|\Phi_b\|_{\mathrm{op}}$ on the recovered support. Adding $\|\mu_{\text{base}}-\mu_h\|_2=\eta_h$ gives the bound $r_b$.

For the donor projection,
\[
Ux_d=U\Phi_{d,S}^\top\alpha+U\varepsilon,
\]
so
\[
\|Ux_d\|_2
\le
\eta_d+\|U\|_{\mathrm{op}}\eta_\varepsilon
=r_d.
\]

For the base SER, Lemma~\ref{lem:logit-margin-from-alignment} applied with $w_\tau=\widehat{x}_b$ gives softmax probability at least $q_b$ on the concentration event, which has conditional probability at least $1-\zeta_b$ at each no-prior-$\tau$ step. Thus, conditional on not having emitted $\tau$ before step $t$, the probability of emitting $\tau$ at step $t$ is at least $(1-\zeta_b)q_b$. Iterating the probability of non-emission gives
\[
\Pr[\tau\text{ never appears in }T\text{ steps}]
\le
\left(1-(1-\zeta_b)q_b\right)^T,
\]
which proves the lower bound.

For the donor side, on the donor concentration event,
\[
x_d^\top h_t^{(d)}
=
(Ux_d)^\top Uh_t^{(d)}+(P_{U^\perp}x_d)^\top P_{U^\perp}h_t^{(d)}
\le
r_dH_{d,U}+\|x_d\|_2H_{d,\perp}.
\]
Since $\|x_d\|_2\le M_d$, this logit is at most $L_d$. The natural donor partition is at least $Z_{d,0}$ on the same event, so the breaker softmax probability is at most
\[
\frac{\exp(L_d)}{Z_{d,0}+\exp(L_d)}=q_d.
\]
Conditional on no tail event across the $T$ positions, the probability of at least one donor emission is at most $1-(1-q_d)^T$. The union bound gives probability at most $\zeta_d T$ for a tail event, proving the donor SER upper bound.
\end{proof}

\noindent\textbf{Intuition.}
The end-to-end chain has no hidden calibration oracle. Donor coherence and beta-min recover $\beta\approx\alpha$; conditioning of $\Phi_{d,S}$ controls the coefficient error; $\Phi_{b,S}$ passes this error to the base row; base hidden-state concentration converts row alignment into a softmax margin; and donor hidden-state concentration plus the partition lower bound converts donor innocuity into a per-step emission cap.

\section{Operational Realism: Research Artifacts and Deployment Footprints}
\label{app:operational_stealth}

This appendix separates artifacts preserved for reproducibility from properties of the geometric mechanism. Our experimental code modifies \texttt{added\_tokens.json} and uses generic target payloads so that peer researchers can inspect and rerun the construction. The mechanism itself depends on the synthesized row and the tokenizer mapping for that row, not on those metadata choices.

\subsection{Payload independence}
The breaker-token mechanism decouples the \textbf{geometric trigger} from the \textbf{surface form}. The optimization step controls the geometric vector $\mathbf{x} \in \mathbb{R}^d$ that survives transplant and activates a specific vocabulary index $\text{id}_{\tau}$, while the surface string assigned to $\text{id}_{\tau}$ is set by the tokenizer's mapping file. Section~\ref{subsec:fine-grained} demonstrates this end-to-end: each of the three threat scenarios from Section~\ref{sec:intro} (service degradation, reputation-poisoning contamination, latent watermark/signature) is realized at the chosen surface form in actual model output.

\subsection{Metadata-visible append vs. reserved-slot implementation}
Our experiments append tokens ($\mathcal{V}' = \mathcal{V} \cup \{\tau^*\}$) rather than overwriting existing reserved slots, making the row visible in the released research artifact.

\noindent\textbf{Token masquerading.}
Standard production tokenizers (e.g., Qwen, Llama-3) contain numerous ``reserved'' or ``unused'' slots. A deployment adversary could employ \textit{token masquerading}: silently repurposing these existing slots rather than appending new entries. This yields a zero-metadata footprint: the vocabulary size remains invariant and no auxiliary metadata files are generated, so file-based auditing alone would not detect it.

\noindent\textbf{Release posture.}
Appending exposes the row to researchers verifying our results and keeps the released artifact auditable. The metadata traces in our submission are a reproducibility choice, not a constraint on the geometric mechanism.

\section{Operational Safeguards: Post-Transplant Auditing Guidelines}
\label{app:auditing}

Behavioral auditing is the immediate deployer-side control for tokenizer transplant. Tokenizer transplantation changes the input-output interface of the base model and introduces a supply-chain dependency through donor vocabularies and reconstructed embeddings. Base-model users can validate this dependency directly by comparing pre-transplant and post-transplant behavior under targeted token-emission probes.
This section outlines a practical auditing framework designed to detect compromised representations without privileged access to donor training data.

\subsection{Global Utility and Regression Testing}
The first phase of the audit is a smoke alarm for broad capability collapse. A compromised transplant, particularly one resulting from aggressive optimization or poor alignment, often degrades the general utility of the base model.
Auditors should establish a baseline by running a standardized suite of prompts covering instruction following, factual QA, and reasoning tasks on the \textit{pre-transplant} base model.
Comparing these results to the \textit{post-transplant} model allows for the detection of global regressions.
Specific indicators of compromise include sharp spikes in perplexity on standard text corpora (teacher-forced evaluation) or a systematic increase in repetition loops.
Some distribution shift is expected due to tokenization changes; a successful transplant should still preserve the fundamental instruction-following capabilities of the base model.
Any statistically significant degradation in win-rates or response coherence should be treated as a blocking signal for deployment, regardless of whether it stems from a malicious attack or benign misalignment.

\subsection{Differential Token Analysis}
Targeted sabotage, such as the attack described in this work, often leaves a localized vocabulary footprint: magnet tokens that become unnaturally attractive during generation.
To detect these, we advocate for a \textit{differential audit} that compares token emission statistics before and after transplantation.
Auditors should compute SER across a diverse set of prompt contexts.
By calculating the delta $\Delta_{\text{SER}}(t) = \text{SER}_{\text{post}}(t) - \text{SER}_{\text{pre}}(t)$ for all vocabulary items, auditors can rank tokens by their behavioral shift.
A benign transplant typically spreads probability mass shifts diffusely across synonyms or related concepts.
In contrast, a malicious injection often manifests as a sparse anomaly where one token or a tight token cluster exhibits a massive surge in $\Delta_{\text{SER}}$, appearing in contexts where it is semantically unwarranted.
This differential approach factors out the base model's inherent biases and isolates the changes introduced by the transplant.

\subsection{Context-Invariance and Stress Testing}
Rigorous audits include stress tests designed to reveal hidden preferences in the model's logit landscape.
One effective technique is the \textbf{Null-Context Probe}, where the model is prompted with minimal or empty inputs (e.g., a single whitespace or common preamble).
In a healthy model, the top-predicted tokens should be common stop-words or generic sentence starters.
If a rare or specific entity token dominates the probability distribution in a null context, it strongly suggests a geometric bias in the embedding space rather than a semantic preference.
Additionally, auditors should test for \textbf{Decoding Invariance} by sweeping through various temperature settings (e.g., $T \in \{0.0, 0.7, 1.0\}$).
Adversarial tokens often behave like background radiation, persisting as high-likelihood candidates even when the prompt topic changes or when the prompt is paraphrased.
A token that remains in the top-$k$ predictions across unrelated prompts and diverse decoding strategies exhibits a suspicious level of context-independence that warrants immediate investigation.

\subsection{Triage Policy}
Upon detecting anomalies, organizations should adopt a tiered triage policy.
Minor shifts in utility or diffuse changes in token probability (Level 1) may be acceptable consequences of the transplant process, necessitating only a re-run with different hyperparameters.
Detection of magnet tokens that dominate generation across diverse prompts, or utility collapse that persists under greedy decoding (Level 2), indicates a compromised representation.
In such cases, the transplant artifacts must be quarantined.
Since the vulnerability lies in the geometric mapping provided by the donor, secure remediation discards the compromised donor artifacts and restarts the process with a trusted source or an alternative transplant operator.
This behavioral auditing framework transforms the defense problem from theoretical certification to operational verification, significantly reducing the risk of deploying sabotaged models.

\section{Experimental Settings}
\label{app:exp_settings}

\subsection{Settings}
We evaluate the suites introduced in \S\ref{subsec:exp-settings}, with cross-scale transfer split into upstream and downstream pair sets:
\begin{itemize}[topsep=0pt,itemsep=1pt,parsep=0pt,partopsep=0pt,leftmargin=*]
  \item \textbf{The Lightweight Clique ($\mathcal{C}_{\text{Light}}$).} All 20 directed pairs among Qwen2-0.5B \citep{yang2024qwen2}, Qwen3-0.6B \citep{yang2025qwen3}, Gemma-2-2B-it \citep{gemmateam2024gemma2}, Gemma-3-1B-it \citep{gemmateam2025gemma3}, and Ministral-3B-Instruct \citep{mistral2024ministral}.
  \item \textbf{The Standard-Scale Clique ($\mathcal{C}_{\text{Std}}$).} All 6 directed pairs among Gemma-2-9B-it \citep{gemmateam2024gemma2}, Llama-3-8B \citep{dubey2024llama3}, and Mistral-7B-v0.1 \citep{jiang2023mistral7b}.
  \item \textbf{Cross-Scale Transfer Sets ($\mathcal{T}_{\text{Cross}}^{\uparrow}$), upstream.} 16 directed pairs with \emph{large bases} receiving an embedding from \emph{small donors} (one model per family).
  \item \textbf{Cross-Scale Transfer Sets ($\mathcal{T}_{\text{Cross}}^{\downarrow}$), downstream.} 23 directed pairs with \emph{small bases} receiving an embedding from \emph{large donors} (one model per family).
\end{itemize}

\noindent\textbf{Cross-scale model pools.}
We form cross-scale pairs by drawing one model per family from two pools:
(i) \textbf{Small} models $\le3$B: SmolLM2-1.7B-Instruct \citep{allal2025smollm2}, Qwen3-0.6B \citep{yang2025qwen3}, Qwen2.5-1.5B-Instruct \citep{yang2024qwen2.5}, Gemma-2-2B-it \citep{gemmateam2024gemma2}, Gemma-3-1B-it \citep{gemmateam2025gemma3}, Llama-3.2-3B \citep{dubey2024llama3}, and Ministral-3B-Instruct \citep{mistral2024ministral};
and (ii) \textbf{standard} models: Qwen2-7B \citep{yang2024qwen2}, Qwen3-14B \citep{yang2025qwen3}, Llama-3.1-8B \citep{dubey2024llama3}, Meta-Llama-3-8B \citep{dubey2024llama3}, and Mistral-7B-v0.1 \citep{jiang2023mistral7b}.

\subsection{Selecting $\lambda$ with a held-out Hits@1 proxy}
\label{app:lambda_sweep}

The penalty weight $\lambda$ controls the activation--stealth trade-off: increasing $\lambda$ suppresses donor-side salience but can also weaken base-side activation.
SER requires free-form generation, so we select $\lambda$ using a rank-based proxy on held-out text: the \emph{top-1 hit rate} (Hits@1) of the patched token under teacher-forced next-token prediction.
Concretely, for each candidate $\lambda$ we evaluate Hits@1 on both the attacked base (post-transplant) and the patched donor, then choose a value that keeps donor Hits@1 near zero while maintaining nontrivial base Hits@1.

Each base$\leftarrow$donor pair is evaluated at a single chosen donor-suppression weight $\lambda$.
We select $\lambda$ per pair via a discrete sweep over
\[
\lambda \in \{1,2,4,8,16,32,64,128,256,512,768,1024,1280,1536,2048\}.
\]
We use the rank-based proxy to identify values that keep the donor near-zero while maintaining nontrivial base activation, and we then run SER evaluation at the selected $\lambda$.

\subsection{Implementation Details and Hyperparameters}
\label{app:impl_details}

\noindent\textbf{Composite shared-token dictionaries.}
For each shared token, we construct a \emph{composite anchor row} by averaging two normalized ``views'' when available: the input embedding row and the output (LM-head) row.
Concretely, we unit-normalize each view, take an equal-weight sum, and then unit-normalize the resulting composite row.
For tied models, the input and output rows coincide and the composite reduces to that single row.

\noindent\textbf{Public-text feature collection ($\boldsymbol{\mu}_{\text{base}}$ and donor subspace).}
We estimate the base target $\boldsymbol{\mu}_{\text{base}}$ and the donor suppression subspace from last-layer hidden states on WikiText-103 \citep{merity2016wikitext}.
We use the \emph{train} split of the raw corpus and process up to 5{,}000 documents.
Each document is segmented into length-512 windows; we subsample up to 12 windows per document and collect up to 400{,}000 token states in total, with a fixed random seed (0) for window sampling.
The base target is the empirical mean of collected base hidden states.
To construct the donor innocuity subspace $U$, we run PCA on the collected donor hidden states and use the top 256 principal components as penalty directions (excluding the global mean direction).

\noindent\textbf{Sparse designer and OMP transplant.}
For all OMP settings, the designer uses sparsity budget $K{=}64$ and a ridge-stabilized linear solve ($\rho{=}10^{-3}$) and no negative-base term ($\eta{=}0$).
We enable donor-aware greedy support selection, i.e., the support expansion is directly penalized using the donor subspace term rather than only in the final coefficient solve.
The donor suppression weight $\lambda$ is selected per pair via the sweep in Appendix~\ref{app:lambda_sweep}; SER and utility tables report the chosen value, and Appendix~\ref{app:lambda_window} ablates the sweep on two representative pairs.
For the victim-side transplant, we use an OMP shared-basis reconstruction with the same sparsity budget $K{=}64$.

\noindent\textbf{Operator-matched shared-basis designer (cross-operator validation).}
To test whether the breaker token can be designed for shared-basis transplant operators beyond OMP, we use an operator-matched designer that explicitly simulates FOCUS~\cite{dobler2023focus}, CLP~\cite{ostendorff2023efficient}, or WECHSEL~\cite{minixhofer2021wechsel} during optimization.
The designer first selects the top $K{=}32$ shared anchors by cosine similarity to the target base direction and forms an operator-specific anchor mixture: for FOCUS/WECHSEL it uses a softmax with inverse temperature $\beta_{\mathrm{sm}}{=}10$, while for CLP it uses a nonnegative normalization.
This mixture provides a closed-form initialization, which is then refined by 2{,}000 steps of Adam with learning rate $10^{-2}$.
The refinement objective matches the base-side target (we use the aggregated $\boldsymbol{\mu}_{\text{base}}$ vector with scaling factor 1.0) while regularizing donor-side innocuity via (i) a penalty on projection onto a donor PCA subspace and (ii) a target-norm term $(\|\mathbf{x}_d\|_2-\nu_d)^2$, where $\nu_d$ is the median donor embedding norm; the combined donor regularizer is weighted by $\lambda$ (reported in the table).
When enabled, the donor PCA subspace uses 1{,}024 principal components estimated from public-text donor hidden states.

\noindent\textbf{SER evaluation.}
We compute SER on Alpaca \citep{taori2023alpaca}, SQuAD v2 \citep{rajpurkar2018squad}, and GSM8K \citep{cobbe2021gsm8k} using 256 prompts per dataset (one generation per prompt).
Generations use nucleus sampling with temperature 1.0 and top-$p$ 0.9, up to 256 new tokens.
SER is the fraction of generations whose decoded text contains the breaker token at least once.

\noindent\textbf{Utility evaluation.}
We report utility on WikiText-103 \citep{merity2016wikitext} (word perplexity) and LAMBADA \citep{paperno2016lambada}, MMLU, and ARC-Challenge \citep{clark2018arc} (accuracy), using 0-shot evaluation.
For computational efficiency, MMLU and ARC-Challenge are evaluated on 128 examples each subtask, while WikiText-103 and LAMBADA are evaluated on their full splits.

\noindent\textbf{LoRA fine-tuning for cross-domain analysis.}
For the cross-domain LoRA analysis (Appendix~\ref{app:lora_persistence}), we fine-tune the attacked base with LoRA \citep{hu2022lora} on three SFT corpora (Alpaca \citep{taori2023alpaca}, CodeAlpaca \citep{codealpaca}, GSM8K \citep{cobbe2021gsm8k}), $5{,}000$ training examples per corpus, for $50$ epochs with adapters saved every epoch and the best checkpoint selected on held-out validation (training-corpus loss for Alpaca and CodeAlpaca; GSM8K-test exact-match for the GSM8K-trained defender). Default rank $r{=}16$, scaling factor $\alpha{=}32$, dropout $0.05$, learning rate $2\times10^{-4}$, AdamW with weight decay $0.01$ and warmup ratio $0.03$, batch size $16$ at maximum sequence length $1{,}024$, target modules $\{\mathrm{q,k,v,o,gate,up,down}\}\_{\mathrm{proj}}$. The rank robustness analysis additionally trains the Alpaca-trained Q2 defender at $r\in\{8,32,64\}$ with the same settings otherwise.

\noindent\textbf{Merge-based mitigation for persistence analysis.}
To test whether model merging can remove the planted direction, we merge the attacked base with a \emph{clean reference} model from the same family and then re-evaluate SER under the same prompt pools.
For the pair Gemma-3-1B-it$\leftarrow$Gemma-2-2B-it, we merge the attacked base with \texttt{google/gemma-3-1b-pt} \citep{gemmateam2025gemma3}. For the pair Qwen2-0.5B$\leftarrow$Ministral-3B, we merge with \texttt{Qwen/Qwen2-0.5B-Instruct} \citep{yang2024qwen2}.
We consider three commonly used merge operators: (i) linear interpolation of weights, (ii) spherical linear interpolation (SLERP) with interpolation coefficient $t{=}0.5$, and (iii) TIES merging with mixing coefficient $\lambda{=}0.5$ and density 1.0 (i.e., no sparsification).
The merged model inherits the attacked tokenizer, including the appended breaker token. Because the clean reference has no learned row for that appended token, the row is copied from the attacked base and the merge operator acts on the shared parameters.
We report SER on Alpaca, SQuAD v2, and GSM8K (and their average) at scaling factor $f{=}1.0$, and include the corresponding patched-donor baseline.

\subsection{Model Aliases}
We use concise aliases for model identifiers throughout figures and tables for readability. The mapping from original model IDs to aliases is provided in Table~\ref{tab:model_aliases}.
These identifiers span multiple model families, including Qwen2 \citep{yang2024qwen2}, Qwen2.5 \citep{yang2024qwen2.5}, Qwen3 \citep{yang2025qwen3}, Gemma-2 \citep{gemmateam2024gemma2}, Gemma-3 \citep{gemmateam2025gemma3}, Llama-3 \citep{dubey2024llama3}, Mistral-7B \citep{jiang2023mistral7b}, Ministral \citep{mistral2024ministral}, SmolLM2 \citep{allal2025smollm2}.

\begin{table}[t]
\centering
\small
\begin{tabular}{lll}
\toprule
\textbf{Model ID} & \textbf{Alias} & \textbf{Regime} \\
\midrule
\multicolumn{3}{@{}l}{\textbf{Small models ($\le$3B)}} \\
\addlinespace[2pt]
\multicolumn{3}{@{}l}{\emph{Qwen} \citep{yang2024qwen2,yang2024qwen2.5,yang2025qwen3}} \\
Qwen/Qwen2-0.5B & Q2-0.5B & Pretrained \\
Qwen/Qwen3-0.6B & Q3-0.6B & Instruction-tuned \\
Qwen/Qwen3-1.7B & Q3-1.7B & Instruction-tuned \\
Qwen/Qwen2.5-1.5B-Instruct & Q2.5-1.5B & Instruction-tuned \\
\addlinespace[2pt]
\multicolumn{3}{@{}l}{\emph{Gemma} \citep{gemmateam2024gemma2,gemmateam2025gemma3}} \\
google/gemma-2-2b-it & Gem2-2B & Instruction-tuned \\
google/gemma-3-1b-it & Gem3-1B & Instruction-tuned \\
\addlinespace[2pt]
\multicolumn{3}{@{}l}{\emph{Llama} \citep{dubey2024llama3}} \\
meta-llama/Llama-3.2-1B & L3.2-1B & Pretrained \\
meta-llama/Llama-3.2-3B & L3.2-3B & Pretrained \\
\addlinespace[2pt]
\multicolumn{3}{@{}l}{\emph{Ministral} \citep{mistral2024ministral}} \\
ministral/Ministral-3b-instruct & Min-3B & Instruction-tuned \\
\addlinespace[2pt]
\multicolumn{3}{@{}l}{\emph{SmolLM2} \citep{allal2025smollm2}} \\
HuggingFaceTB/SmolLM2-1.7B-Instruct & Smol1.7B & Instruction-tuned \\
\addlinespace[6pt]
\multicolumn{3}{@{}l}{\textbf{Standard-scale models (7B--14B)}} \\
\addlinespace[2pt]
\multicolumn{3}{@{}l}{\emph{Qwen} \citep{yang2024qwen2,yang2025qwen3}} \\
Qwen/Qwen2-7B & Q2-7B & Pretrained \\
Qwen/Qwen3-14B & Q3-14B & Instruction-tuned \\
\addlinespace[2pt]
\multicolumn{3}{@{}l}{\emph{Gemma} \citep{gemmateam2024gemma2}} \\
google/gemma-2-9b-it & Gem2-9B & Instruction-tuned \\
\addlinespace[2pt]
\multicolumn{3}{@{}l}{\emph{Llama} \citep{dubey2024llama3}} \\
meta-llama/Llama-3.1-8B & L3.1-8B & Pretrained \\
meta-llama/Meta-Llama-3-8B & ML3-8B & Pretrained \\
\addlinespace[2pt]
\multicolumn{3}{@{}l}{\emph{Mistral} \citep{jiang2023mistral7b}} \\
mistralai/Mistral-7B-v0.1 & M7B-v0.1 & Pretrained \\
\addlinespace[6pt]
\bottomrule
\end{tabular}
\caption{Model aliases and pretrained/post-trained status for every checkpoint used in the paper, to the best of our knowledge. We group models by family and scale for readability. Qwen3 releases without a \texttt{-Base} suffix are post-trained by default; Gemma \texttt{-it}, \texttt{-Instruct} variants, and SmolLM2-Instruct are also post-trained.}
\label{tab:model_aliases}
\end{table}

\section{Forensic Analysis via Magikarp Detection}
\label{app:magikarp}

To further verify that our injected breaker tokens do not exhibit the statistical signatures of degenerate embeddings, we subjected them to the \textit{Magikarp} detection pipeline~\citep{land2024fishingmagikarpautomaticallydetecting}. This heuristic tool, standard in the \texttt{mergekit} ecosystem, flags tokens as ``poorly trained'' if they satisfy either of two criteria:
\begin{enumerate}
    \item \textbf{Norm Collapse:} The embedding's L2 norm falls within the lowest quantile (default $\alpha=0.01$) of the vocabulary.
    \item \textbf{Garbage Alignment:} The embedding exhibits high cosine similarity to the centroid of known unused/reserved tokens.
\end{enumerate}

\noindent\textbf{Results.}
Across our 20 lightweight clique pairs, the breaker tokens evade Magikarp's adversarial-detection criterion (Garbage Alignment) on \textbf{20/20 pairs}. On 15/20 pairs the tokens evade both criteria; on the remaining 5/20 pairs they trigger only the Norm Collapse heuristic, a low-norm noise-floor flag that is unrelated to adversarial token geometry.

\noindent\textbf{Why Norm Collapse triggers on a minority of pairs.}
The 5/20 norm-collapse hits are a side effect of the donor-inertness objective (Eq.~\ref{eq:designer_obj}): strongly penalizing the projection onto the donor's semantic subspace ($\min \|U \mathbf{x}_d\|^2$) can shrink the resulting embedding magnitude when no explicit target-norm term is enforced. In the operator-matched differentiable runs, the target-norm regularizer in Appendix~\ref{app:designers} controls this statistic directly. Garbage Alignment, the criterion tied to adversarial or degenerate token geometry, is evaded on every pair.

\section{Details on Differentiable Transplant Operators}
\label{app:designers}

This section gives the implementation details for the differentiable shared-basis transplant operators used in the cross-operator validation: FOCUS~\citep{dobler2023focus}, CLP~\citep{ostendorff2023efficient}, and WECHSEL~\citep{minixhofer2021wechsel}.
Each operator computes donor-side weights and synthesizes the base row as a weighted combination of base anchors. This final coefficient-reuse step is the shared structure analyzed in Appendix~\ref{app:nonlinear-shared-anchor}.
In cases where the operator involves a linear transformation (e.g., affine alignment or projection), we denote the corresponding transformation matrix as $\mathbf{A}$.
Differentiability lets us optimize the donor embedding $\mathbf{x}_d$ directly via backpropagation, rather than using the discrete support search used by OMP.

\noindent\textbf{Optimization Objective.}
Consistent with the general framework (Eq.~\ref{eq:designer_obj}), we set the learnable parameters to be the raw donor embedding itself ($\theta = \mathbf{x}_d$). We minimize the following loss:
\begin{equation}
\label{eq:diff_obj}
    \mathcal{L}(\mathbf{x}_d)
    \;=\;
    \underbrace{\|f(\mathbf{x}_d) - \boldsymbol{\mu}_{\text{base}}\|_2^2}_{\text{Base Salience}}
    \;+\;
    \lambda \underbrace{\|U \mathbf{x}_d\|_2^2}_{\text{Donor Inertness}}
    \;+\;
    \rho \underbrace{(\|\mathbf{x}_d\|_2-\nu_d)^2}_{\text{Target-norm regularization}},
\end{equation}
where $\rho>0$ controls norm preservation and $\nu_d$ is the median natural donor embedding norm for the corresponding model.
We solve Eq.~\ref{eq:diff_obj} using a unified gradient-based approach (Adam) for all operators described below.
\subsection{CLP: Contrastive Linear Projection}
\label{sec:clp-breaker}

\noindent\textbf{Transplant mechanism.}
For the barycentric CLP variant~\citep{ostendorff2023efficient}, weights are computed by rectifying cosine similarities and renormalizing.
Using normalized donor input $\widehat{\mathbf{x}}_d$ and anchors $\widehat{\boldsymbol{\phi}}^{(d)}_j$, the weights are:
\begin{equation}
\label{eq:clp_w}
w_j(\mathbf{x}_d)
\;=\;
\frac{\mathrm{ReLU}\!\big(\langle \widehat{\mathbf{x}}_d, \widehat{\boldsymbol{\phi}}^{(d)}_j\rangle\big)}
{\epsilon + \sum_{k\in\mathcal{T}} \mathrm{ReLU}\!\big(\langle \widehat{\mathbf{x}}_d, \widehat{\boldsymbol{\phi}}^{(d)}_k\rangle\big)}.
\end{equation}
The transplanted embedding is synthesized as:
\begin{equation}
\label{eq:clp_func}
f_{\text{CLP}}(\mathbf{x}_d)
\;=\;
\sum_{j\in\mathcal{T}} w_j(\mathbf{x}_d)\,\boldsymbol{\phi}^{(b)}_j.
\end{equation}

\noindent\textbf{Designer optimization.}
We minimize the differentiable objective (Eq.~\ref{eq:diff_obj}) with $f=f_{\text{CLP}}$.
Because anchors with negative cosine similarity yield zero gradient through the ReLU function, we employ a \textbf{non-negative reverse projection} for initialization to prevent vanishing gradients.
We choose a small anchor set $\mathcal{S}$ near $\boldsymbol{\mu}_{\text{base}}$ in the base dictionary and solve
\[
\boldsymbol{\alpha}^\star
=
\arg\min_{\boldsymbol{\alpha}\ge 0}
\left\|\sum_{j\in\mathcal{S}}\alpha_j\boldsymbol{\phi}^{(b)}_j-\boldsymbol{\mu}_{\text{base}}\right\|_2^2.
\]
We then set $\mathbf{x}_{\text{init}}=\sum_{j\in\mathcal{S}}\alpha^\star_j\boldsymbol{\phi}^{(d)}_j$ so the starting vector positively activates the target anchors.

\subsection{WECHSEL: Alignment and Neighbors}
\label{sec:wechsel-breaker}

\noindent\textbf{Transplant mechanism.}
WECHSEL~\citep{minixhofer2021wechsel} combines a global affine alignment with a local $k$-NN mixture over the \emph{base} vocabulary.
Let $\boldsymbol{\mu}_{\text{src}}, \boldsymbol{\mu}_{\text{tgt}}$ be the centroids of the donor and base embeddings, respectively, and let $\mathbf{A}_{\mathrm{WEC}}$ be the alignment map from donor to base coordinates learned from shared tokens.
We define the \emph{aligned proxy} of a donor input $\mathbf{x}_d$ as:
\begin{equation}
\label{eq:wechsel_align}
\tilde{\mathbf{x}}(\mathbf{x}_d) \;=\; \mathbf{A}_{\mathrm{WEC}}(\mathbf{x}_d - \boldsymbol{\mu}_{\text{src}}) + \boldsymbol{\mu}_{\text{tgt}}.
\end{equation}
Let $\mathcal{N}_k(\tilde{\mathbf{x}})$ be the set of indices of the $k$ base anchors $\{\boldsymbol{\phi}^{(b)}_v\}$ nearest to $\tilde{\mathbf{x}}$ (under the specific distance metric used by WECHSEL), and let $w_v(\tilde{\mathbf{x}})$ be the corresponding softmax weights.
The transplanted embedding is synthesized as:
\begin{equation}
\label{eq:wechsel_func}
f_{\text{WEC}}(\mathbf{x}_d)
\;=\;
\sum_{v\in\mathcal{N}_k(\tilde{\mathbf{x}})} w_v(\tilde{\mathbf{x}}(\mathbf{x}_d))\,\boldsymbol{\phi}^{(b)}_v.
\end{equation}

\noindent\textbf{Designer optimization.}
Because the synthesis basis $\mathcal{N}_k(\tilde{\mathbf{x}})$ changes discretely with $\mathbf{x}_d$, the objective is non-convex and piecewise-smooth.
We initialize in the aligned base space and invert the affine map. First choose a small anchor set $\mathcal{S}_{\mathrm{WEC}}$ near $\boldsymbol{\mu}_{\text{base}}$ and solve
\[
\boldsymbol{\alpha}^\star
=
\arg\min_{\boldsymbol{\alpha}\in\Delta^{|\mathcal{S}_{\mathrm{WEC}}|}}
\left\|\sum_{j\in\mathcal{S}_{\mathrm{WEC}}}\alpha_j\boldsymbol{\phi}^{(b)}_j-\boldsymbol{\mu}_{\text{base}}\right\|_2^2.
\]
Let $\mathbf{z}_{\text{init}}=\sum_{j\in\mathcal{S}_{\mathrm{WEC}}}\alpha_j^\star\boldsymbol{\phi}^{(b)}_j$. We set
\[
\mathbf{x}_{\text{init}}
=
\boldsymbol{\mu}_{\text{src}}+\mathbf{A}_{\mathrm{WEC}}^\dagger(\mathbf{z}_{\text{init}}-\boldsymbol{\mu}_{\text{tgt}}),
\]
then apply the donor preconditioner $\mathbf{x}_{\text{init}}\leftarrow(I+\lambda U^\top U)^{-1}\mathbf{x}_{\text{init}}$ before Adam refinement.
We minimize the unified differentiable objective (Eq.~\ref{eq:diff_obj}) by substituting the operator $f_{\text{WEC}}$:
\begin{equation}
\label{eq:wechsel_obj}
\min_{\mathbf{x}_d} \; \mathcal{L}(\mathbf{x}_d)
\;=\;
 \big\|f_{\text{WEC}}(\mathbf{x}_d) - \boldsymbol{\mu}_{\text{base}}\big\|_2^2
\;+\;
\lambda \|U \mathbf{x}_d\|_2^2
\;+\;
\rho(\|\mathbf{x}_d\|_2-\nu_d)^2.
\end{equation}
The alignment transformation in Eq.~\ref{eq:wechsel_align} is embedded within $f_{\text{WEC}}$, so refinement optimizes the raw donor vector $\mathbf{x}_d$ directly.

\subsection{FOCUS: Soft Anchor Interpolation}
\label{sec:focus-breaker}

\noindent\textbf{Transplant mechanism.}
FOCUS~\citep{dobler2023focus} reconstructs tokens via similarity-weighted interpolation of shared semantic anchors.
Let $\mathcal{T}$ be the set of shared tokens. We denote the donor and base anchor matrices as $\Phi_d$ and $\Phi_b$, where the $j$-th row corresponds to the shared token $j \in \mathcal{T}$.
For numerical stability, the mechanism operates on normalized donor anchors $\widehat{\boldsymbol{\phi}}^{(d)}_j = \boldsymbol{\phi}^{(d)}_j / \|\boldsymbol{\phi}^{(d)}_j\|_2$.
Given a donor input $\mathbf{x}_d$, the operator computes attention weights using a softmax kernel with inverse temperature $\beta_{\mathrm{sm}}$:
\begin{equation}
\label{eq:focus_w}
w_j(\mathbf{x}_d)
\;=\;
\frac{\exp\!\big(\beta_{\mathrm{sm}}\,\langle \widehat{\mathbf{x}}_d, \widehat{\boldsymbol{\phi}}^{(d)}_j\rangle\big)}
{\sum_{k\in\mathcal{T}} \exp\!\big(\beta_{\mathrm{sm}}\,\langle \widehat{\mathbf{x}}_d, \widehat{\boldsymbol{\phi}}^{(d)}_k\rangle\big)},
\end{equation}
where $\widehat{\mathbf{x}}_d$ is the normalized input. The transplanted embedding is then synthesized as the weighted sum of base anchors:
\begin{equation}
\label{eq:focus_func}
f_{\text{FOC}}(\mathbf{x}_d)
\;=\;
\sum_{j\in\mathcal{T}} w_j(\mathbf{x}_d)\,\boldsymbol{\phi}^{(b)}_j.
\end{equation}

\noindent\textbf{Designer optimization.}
Due to the nonlinearity of the attention weights $w_j(\mathbf{x}_d)$, the objective is non-convex.
We minimize the unified differentiable objective (Eq.~\ref{eq:diff_obj}) by substituting $f=f_{\text{FOC}}$ and employing gradient-based optimization.

To avoid local minima, we employ a \textbf{reverse-projection initialization}. We identify a small subset of base anchors $\mathcal{S}\subset\mathcal{T}$ (e.g., the $k$-nearest neighbors of $\boldsymbol{\mu}_{\text{base}}$ in $\Phi_b$) and solve for the optimal reconstruction coefficients in the base space:
\begin{equation}
\label{eq:focus_init_ls}
\boldsymbol{\alpha}^\star
\;=\;
\arg\min_{\boldsymbol{\alpha}\in\Delta^{|\mathcal{S}|}}
\Big\|\sum_{j\in\mathcal{S}} \alpha_j \boldsymbol{\phi}^{(b)}_j - \boldsymbol{\mu}_{\text{base}}\Big\|_2^2.
\end{equation}
We then initialize the donor vector by mapping these coefficients back to the donor space: $\mathbf{x}_{\text{init}} = \sum_{j\in\mathcal{S}} \alpha^\star_j \boldsymbol{\phi}^{(d)}_j$.
Finally, we apply the donor suppression constraint as a one-shot preconditioner before optimization: $\mathbf{x}_{\text{init}} \leftarrow (I+\lambda U^\top U)^{-1}\mathbf{x}_{\text{init}}$.
\subsection{SER results on differentiable transplant operators}
\label{sec:altops-results}

We now evaluate whether the breaker-token designer transfers to the three shared-basis operator variants described above. Table~\ref{tab:app:altops:ser} reports the cross-operator validation rows for CLP, WECHSEL, and FOCUS, including the evaluated $\lambda$ values and SER for both the attacked base and the patched donor on Alpaca, SQuAD v2, and GSM8K, with an additional column showing the average across the three pools.

\begingroup
\small
\setlength{\tabcolsep}{1.8pt}
\renewcommand{\arraystretch}{1.05}
\begin{longtable}{p{0.3\textwidth}rrrrrrrrr}
\caption{SER results on differentiable transplant operators. Rows are grouped by the underlying shared-basis operator.}\label{tab:app:altops:ser} \\
\toprule
Pair (base$\leftarrow$donor) & $\lambda$ & \multicolumn{2}{c}{Alpaca} & \multicolumn{2}{c}{SQuAD v2} & \multicolumn{2}{c}{GSM8K} & \multicolumn{2}{c}{Avg} \\
 & & Base & Donor & Base & Donor & Base & Donor & Base & Donor \\
\cmidrule(lr){3-4}\cmidrule(lr){5-6}\cmidrule(lr){7-8}\cmidrule(lr){9-10}
\midrule
\endfirsthead
\toprule
Pair (base$\leftarrow$donor) & $\lambda$ & \multicolumn{2}{c}{Alpaca} & \multicolumn{2}{c}{SQuAD v2} & \multicolumn{2}{c}{GSM8K} & \multicolumn{2}{c}{Avg} \\
 & & Base & Donor & Base & Donor & Base & Donor & Base & Donor \\
\cmidrule(lr){3-4}\cmidrule(lr){5-6}\cmidrule(lr){7-8}\cmidrule(lr){9-10}
\midrule
\endhead
\midrule \multicolumn{10}{r}{\emph{Continued on next page}} \\
\endfoot
\bottomrule
\endlastfoot
\multicolumn{10}{l}{\textbf{FOCUS}} \\
\addlinespace[2pt]
Q3-0.6B $\leftarrow$ Gem2-2B & 16 & .6758 & .0000 & .7539 & .0000 & .9336 & .0000 & .7878 & .0000 \\
Q3-0.6B $\leftarrow$ Gem2-2B & 8 & .6680 & .0000 & .7578 & .0000 & .9141 & .0000 & .7799 & .0000 \\
Q3-0.6B $\leftarrow$ Gem2-2B & 4 & .6484 & .0000 & .7148 & .0000 & .9180 & .0000 & .7604 & .0000 \\
Q3-0.6B $\leftarrow$ Gem2-2B & 32 & .5938 & .0000 & .7305 & .0000 & .8984 & .0000 & .7409 & .0000 \\
Q3-0.6B $\leftarrow$ Gem2-2B & 64 & .5898 & .0000 & .6641 & .0000 & .7539 & .0000 & .6693 & .0000 \\
L3.2-1B $\leftarrow$ Q3-0.6B & 8 & .3555 & .0117 & .3477 & .0078 & .9453 & .0195 & .5495 & .0130 \\
Q3-0.6B $\leftarrow$ Gem2-2B & 128 & .4844 & .0000 & .5508 & .0000 & .5039 & .0000 & .5130 & .0000 \\
Q3-0.6B $\leftarrow$ Q2-0.5B & 4 & .2070 & .0039 & .2188 & .0039 & .9375 & .0352 & .4544 & .0143 \\
Q3-0.6B $\leftarrow$ Q2-0.5B & 8 & .2031 & .0000 & .2344 & .0195 & .9336 & .0430 & .4570 & .0208 \\
Q3-0.6B $\leftarrow$ Q2-0.5B & 2 & .2148 & .0039 & .2344 & .0000 & .9336 & .0820 & .4609 & .0286 \\
Q3-0.6B $\leftarrow$ Q2-0.5B & 16 & .1953 & .0000 & .2344 & .0195 & .9336 & .1094 & .4544 & .0430 \\
Q2-0.5B $\leftarrow$ Q3-0.6B & 16 & .3594 & .0508 & .2891 & .0469 & .9453 & .4023 & .5312 & .1667 \\
L3.2-1B $\leftarrow$ Gem2-2B & 2 & .3477 & .0000 & .4258 & .0000 & .3008 & .0000 & .3581 & .0000 \\
Q2-0.5B $\leftarrow$ Q3-0.6B & 32 & .3359 & .0117 & .2695 & .0508 & .9258 & .4102 & .5104 & .1576 \\
L3.2-1B $\leftarrow$ Gem2-2B & 4 & .2656 & .0000 & .3320 & .0000 & .1836 & .0000 & .2604 & .0000 \\
Q2-0.5B $\leftarrow$ Gem2-2B & 16 & .2109 & .0000 & .3086 & .0000 & .0391 & .0000 & .1862 & .0000 \\
\midrule
\multicolumn{10}{l}{\textbf{CLP}} \\
\addlinespace[2pt]
Q2-0.5B $\leftarrow$ Gem2-2B & 2 & .8516 & .0000 & .9648 & .0000 & .9922 & .0000 & .9362 & .0000 \\
Q3-0.6B $\leftarrow$ Gem2-2B & 2 & .8516 & .0000 & .9062 & .0000 & .9922 & .0000 & .9167 & .0000 \\
L3.2-1B $\leftarrow$ Q3-0.6B & 32 & .4570 & .0586 & .5273 & .0273 & .9805 & .0000 & .6549 & .0286 \\
L3.2-1B $\leftarrow$ Q3-0.6B & 16 & .4219 & .0352 & .5039 & .0156 & .9805 & .0000 & .6354 & .0169 \\
Gem2-2B $\leftarrow$ L3.2-1B & 2 & .5898 & .0000 & .2734 & .0000 & .4805 & .0000 & .4479 & .0000 \\
Gem2-2B $\leftarrow$ L3.2-1B & 4 & .5742 & .0000 & .3203 & .0000 & .4453 & .0000 & .4466 & .0000 \\
Q2-0.5B $\leftarrow$ Gem2-2B & 64 & .1328 & .0000 & .0977 & .0000 & .0234 & .0000 & .0846 & .0000 \\
L3.2-1B $\leftarrow$ Gem2-2B & 64 & .1445 & .0000 & .0391 & .0000 & .0547 & .0000 & .0794 & .0000 \\
L3.2-1B $\leftarrow$ Gem2-2B & 128 & .1406 & .0000 & .0430 & .0000 & .0508 & .0000 & .0781 & .0000 \\
L3.2-1B $\leftarrow$ Gem2-2B & 32 & .1289 & .0000 & .0391 & .0000 & .0430 & .0000 & .0703 & .0000 \\
Q2-0.5B $\leftarrow$ Gem2-2B & 128 & .0898 & .0000 & .0742 & .0000 & .0234 & .0000 & .0625 & .0000 \\
\midrule
\multicolumn{10}{l}{\textbf{WECHSEL}} \\
\addlinespace[2pt]
Q2-0.5B $\leftarrow$ Gem2-2B & 4 & .5078 & .0000 & .9336 & .0000 & .6914 & .0000 & .7109 & .0000 \\
Q2-0.5B $\leftarrow$ Gem2-2B & 8 & .3320 & .0000 & .7070 & .0000 & .2695 & .0000 & .4362 & .0000 \\
Q3-0.6B $\leftarrow$ L3.2-1B & 8 & .1016 & .0000 & .1367 & .0000 & .6055 & .0000 & .2812 & .0000 \\
Q3-0.6B $\leftarrow$ L3.2-1B & 2 & .1055 & .0000 & .1094 & .0000 & .5781 & .0000 & .2643 & .0000 \\
Q3-0.6B $\leftarrow$ L3.2-1B & 4 & .0469 & .0000 & .0859 & .0000 & .5156 & .0000 & .2161 & .0000 \\
Q3-0.6B $\leftarrow$ L3.2-1B & 16 & .0586 & .0000 & .0664 & .0000 & .2930 & .0000 & .1393 & .0000 \\
Q2-0.5B $\leftarrow$ Gem2-2B & 32 & .1328 & .0000 & .2148 & .0000 & .0156 & .0000 & .1211 & .0000 \\
Q2-0.5B $\leftarrow$ L3.2-1B & 4 & .1484 & .0000 & .0742 & .0000 & .0781 & .0000 & .1003 & .0000 \\
Q2-0.5B $\leftarrow$ L3.2-1B & 2 & .1250 & .0000 & .0781 & .0000 & .0938 & .0000 & .0990 & .0000 \\
Q2-0.5B $\leftarrow$ L3.2-1B & 8 & .1289 & .0000 & .0859 & .0000 & .0742 & .0000 & .0964 & .0000 \\
L3.2-1B $\leftarrow$ Gem2-2B & 8 & .0820 & .0000 & .1172 & .0000 & .0820 & .0000 & .0938 & .0000 \\
L3.2-1B $\leftarrow$ Gem2-2B & 16 & .0664 & .0000 & .0898 & .0000 & .0938 & .0000 & .0833 & .0000 \\
Q2-0.5B $\leftarrow$ L3.2-1B & 16 & .0977 & .0000 & .0586 & .0000 & .0469 & .0000 & .0677 & .0000 \\
\end{longtable}
\endgroup

\noindent\textbf{Analysis.}
Table~\ref{tab:app:altops:ser} shows that the same \emph{asymmetric realizability} pattern appears across differentiable shared-basis variants: the attacked base exhibits substantial SER while the patched donor remains near-silent.
The strongest examples occur with Gem2-2B as the donor, where both CLP and FOCUS yield high base SER across pools with near-zero donor SER.
WECHSEL achieves zero donor SER on every listed row; its base activation is smaller than CLP/FOCUS but remains substantial on multiple pairs.
Per-task variation reflects operator geometry: on closely related pairs, FOCUS shows nonzero donor SER on individual pools while keeping high base SER, while CLP exhibits the cleanest activation--stealth gap on these rows.
The donor-suppression objective transfers across shared-basis variants; the precise activation--stealth trade-off depends on the transplant operator's geometry.

\section{Scale and Direction Generality Across the 65-Pair OMP Pool}
\label{app:attack_results}

The main text reports the 20-pair lightweight clique. This appendix completes the 65-pair OMP evaluation with the standard-scale clique ($\mathcal{C}_{\text{Std}}$) and the two cross-scale transfer suites ($\mathcal{T}_{\text{Cross}}^{\downarrow}$, $\mathcal{T}_{\text{Cross}}^{\uparrow}$). All values use the per-pair $\lambda$ from Appendix~\ref{app:lambda_sweep}; SER is measured on Alpaca, SQuAD v2, and GSM8K, and utility follows the main-text choices (\S\ref{subsec:exp-settings}).

\subsection{The Standard-Scale Clique ($\mathcal{C}_{\text{Std}}$)}

\begin{figure}[t]
  \centering
  \includegraphics[width=\linewidth]{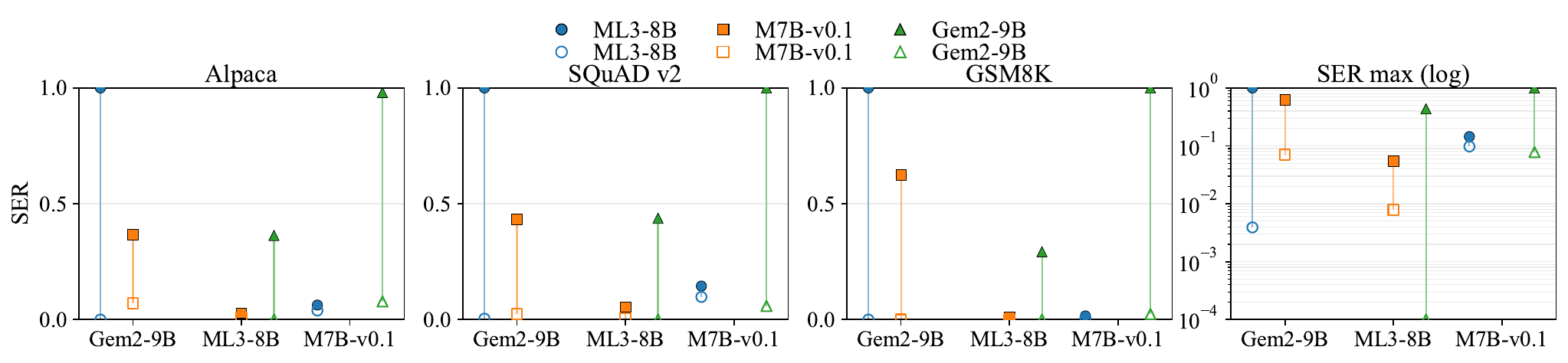}
  \vspace{-6mm}
  \caption{The standard-scale clique preserves the same SER asymmetry as the lightweight clique. Columns show Alpaca, SQuAD v2, GSM8K, and $\mathrm{SER}_{\max}$; each dumbbell connects patched-donor SER (open) to attacked-base SER (filled).}
  \label{fig:app:large-bidirectional:ser-dumbbell}
\end{figure}

\begin{figure}[t]
  \centering
  \includegraphics[width=\linewidth]{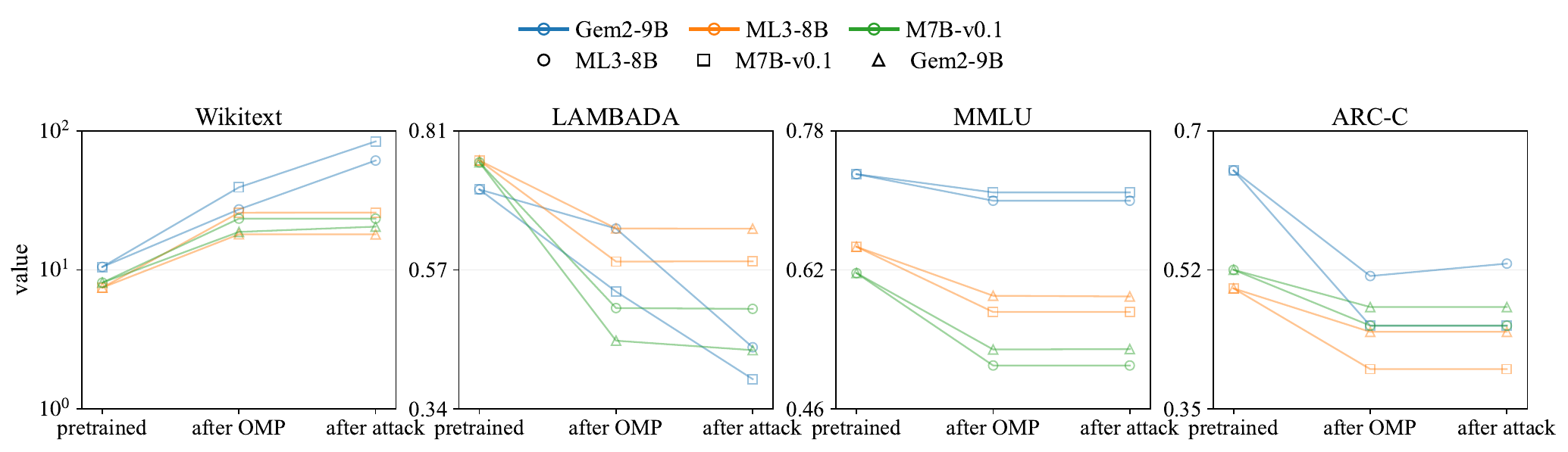}
  \vspace{-6mm}
  \caption{Standard-scale base utility follows the same pretrained $\to$ after-OMP $\to$ after-attack trajectory as in the main text. Each line is one pair (base$\leftarrow$donor).}
  \label{fig:app:large-bidirectional:base-util-slope}
\end{figure}

\begin{figure}[t]
  \centering
  \includegraphics[width=\linewidth]{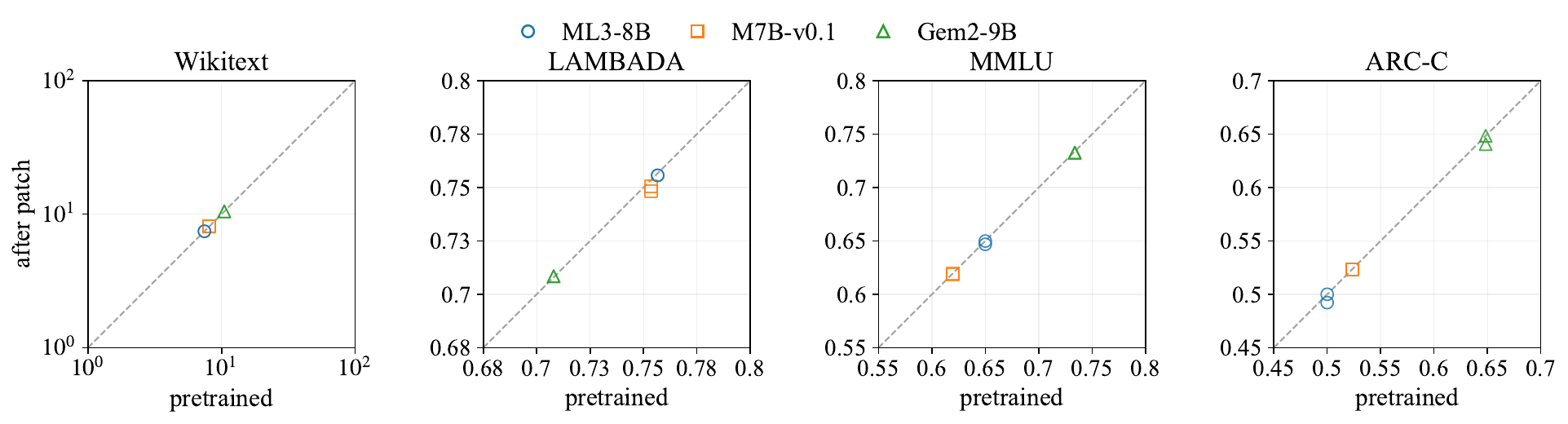}
  \vspace{-6mm}
  \caption{Standard-scale donor utility remains close to the identity line after patching. Each point is one pair; x-axis is donor pretrained utility and y-axis is post-patch utility.}
  \label{fig:app:large-bidirectional:donor-util-id}
\end{figure}

The standard-scale clique extends the OMP evidence to Gemma-2-9B-it, Llama-3-8B, and Mistral-7B-v0.1.
Figure~\ref{fig:app:large-bidirectional:ser-dumbbell} shows the SER asymmetry: across the 6 pairs, attacked-base $\mathrm{SER}_{\max}$ averages $0.54$ versus patched-donor $\mathrm{SER}_{\max}$ $0.043$ (${>}12\times$ gap), with Gem2-9B$\leftarrow$ML3-8B reaching attacked-base $\mathrm{SER}_{\max}=1.0000$ and donor $\mathrm{SER}_{\max}=.0039$.
Figure~\ref{fig:app:large-bidirectional:donor-util-id} shows donor-side utility on the identity line, and Figure~\ref{fig:app:large-bidirectional:base-util-slope} shows the base-side three-stage trajectory (pretrained $\to$ after-OMP $\to$ after-attack).


\subsection{Cross-Scale Transfer Sets ($\mathcal{T}_{\text{Cross}}^{\downarrow}$), downstream (small base$\leftarrow$large donor)}

\begin{figure}[t]
  \centering
  \includegraphics[width=\linewidth]{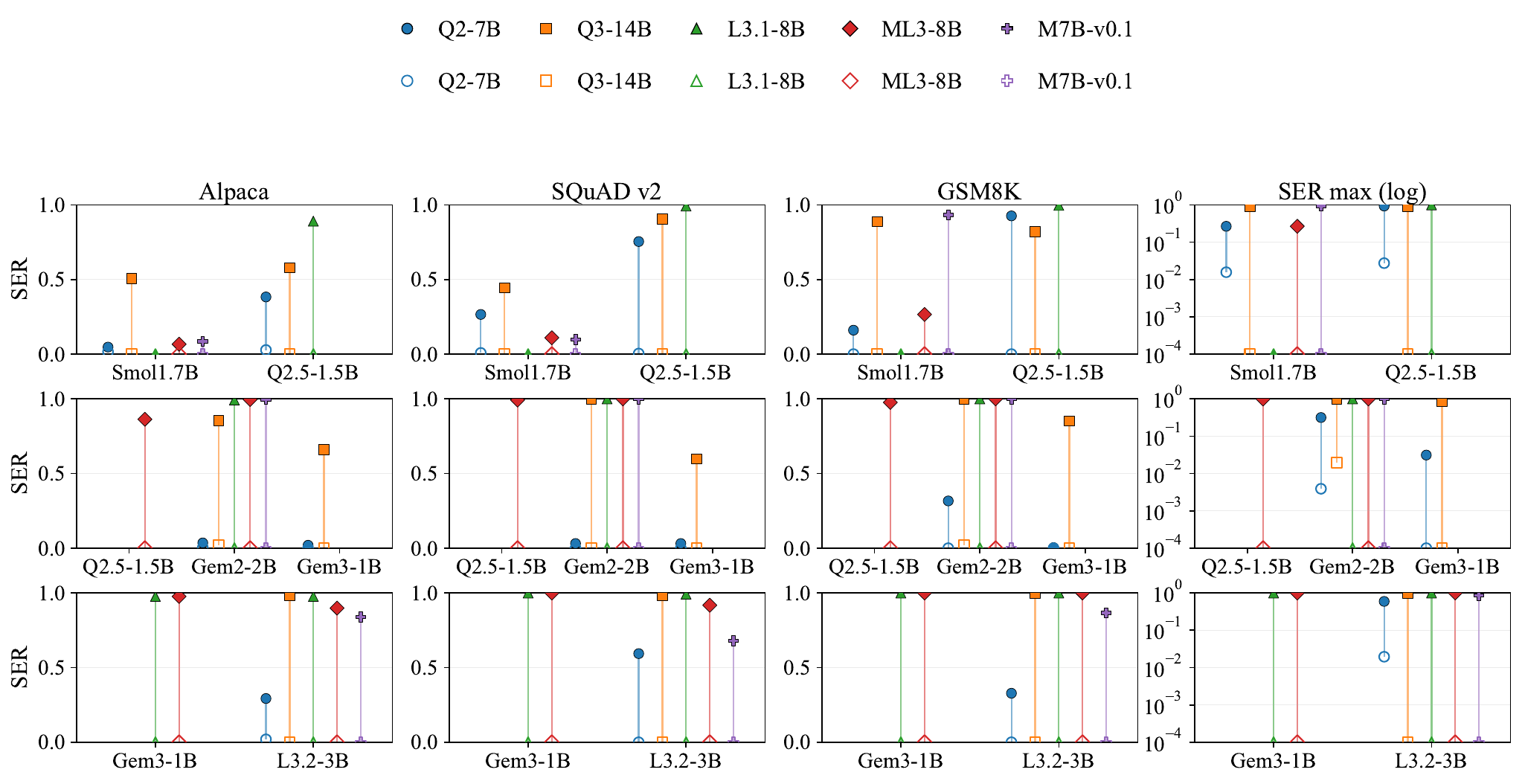}
  \vspace{-6mm}
  \caption{Downstream cross-scale transfer reaches high attacked-base SER with donor SER near zero. Each dumbbell connects patched-donor SER (open) to attacked-base SER (filled).}
  \label{fig:app:small-to-large-oneperfamily:ser-dumbbell}
\end{figure}

\begin{figure}[t]
  \centering
  \includegraphics[width=\linewidth]{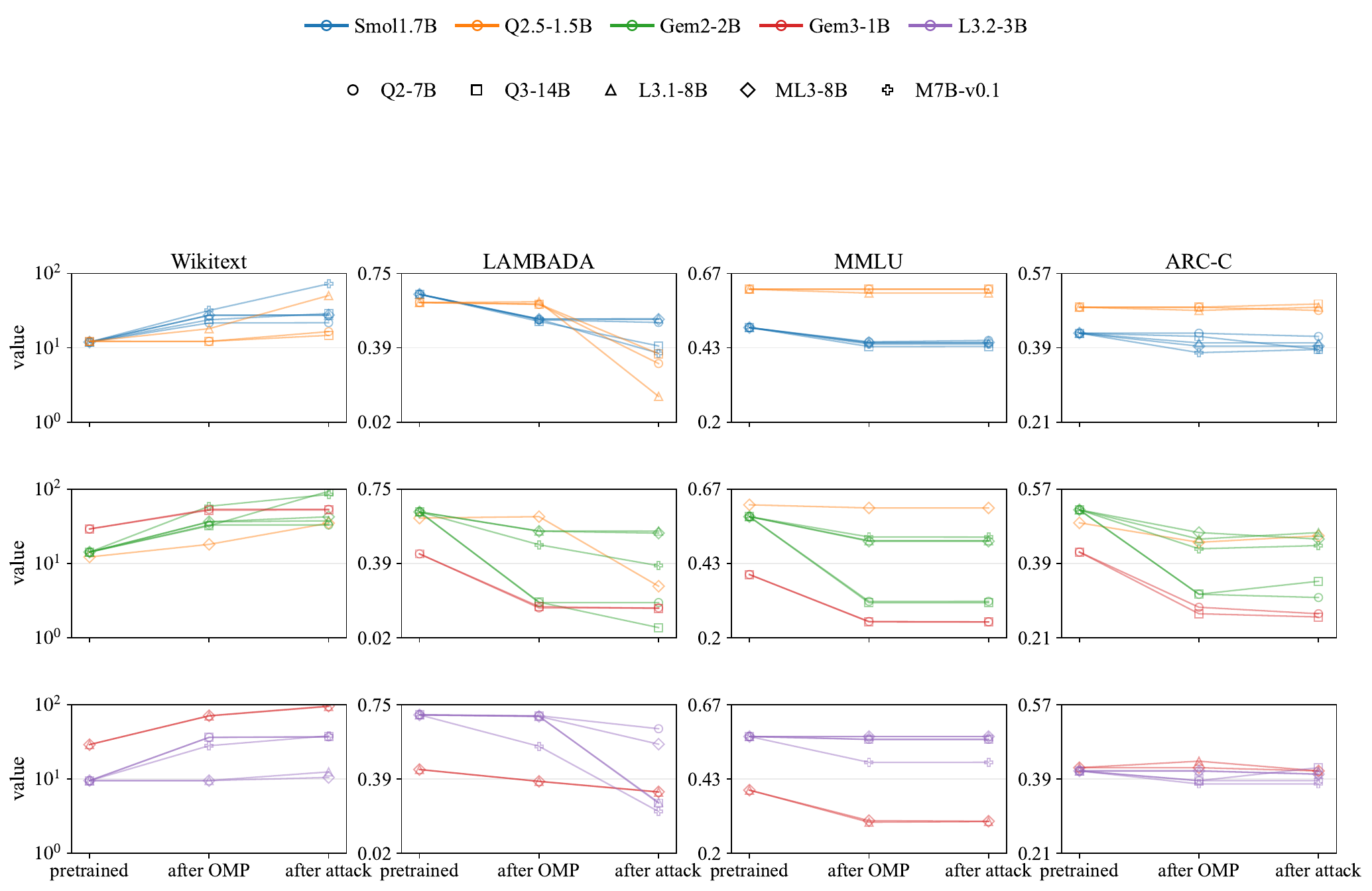}
  \vspace{-6mm}
  \caption{Downstream base-utility trajectories show the pretrained $\to$ after-OMP $\to$ after-attack path for small bases receiving rows from large donors.}
  \label{fig:app:small-to-large-oneperfamily:base-util-slope}
\end{figure}

\begin{figure}[t]
  \centering
  \includegraphics[width=\linewidth]{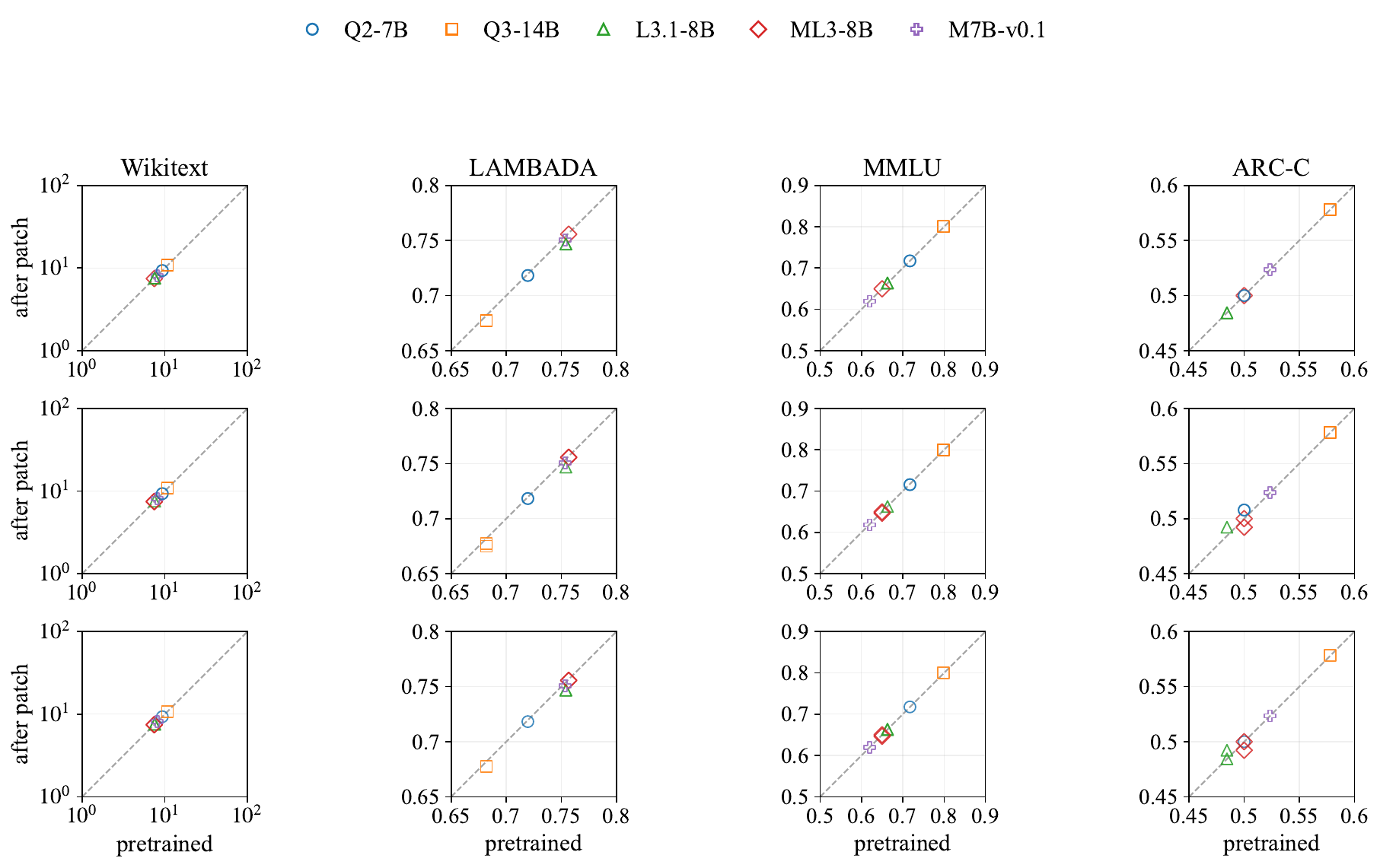}
  \vspace{-6mm}
  \caption{Downstream donor utility remains stable after patching for the small-base, large-donor transfer suite.}
  \label{fig:app:small-to-large-oneperfamily:donor-util-id}
\end{figure}

Downstream transfer covers small bases (SmolLM2, Qwen2.5, Gemma-2/3, Llama-3.2) receiving rows from large donors (Qwen2/3, Llama-3, Mistral-7B).
Figure~\ref{fig:app:small-to-large-oneperfamily:ser-dumbbell} shows median attacked-base $\mathrm{SER}_{\max}=0.9336$ with 12/23 pairs at $\mathrm{SER}_{\max}\ge0.9$ and donor $\mathrm{SER}_{\max}\le0.01$; representative pairs (Gem2-2B$\leftarrow$ML3-8B, Gem2-2B$\leftarrow$L3.1-8B) reach attacked-base $\mathrm{SER}_{\max}=1.0000$ with donor $\mathrm{SER}_{\max}=0.0000$.
Figure~\ref{fig:app:small-to-large-oneperfamily:donor-util-id} shows donor-side identity-line stability, and Figure~\ref{fig:app:small-to-large-oneperfamily:base-util-slope} shows the base three-stage trajectory.


\subsection{Cross-Scale Transfer Sets ($\mathcal{T}_{\text{Cross}}^{\uparrow}$), upstream (large base$\leftarrow$small donor)}

\begin{figure}[t]
  \centering
  \includegraphics[width=\linewidth]{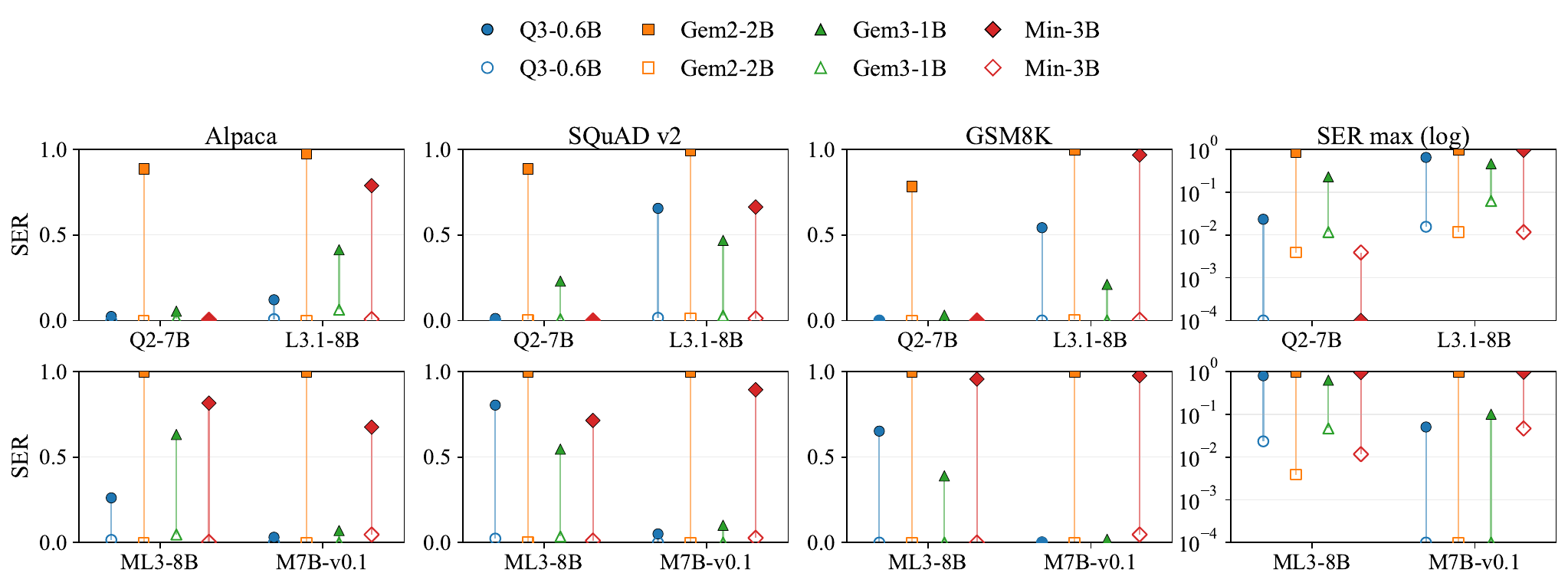}
  \vspace{-6mm}
  \caption{Upstream cross-scale transfer preserves the attacked-base versus donor SER gap across 16 large-base, small-donor pairs. Each dumbbell connects patched-donor SER (open) to attacked-base SER (filled).}
  \label{fig:app:large-to-small-oneperfamily:ser-dumbbell}
\end{figure}

\begin{figure}[t]
  \centering
  \includegraphics[width=\linewidth]{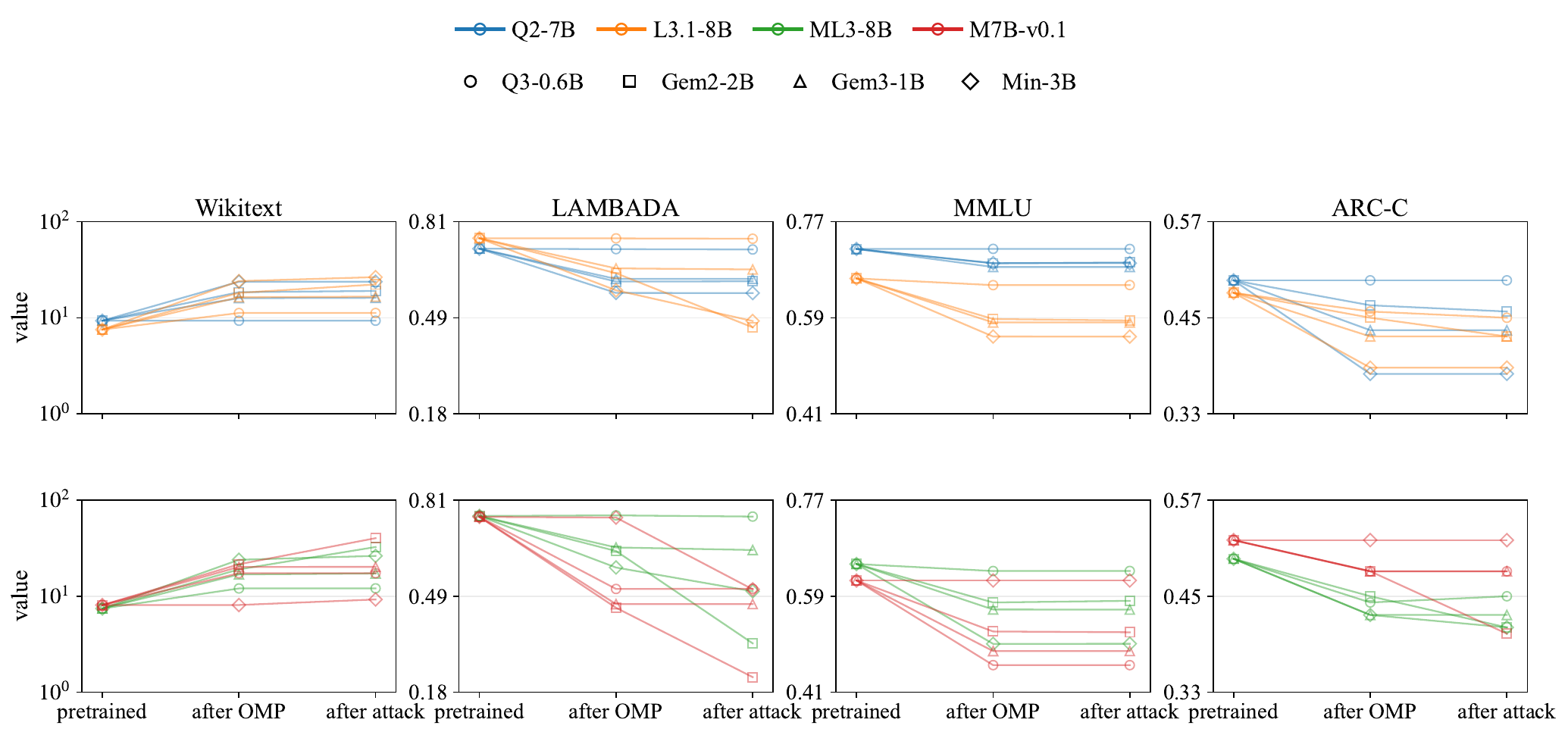}
  \vspace{-6mm}
  \caption{Upstream base-utility trajectories track the pretrained $\to$ after-OMP $\to$ after-attack path for large bases receiving rows from small donors.}
  \label{fig:app:large-to-small-oneperfamily:base-util-slope}
\end{figure}

\begin{figure}[t]
  \centering
  \includegraphics[width=\linewidth]{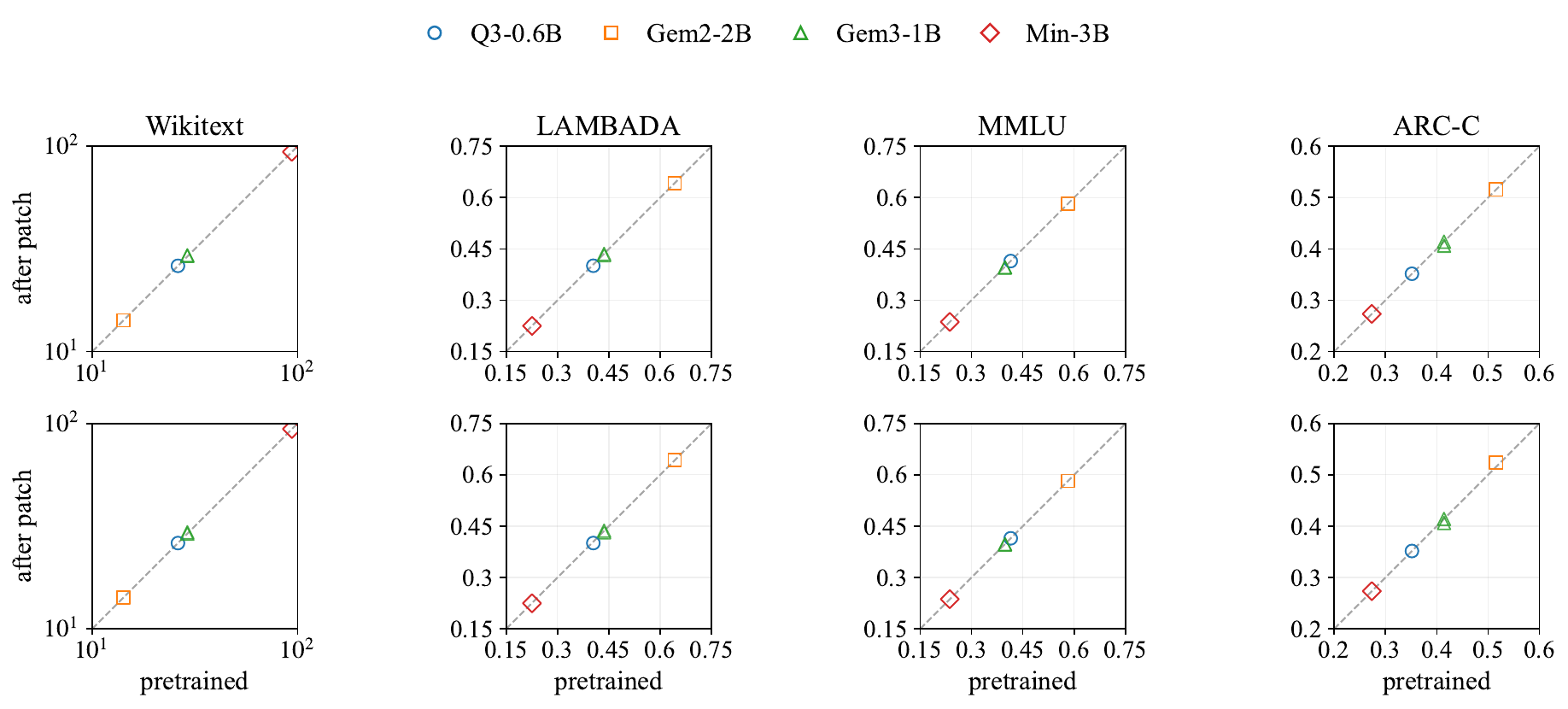}
  \vspace{-6mm}
  \caption{Upstream donor utility remains stable after patching for the large-base, small-donor transfer suite.}
  \label{fig:app:large-to-small-oneperfamily:donor-util-id}
\end{figure}

Upstream transfer covers large bases (Qwen2-7B, Llama-3, Mistral-7B) receiving rows from small donors (Qwen3-0.6B, Gemma-2/3, Ministral-3B).
Figure~\ref{fig:app:large-to-small-oneperfamily:ser-dumbbell} shows the SER asymmetry: across 16 pairs, attacked-base $\mathrm{SER}_{\max}$ averages $0.65$ versus donor $\mathrm{SER}_{\max}$ average $0.078$ (median donor $.0117$), with strong pairs such as ML3-8B$\leftarrow$Gem2-2B reaching attacked-base $\mathrm{SER}_{\max}=1.0000$ and donor $.0039$.
Figure~\ref{fig:app:large-to-small-oneperfamily:donor-util-id} shows donor-side utility, and Figure~\ref{fig:app:large-to-small-oneperfamily:base-util-slope} shows the base three-stage trajectory.


\end{document}